\documentclass[11pt]{article}

\usepackage[preprint]{acl}

\usepackage{times}
\usepackage{latexsym}

\usepackage[T1]{fontenc}

\usepackage[utf8]{inputenc}

\usepackage{microtype}

\usepackage{inconsolata}

\usepackage{graphicx}

\usepackage[most]{tcolorbox}
\usepackage{algorithm}
\usepackage{algpseudocode}  

\usepackage[utf8]{inputenc} 
\usepackage[T1]{fontenc}    
\usepackage{url}            
\usepackage{booktabs}       
\usepackage{amsfonts}       
\usepackage{nicefrac}       
\usepackage{microtype}      
\usepackage{bm}
\usepackage{enumitem}
\usepackage{graphicx}
\usepackage{amsmath}
\usepackage{subfigure}
\usepackage{amssymb}
\usepackage{caption}

\usepackage{comment}
\usepackage{array}
\usepackage{multirow}
\usepackage{wrapfig,lipsum,booktabs}
\usepackage{makecell}
\usepackage{pifont}
\usepackage{tabularx}
\usepackage{color,colortbl}
\definecolor{Gray}{gray}{0.9}

\usepackage{longtable}
\usepackage{fontawesome5}  
\usepackage{etoc}

\usepackage{listings}
\usepackage{xcolor}
\lstdefinestyle{mypython}{
  language=Python,
  basicstyle=\ttfamily\tiny,
  keywordstyle=\color{blue!70!black}\bfseries,
  commentstyle=\color{teal!60!black}\itshape,
  stringstyle=\color{orange!70!black},
  showstringspaces=false,
  breaklines=true,
  breakatwhitespace=true,
  tabsize=4,
  numbers=left,
  numberstyle=\tiny,
  frame=single,
  framerule=0.3pt,
  rulecolor=\color{black!20},
  columns=fullflexible
}

\usepackage{minted}
\setminted{
  autogobble,
  breaklines,
  breakanywhere,
  fontsize=\tiny,
  linenos
}

\usepackage{listings}
\usepackage{xcolor}

\definecolor{codegray}{rgb}{0.5,0.5,0.5}
\definecolor{codepurple}{rgb}{0.58,0,0.82}
\definecolor{backcolour}{rgb}{0.95,0.95,0.92}

\usepackage{hyperref}
\usepackage{url}
\usepackage{graphicx}
\usepackage{colortbl}
\usepackage{booktabs}
\usepackage{multirow}
\usepackage{enumitem}
\usepackage[capitalise]{cleveref}
\definecolor{color5}{HTML}{006795}
\hypersetup{
  colorlinks   = true, 
  urlcolor     = blue, 
  linkcolor    = color5, 
  citecolor   = color5 
}
\usepackage[framemethod=TikZ]{mdframed}
\definecolor{UserExampleBg}{HTML}{ffffff}
\definecolor{UserExampleTitle}{HTML}{618197}
\newmdenv[
    roundcorner=5pt,
    backgroundcolor=UserExampleBg,
    linecolor=UserExampleTitle,
    outerlinewidth=0.5pt,
    frametitlebackgroundcolor=UserExampleTitle,
    frametitlefont={\bfseries\color{white}},
]{user_example}

\AtBeginEnvironment{user_example}

\newcommand{\uam}{\textbf{\texttt{UAM} }}
\newcommand{\uamnospace}{\textbf{\texttt{UAM}}}

\newcommand{\uar}{\textbf{\texttt{UAR} }}
\newcommand{\uarnospace}{\textbf{\texttt{UAR}}}

\newcommand{\auq}{\textbf{\texttt{AUQ} }}
\newcommand{\auqnospace}{\textbf{\texttt{AUQ}}}

\title{Agentic Uncertainty Quantification}

\author{Jiaxin Zhang \quad Prafulla Kumar Choubey \quad  Kung-Hsiang Huang \\
{\bf Caiming Xiong \quad  Chien-Sheng Wu}
\\ Salesforce AI Research \\ \texttt{\{jiaxin.zhang, pchoubey, kh.huang, cxiong, wu.jason\}@salesforce.com}
}

\begin{document}
\maketitle

\begin{abstract}
Although AI agents have demonstrated impressive capabilities in long-horizon reasoning, their reliability is severely hampered by the ``Spiral of Hallucination,'' where early epistemic errors propagate irreversibly. Existing methods face a dilemma: uncertainty quantification (UQ) methods typically act as passive sensors, only diagnosing risks without addressing them, while self-reflection mechanisms suffer from continuous or aimless corrections. To bridge this gap, we propose a unified Dual-Process Agentic UQ (\auqnospace) framework that transforms verbalized uncertainty into active, bi-directional control signals. Our architecture comprises two complementary mechanisms: System 1 (Uncertainty-Aware Memory, \uamnospace), which implicitly propagates verbalized confidence and semantic explanations to prevent blind decision-making; and System 2 (Uncertainty-Aware Reflection, \uarnospace), which utilizes these explanations as rational cues to trigger targeted inference-time resolution only when necessary. This enables the agent to balance efficient execution and deep deliberation dynamically. Extensive experiments on closed-loop benchmarks and open-ended deep research tasks demonstrate that our training-free approach achieves superior performance and trajectory-level calibration. We believe this principled framework \auq represents a significant step towards reliable agents.
\end{abstract}

\section{Introduction}
\label{sec:introduction}
\begin{figure}[t]
    \centering
    \includegraphics[width=\linewidth]{  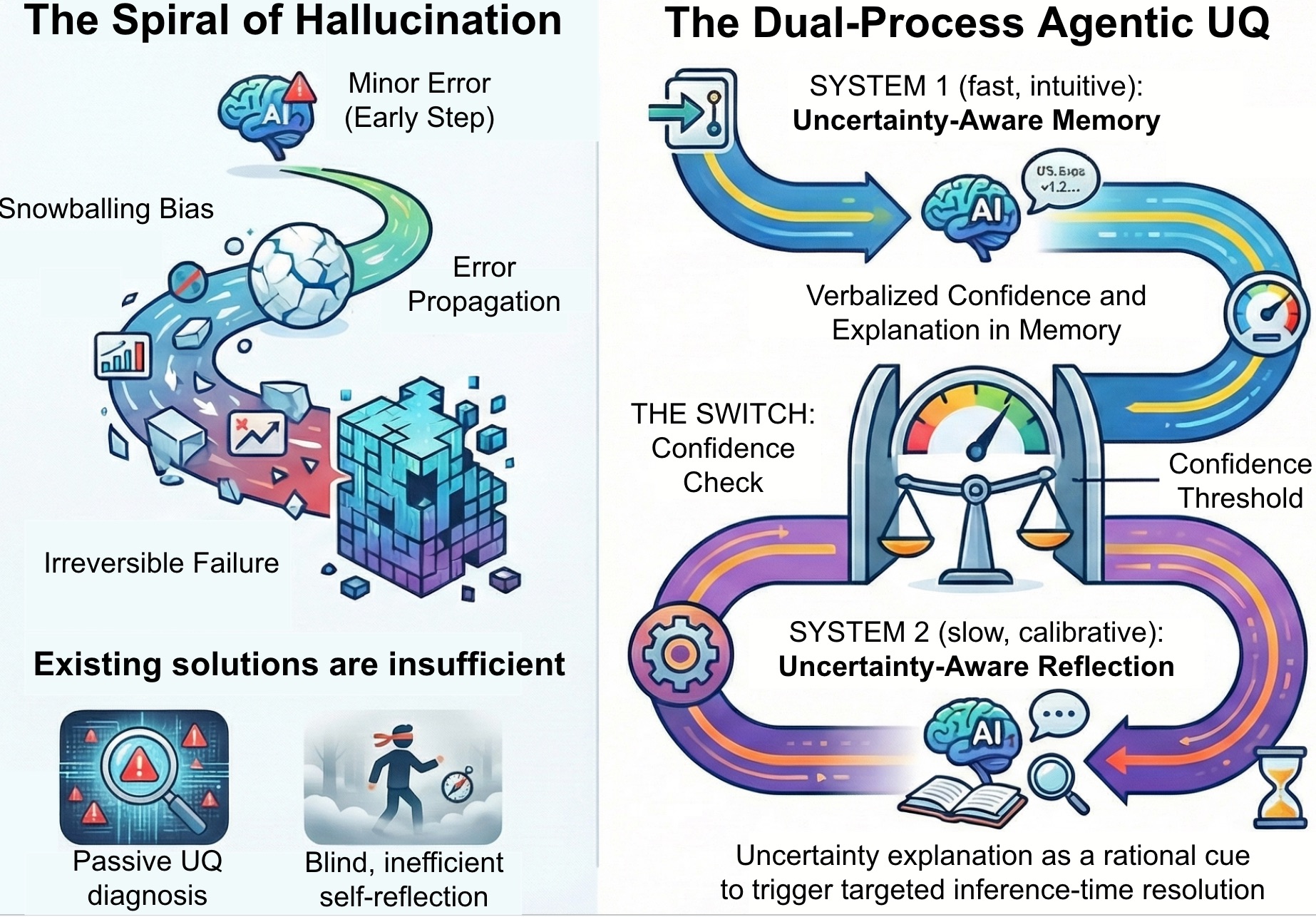}
\caption{Overview of Dual-Process Agentic UQ Framework. To address the spiral of hallucination challenges in long-horizon agents, \auq transforms verbalized uncertainty into active, bi-directional control signals,  comprising two complementary mechanisms: System 1 (Uncertainty-Aware Memory, \uamnospace), which implicitly propagates verbalized confidence and semantic explanations to prevent blind decision-making; and System 2 (Uncertainty-Aware Reflection, \uarnospace), which utilizes these explanations as rational cues to trigger targeted inference-time resolution only when necessary. This allows the agent to balance efficient execution and deep deliberation dynamically.}
    \label{fig:overview}
\end{figure}

The transition from LLMs to autonomous agents has unlocked capabilities in long-horizon reasoning and tool use \citep{yao2022react-auq, schick2023toolformer-auq}. However, as agents operate in dynamic environments, they face a critical reliability bottleneck: the \emph{“Curse of Recursion”} \cite{cemri2025multi-auq}. Unlike single generation, agentic workflows are prone to \textit{“Spiral of Hallucination”} \citep{dziri2023faith-auq,zhang2024language-auq,kalai2025language-auq}, where a minor grounding error in an early reasoning step propagates through the context window, biasing all subsequent planning towards an irreversible failure state \cite{zhangagent-auq}.
To deploy agents in high-risk real-world scenarios, they need not only reasoning capabilities but also the ability to detect when they are deviating from the correct path before errors propagate.

Addressing this fragility of long horizons necessitates a shift from blind execution to conscious control. For an agent to be reliable, it must be aware of its own limitations, and know when it knows and when it does not \cite{kadavath2022language-auq}. Recent work like UProp \citep{duan2025uprop-auq} and SAUP \citep{zhao2025uncertainty-auq} have formally characterized \textit{uncertainty propagation}, mathematically demonstrating how local epistemic errors compound into global failures \cite{zhang2021modern-auq}. While they successfully \textit{quantify} the risk, they do not inherently provide a mechanism to \textit{resolve} it. Conversely, current mitigation strategies rely on \textit{self-reflection} mechanisms \citep{shinn2023reflexion-auq, madaan2023self-auq}. However, without access to ground-truth labels (unavailable in open-ended tasks), standard reflection is often triggered blindly or incessantly, leading to computational inefficiency or the \textit{sycophancy effect} \citep{huanglarge-auq}, where the model hallucinates justifications for its errors \cite{renze2024self-auq}. More discussions can be found in the related work section in Appendix \ref{sec:related_work}.

We believe that reliable long-term intelligent agents need to bridge this gap by transforming uncertainty into bidirectional control signals: \emph{Forward Propagation} for constraint and \emph{Inverse Calibration} for problem-solving. To this end, we propose \textbf{Uncertainty-Aware Memory} (\uamnospace) to explicitly retain verbalized confidence and explanations in the agent's context window. This leverages the Transformer's attention mechanism to naturally suppress overconfidence, creating a soft cognitive constraint that prevents errors from solidifying (Forward). Simultaneously, when suppression signals critical instability, the retained explanation serves as a \textbf{rational cue}, transforming vague epistemic anxiety into targeted diagnosis via \textbf{Uncertainty-Aware Reflection} (\uarnospace) (Inverse). Unlike single-step generation (where the process ends at the output), our agent utilizes this cue to trigger targeted actions, such as switching tools or expanding retrieval, effectively transforming epistemic uncertainty into an {information-search strategy}.


This dual mechanism transforms uncertainty from a passive indicator into an active control signal, implicitly constraining the operation of System 1 (fast, intuitive) using memory and explicitly guiding the operation of System 2 (slow, reflective) using interpretation \cite{kahneman2011thinking-auq, li2025system-auq}. To realize this vision, we propose a unified framework for Agentic UQ based on a dual-process architecture. Our key contributions are threefold:
\vspace{1mm}
\begin{itemize} [leftmargin=10pt,nosep]
    \item We formally decouple Agentic UQ into two complementary mathematical problems: \textit{Forward Uncertainty Propagation} (preventing epistemic errors from solidifying into history) and \textit{Inverse Uncertainty Calibration} (using inference-time compute to correct deviations). To our knowledge, this is the first work to frame agent reliability through this dual-process lens.
    \vspace{1mm}
    \item We introduce a novel dual-process architecture, combining \emph{System 1 (Uncertainty-Aware Memory)} and \emph{System 2 (Uncertainty-Aware Reflection)}, as well as a trajectory-level evaluation metric, tailored for long-horizon reliability calibration. Our approach is a \textit{training-free} framework, which enables the agent to dynamically balance efficient execution with deep deliberation.
    \vspace{1mm}
    \item We validate the framework across diverse domains, spanning embodied decision-making (\texttt{ALFWorld}), web agents (\texttt{WebShop}), and open-ended reasoning (\texttt{DeepResearch}). Our results demonstrate that \auq achieves superior performance, effectively bridging the gap between passive UQ and active agentic control, while yielding superior trajectory-level calibration.

\end{itemize}

\section{Preliminaries}
\label{sec:preliminaries}

\subsection{Long-horizon Agentic Process}
We model the long-horizon agent's system as a Partially Observable Markov Decision Process (POMDP) defined by the tuple $\mathcal{E} = (\mathcal{S}, \mathcal{A}, \Omega, \mathcal{T}, \mathcal{R})$. At step $t$, the agent acts based on a history-dependent policy $\pi(a_t | h_t)$, where $h_t = (o_0, a_0, \dots, o_t)$ represents the observed trajectory. Unlike single-step LLM, where the full context is visible, autonomous agents operate under \emph{state uncertainty} (the true state $s_t$ is latent). The agent maintains an implicit \textit{belief state} $b_t(s_t) = P(s_t | h_t)$.
In this context, reliability failure occurs when the agent's internal belief $b_t$ diverges from the true state $s_t$, while the policy $\pi$ continues to act on this flawed belief. Therefore, the core objective of Agentic UQ is to quantify the uncertainty of $b_t$ and its propagation over trajectories.

\subsection{Decomposing Uncertainty: From Static to Agentic}
Classical UQ distinguishes between aleatoric and epistemic uncertainty \cite{kendall2017uncertainties-auq}, but these definitions undergo a semantic shift, particularly regarding their temporal interaction.

\vspace{1mm}
\noindent \textbf{Aleatoric Uncertainty (Environmental Stochasticity)}: typically refers to inherent randomness of systems and irreducible noise in data. In agentic settings, it maps to the \textit{stochasticity of the environment interface}.  This represents the objective, external friction of the world that the agent cannot eliminate but must robustly handle. 

\vspace{1mm}
\noindent \textbf{Epistemic Uncertainty (Cognitive Deficiency)}:
Traditionally, this stems from a lack of knowledge and ignorance of model parameters. In Agentic Context: it maps to the \textit{agent's reasoning limitations}, manifesting as hallucinations, logic gaps, or memory failures. Crucially, this is reducible via better reasoning strategies or external knowledge.

\vspace{1mm}
\noindent \textbf{The Entanglement in Long Trajectories}: A key insight of our work is that these two uncertainties are \textit{not independent} in agentic UQ. When an agent acts with high epistemic uncertainty at step $t$ and commits it to the history $h_t$, this error becomes part of the ``ground truth'' context for step $t+1$. This phenomenon, which we term the \emph{Spiral of Hallucination}, effectively transforms the agent's internal cognitive error into an external ``environmental'' constraint for future steps. 
This unique dynamics necessitates a dual approach: preventing the propagation of epistemic errors (Forward) and correcting them before they solidify (Inverse).

\subsection{Problem Formulation}
\label{sec:problem_formulation}
We decouple Agentic UQ into two complementary mathematical problems, mapping the intuitive concepts of \textit{Propagation} and \textit{Calibration} to rigorous probabilistic formulations.

\noindent \textbf{Forward Problem: Uncertainty Propagation.} As early errors infect future steps in long-horizon trajectories, it requires the agent to propagate its confidence through time, ensuring that the current uncertainty estimate $c_t$ accounts for the accumulated risk of the history. We thus define the \emph{Forward Problem} as estimating the joint probability of the trajectory's validity up to time $t$. Let $V_t$ be a binary variable indicating if the trajectory $h_t$ is valid (free of critical errors). The goal is to estimate:
\begin{equation}
    P(V_t | h_t) = f_{\text{p}}(P(V_{t-1}|h_{t-1}), \pi(a_t|h_t)) \label{eq:forward}
\end{equation}  
where $f_{\text{p}}(\cdot)$ denotes the prompt formatting function that wraps the history into the textual input. 

\noindent \textbf{Inverse Problem: Uncertainty Calibration.} The inverse problem aims to infer the optimal latent process or action $a^*$ given a desired outcome or constraints.  We define the Inverse Problem as a \textbf{posterior optimization} task. Given a low forward estimate $P(V_t|h_t) < \delta$ (indicating a potential divergence between belief $b_t$ and state $s_t$), the objective is to find a corrected action $a^*$ that maximizes the likelihood of task success, potentially by inferring a latent reasoning path $z$:
\begin{equation}
    a^* = \operatorname*{argmax}_{a} \int P(a | z, h_t) P(z | \text{Succ}, h_t) dz \label{eq:inverse}
\end{equation}
We view this as \emph{Test-Time Calibration} \citep{cobbe2021training-auq}. When the forward process yields low confidence, we treat the generation of a reliable plan as an inverse search problem (see Appendix \ref{app:formal_math}).


\section{Methodology}
\label{sec:methodology}

We propose a unified Dual-Process UQ Framework that dynamically decouples the agent's mechanism into two distinct modes based on uncertainty: a fast, memory-augmented \textbf{System 1} for forward uncertainty propagation, and a slow, reflection-based \textbf{System 2} for inverse uncertainty calibration. 

\vspace{-2mm}
\subsection{System 1: Forward UQ via Uncertainty-Aware Memory (\uamnospace)}
\label{subsec:forward_uam}


The primary objective of System 1 is to maintain \textit{Cognitive Continuity}. It acts as the default ``Fast Path,'' propagating uncertainty constraints implicitly through the agent's context window. Instead of relying on opaque logit probabilities, which are often miscalibrated in RLHF-aligned models \cite{tian2023just-auq}, we implement a \emph{verbalized confidence sensor} by defining an elicitation mapping $\Phi: h_t \to ({a}_t, \hat{c}_t, \hat{e}_t)$, where the agent generates the action ${a}_t$, a confidence scalar $\hat{c}_t \in [0,1]$, and a natural language explanation $\hat{e}_t$. This explanation $\hat{e}_t$ serves to explicitly verbalize latent epistemic uncertainties, see more discussion in Appendix \ref{app:appendix_verbalized_justification}).


We formalize \uam as an augmented history structure that approximates the agent's belief state over time. Unlike standard agents that only retain the trajectory of observations and actions $h_t = \{(o_i, a_i)\}_{i=0}^{t-1}$, our framework explicitly constructs an \textit{Uncertainty-Aware Memory} that retains the agent's metacognitive states:
\begin{equation}
    \mathcal{M}_t = \{ (o_i, a_i, {\hat{c}_i}, {\hat{e}_i}) \}_{i=0}^{t-1}
\end{equation}
Here, $\hat{c}_i$ and $\hat{e}_i$ are the verbalized confidence and semantic explanation generated at step $i$. By preserving $(\hat{c}_i, \hat{e}_i)$, the agent maintains a record of its own reliability, preventing the loss of critical risk signals as the context window slides. 

The core mechanism of \uam is \emph{Semantic Uncertainty Propagation}, denoted by $\pi_\text{fwd}$ policy. While we do not apply hard symbolic rules to block actions, retaining $\hat{e}_t$ in the context window imposes a \emph{Soft Cognitive Constraint} via the Transformer's self-attention mechanism.
Mathematically, the probability of the next action $P(a_{t+1} | \mathcal{M}_t)$ is conditioned on the explicit articulation of prior doubts. When the attention heads attend to tokens in $\hat{e}_i$ describing uncertainty, the generative distribution naturally shifts away from high-commitment actions (Exploitation) towards information-gathering actions (Exploration). This effectively propagates uncertainty forward in time, allowing the agent to dynamically adjust its behavior based on the accumulated uncertainty in its memory. To implement System 1 without fine-tuning, we employ a \textbf{Confidence Elicitation Protocol} by appending a structured instruction to the inference prompt, requiring the agent to output a confidence score $\hat{c}_t \in [0,1]$ and a semantic explanation $\hat{e}_t$ alongside its action 

\vspace{-2mm}
\subsection{System 2: Inverse UQ via Uncertainty-Aware Reflection (\uarnospace)}
\label{subsec:inverse_uar}

System 2 acts as the slow path intervention. It treats reliability as an \textit{Inverse Optimization Problem}: given the detected diagnostic signal $\hat{e}_t$, we aim to infer a corrected action $a^*$ that maximizes information completeness and logic consistency. 

\vspace{1mm}
\noindent \textbf{The Reflection Operator.}
Standard reflection often suffers from degenerate feedback loops \cite{renze2024self-auq}. To enforce effective correction, we utilize the verbalized explanation $\hat{e}_t$ as a \textbf{Rational Cue} from System 1 as an explicit diagnostic constraint. We define a reflection operator $\pi_{\text{inv}}$ that transforms the unconditioned policy into a conditional distribution parameterized by the identified knowledge gaps. For an initial action $a_{init}$ and explanation $\hat{e}_{init}$, the operator constructs a reflection prompt $\mathcal{P}_{ref}(\hat{e}_{init})$ to guide the resampling:
   $ a_{new}, \hat{c}_{new}, \hat{e}_{new} \sim \pi_{\text{inv}} \left(\cdot \mid h_t, a_{init} \right)$.
Crucially, we inject $\hat{e}_{init}$ into $\mathcal{P}_{ref}$  (see Appendix \ref{sec:appendix_prompts_alfworld}), explicitly instructing the model to ``address the concerns mentioned in: $\hat{e}_{init}$''.  This reflection occurs \textit{within} the same time step $t$. If System 2 is triggered, the initial proposal ${a}_{init}$ is discarded, and the selected candidate $a_{new}$ becomes the realized action $a_t$ sent to the environment.

\vspace{1mm}
\noindent \textbf{Consistency-Weighted Reflection.}
We employ a \emph{Best-of-N} sampling strategy to generate $N$ parallel reflection trajectories $\{ (a^{(k)}_{new}, \hat{c}^{(k)}_{new}) \}_{k=1}^N$ to encourage diversity. Instead of simply selecting the maximum confidence, we calculate a consistency-weighted score $S_{cons}(a)$ for each action candidate:
\begin{equation}
    S_{cons}(a) = \frac{1}{N} \sum_{k=1}^N \hat{c}^{(k)}_{new} \cdot \mathbb{I}(a^{(k)}_{new} \equiv a) \label{eq:cw}
\end{equation}
where $\mathbb{I}(\cdot)$ is an indicator function for semantic equivalence, see Appendix \ref{app:implementation} for more explanations. This metric rewards answers that are both \textit{high-confidence} and \textit{consistent}. The provisional optimal action is $a^*_{new} = \arg\max_a S_{cons}(a)$.

\vspace{2mm}
\noindent \textbf{Adaptive Memory Expansion.}
To balance efficiency and context retention, our agent typically operates with a \emph{Limited Memory} window ($\mathcal{M}_{limit} = h_{t-k:t}$) during inference. However, epistemic uncertainty often stems from forgetting \cite{kirsch2024implicit-auq,liu2024lost-auq}. We introduce a \emph{Memory Expansion Mechanism}. After the initial reflection, if the aggregated score $S_{cons}(a^*)$ remains below the reliability threshold $\tau$, the agent triggers a \textit{Context Retrieval} operation, loading the \emph{Full Memory} $\mathcal{M}_{full} = h_{0:t}$ and re-executing the reflection. 
This mechanism ensures that expensive long-context processing is only utilized when local reasoning fails to resolve uncertainty, effectively creating a tiered defense against error propagation. More details are provided in Appendix \ref{sec:appendix_prompts_alfworld}.

\vspace{-2mm}
\subsection{Dual-Process Agentic UQ}
\label{sec:dual_process_control}
Having defined the two systems, we now formalize the \textbf{Dual-Process Policy} $\pi_{\text{dual}}$ that integrates them. To balance reliability and efficiency, we introduce a switching function $S(h_t)$ that governs the transition between systems:
\begin{equation*}
    \pi_{\text{dual}}(a|h_t) = 
    \begin{cases} 
    \pi_{\text{fwd}}(a|h_t, \mathcal{M}_t), & \text{if } S(h_t) = 0 \\
    \pi_{\text{inv}}(a|h_t), & \text{if } S(h_t) = 1 
    \end{cases}
\end{equation*}
where $S(h_t) = \mathbb{I}(\hat{c}_t < \tau)$ is an indicator function triggered when the agent's self-evaluated confidence $\hat{c}_t$ falls below a reliability threshold $\tau$, which is a threshold determined empirically (typically $\tau \in [0.8, 1)$ based on validation datasets). When $\hat{c}_t \ge \tau$, the agent executes the System 1 action directly, incurring minimal cost. When $\hat{c}_t < \tau$, the agent activates the System 2 loop. The corrected result is then written back to $\mathcal{M}_t$ with its updated confidence, ensuring that the resolved uncertainty is propagated to future steps.  This mechanism realizes \emph{dynamic inference budgeting}, allocating expensive reflection only to high-uncertainty steps. The complete execution flow is detailed in Algorithm~\ref{alg:dual_process}.

\begin{table*}[!h]
\centering
\small
\resizebox{\textwidth}{!}{
\begin{tabular}{ll | cc | cc | cc}
\toprule
\multirow{2}{*}{\textbf{Dataset}} & \multirow{2}{*}{\textbf{Method}} & \multicolumn{2}{c|}{\textbf{End-State Calibration} ($\Phi_{\text{last}}$)} & \multicolumn{2}{c|}{\textbf{Overall Quality} ($\Phi_{\text{avg}}$)} & \multicolumn{2}{c}{\textbf{Process Reliability} ($\Phi_{\text{min}}$)} \\
\cmidrule(lr){3-4} \cmidrule(lr){5-6} \cmidrule(lr){7-8}
& & \textbf{ECE} ($\downarrow$) & \textbf{Brier Score} ($\downarrow$) & \textbf{ECE} ($\downarrow$) & \textbf{Brier Score} ($\downarrow$) & \textbf{ECE} ($\downarrow$) & \textbf{Brier Score} ($\downarrow$) \\
\midrule
\multirow{6}{*}{\textbf{ALFWorld}} 
& ReAct & 0.306 & 0.258 & 0.286 & 0.272 & 0.255 & 0.236 \\
& Reflexion & 0.279 & 0.237 & 0.234 & 0.264 & 0.199 & 0.219 \\
& Self-Reflection & 0.264 & 0.227 & 0.223 & 0.254 & 0.185 & 0.218 \\
& CoT-SC & 0.185 & 0.224 & 0.177 & 0.215 & 0.178 & 0.206 \\
\cmidrule{2-8}
\rowcolor{gray!10} & Forward UQ (\uamnospace) & \textbf{0.109} & 0.220 & \textbf{0.160} & 0.222 & 0.104 & 0.217 \\
\rowcolor{gray!10} & Inverse UQ (\uarnospace) & 0.205 & \textbf{0.202} & 0.207 & 0.205 & 0.131 & 0.191 \\
\rowcolor{gray!10}  & \textbf{Dual-Process} (\auqnospace) & {0.174} & {0.218} & {0.162} & \textbf{0.198} & \textbf{0.093} & \textbf{0.176} \\
\midrule
\midrule
\multirow{6}{*}{\textbf{WebShop}} 
& ReAct & 0.335 & 0.288 & 0.342 & 0.305 & 0.329 & 0.298 \\
& Reflexion & 0.352 & 0.275 & 0.334 & 0.296 & 0.272 & 0.284 \\
& Self-Reflection & 0.310 & 0.278 & 0.295 & 0.288 & 0.298 & 0.276 \\
& CoT-SC & 0.225 & 0.263 & 0.269 & 0.253 & 0.242 & 0.250 \\
\cmidrule{2-8}
\rowcolor{gray!10} & Forward UQ (\uamnospace) & \textbf{0.185} & 0.245 & 0.221 & 0.228 & \textbf{0.138} & 0.268 \\
\rowcolor{gray!10} & Inverse UQ (\uarnospace) & 0.209 & \textbf{0.231} & 0.242 & \textbf{0.222} & 0.262 & 0.248 \\
\rowcolor{gray!10} & \textbf{Dual-Process} (\auqnospace) & {0.188} & {0.242} & \textbf{0.215} & {0.236} & {0.210} & \textbf{0.235} \\

\bottomrule
\end{tabular}
}
\vspace{-2mm}
\caption{{Trajectory-Level Reliability Analysis.} We report Trajectory-ECE (T-ECE, lower is better) and Brier Score (T-BS, lower is better) on ALFWorld and WebShop. \textbf{Forward (UAM)} achieves the best calibration (lowest ECE) by effectively dampening overconfidence, while \textbf{Inverse (UAR)} achieves the best sharpness (lowest BS) by resolving uncertainty through reflection. The Dual-Process framework strikes an optimal balance.}
\label{tab:calibration_main}
\end{table*}

\vspace{-2mm}
\subsection{Trajectory-Level Evaluation Metrics}
Standard calibration metrics (e.g., ECE) are ill-suited for long-horizon agentic tasks because they decouple local confidence from global success. A single failure step often invalidates the entire trajectory, regardless of the confidence of other steps. To address this, we introduce a generalized framework for \textbf{Trajectory-Level Calibration}. We define a trajectory as a sequence of confidence scores $\mathbf{c} = \{\hat{c}_1, \dots, \hat{c}_T\}$. We posit that the effective ``trajectory belief'' $C(\tau)$ is determined by a trajectory confidence aggregation function $\Phi$: $C(\tau) = \Phi(\mathbf{c})$, consisting of three distinct aggregation strategies to capture different dimensions: (1) \textbf{End-State Belief ($\Phi_{\text{last}}$):} $C(\tau) = \hat{c}_T$. This assumes reliability depends solely on the last-step final decision; (2) \textbf{Process Reliability ($\Phi_{\text{min}}$):} $C(\tau) = \min_{t=1}^T \hat{c}_t$. Based on the \textit{Weakest Link Principle} \cite{zhonglaw-auq}, a trajectory is only as reliable as its most uncertain step; and (3) \textbf{Overall Quality ($\Phi_{\text{avg}}$):} $C(\tau) = \frac{1}{T} \sum_{t=1}^T \hat{c}_t$. This represents the average confidence throughout the task. The detailed metrics of trajectory-ECE, Brier Score (BS), and AUROC are described in Appendix \ref{app:metrics}.

\begin{table*}[!h]
    \centering
    \resizebox{0.9\linewidth}{!}{
    \begin{tabular}{l|cccc|cccc}
        \toprule
        \multirow{3}{*}{\textbf{Method}} & \multicolumn{4}{c|}{\textbf{ALFWorld}} & \multicolumn{4}{c}{\textbf{WebShop}} \\
        \cmidrule(lr){2-5} \cmidrule(lr){6-9}
         & \textbf{Success Rate} ($\uparrow$) & \multicolumn{3}{c|}{\textbf{AUROC} ($\uparrow$)} & \textbf{Success Rate} ($\uparrow$) & \multicolumn{3}{c}{\textbf{AUROC} ($\uparrow$)} \\
        \cmidrule(lr){2-5} \cmidrule(lr){6-9}
         & (\%) & $\Phi_{\text{last}}$ & $\Phi_{\text{avg}}$ & $\Phi_{\text{min}}$ & (\%) & $\Phi_{\text{last}}$ & $\Phi_{\text{avg}}$ & $\Phi_{\text{min}}$ \\
        \midrule
        ReAct & 63.6 & 0.913 & 0.783 & 0.667 & 29.3 & 0.863 & 0.758 & 0.608 \\
        Reflexion & 67.9 & 0.925 & 0.820 & 0.721 & 30.7 & 0.840 & 0.742 & 0.630 \\
        Self-Reflection & 66.4 & 0.922 & 0.831 & 0.718 & 33.4 & 0.855 & 0.735 & 0.615 \\
        CoT-SC & 69.5 & 0.948 & 0.847 & 0.763 & 37.1 & 0.861 & 0.769 & 0.663  \\
        \midrule
        \rowcolor{gray!10} Forward UQ (\uamnospace) & 65.7 & 0.963 & 0.841 & \textbf{0.807} & 31.4 & 0.865 & 0.775 & 0.725 \\
        \rowcolor{gray!10} Inverse UQ (\uarnospace) & {72.9} & 0.958 & 0.856 & 0.774 & 38.6 & 0.874 & \textbf{0.791} & 0.710 \\
        \rowcolor{gray!10} \textbf{Dual-Process} (\auqnospace) & \textbf{74.3} & \textbf{0.968} & \textbf{0.905} & {0.791} & \textbf{42.9} & \textbf{0.888} & {0.782} & \textbf{0.755} \\
        \bottomrule
    \end{tabular}
    }    

    \vspace{-2mm}
    \caption{{Performance and Discrimination Analysis.} We compare Success Rate (SR) and Discriminative AUROC. {Dual-Process (\auqnospace)} consistently outperforms strong baselines.}
    \label{tab:performance_main}
\end{table*}

\section{Experiments}
\label{sec:experiments}
\vspace{-2mm}
\subsection{Experimental Setup}
\label{sec:setup}

\paragraph{Datasets and Benchmarks.}
To demonstrate the broad applicability of our framework across distinct agentic regimes, we evaluate on three diverse benchmarks: (1) \textbf{ALFWorld} \citep{shridharalfworld-auq}: A precise embodied decision-making suite. 
(2) \textbf{WebShop} \citep{yao2022webshop-auq}: A realistic e-commerce environment characterized by high observation noise. 
(3) \textbf{DeepResearch Bench} \citep{du2025deepresearch-auq}: A benchmark for open-ended information seeking, consisting of {100 PhD-level research tasks}. 
Detailed task descriptions and evaluation protocols are provided in Appendix~\ref{app:datasets}.

\vspace{-3mm}
\paragraph{Baselines.}
We compare our framework against three categories of methods to isolate specific agentic capabilities: (1) \textbf{ReAct} \citep{yao2022react-auq}: The standard System 1 baseline. The agent reasons and acts in an interleaved manner without explicit self-evaluation or reflection. (2) \textbf{Reflexion} \citep{shinn2023reflexion-auq}: A strong \textit{inter-episode} learning baseline representing System 2 self-correction without uncertainty awareness. (3) \textbf{Self-Reflection} \cite{renze2024self-auq}: An \textit{intra-episode} baseline that triggers reflection on every step (or uses a heuristic), by simply asking the model to validate its action for unguided compute scaling. (4) \textbf{CoT-SC} \citep{wangself-auq}: An ensemble baseline, verifying whether our gains stem purely from sampling diversity rather than targeted reflection. To dissect the Dual-Process architecture, we evaluate three variants of our method: {Forward (\uamnospace-only)}, {Inverse (\uarnospace-only)}, and {Dual-Process (\auqnospace)}. The detailed descriptions are provided in Appendix~\ref {app:baselines}.

\vspace{-2mm}
\paragraph{Models and Implementation Details.}
We conduct a comprehensive evaluation across a spectrum of closed and open-source LLMs, including \textbf{GPT-5.1}, \textbf{GPT-4.1}, and \textbf{GPT-4o}~\citep{achiam2023gpt-auq}; \textbf{Gemini-2.5-Pro} and \textbf{Gemini-2.5-Flash}~\cite{comanici2025gemini-auq}; \textbf{Qwen3-235B}~\citep{yang2025qwen3-auq} and \textbf{DeepSeek-V3.1}~\citep{bi2024deepseek-auq}, representing the frontier of open-source capabilities. 
Our Inverse UQ module employs a \textbf{Best-of-N} strategy ($N=3$ parallel paths). For scenarios with limited memory context ($h=5$), we activate the \textbf{Memory Expansion} mechanism upon reflection failure.  All experiments on ALFWorld and WebShop are capped at a maximum of 50 steps. To demonstrate generalizability, we integrated \auq into the \textit{Enterprise Deep Research (EDR)} framework \cite{prabhakar2025enterprise-auq}. 
See more details in Appendix~\ref{app:implementation}.

\vspace{-2mm}
\paragraph{Evaluation Metrics.}
We adopt an evaluation protocol to assess agent performance across three dimensions: (1) \textbf{Performance:} We report Success Rate (SR) for ALFWorld and WebShop. For the open-ended DeepResearch Bench, we utilize their RACE rubric. We employ {Gemini-2.5-Pro} as an impartial LLM-as-a-Judge to score the generated reports. (2) \textbf{Calibration:} We assess how well the agent's confidence aligns with reality using \textbf{Trajectory-ECE (T-ECE)} and \textbf{Trajectory Brier Score (T-BS)}. (3) \textbf{Discrimination:} We report the (\textbf{AUROC}) to quantify the system's ability to distinguish between success and failure modes purely based on internal confidence signals. 

\subsection{Main Results}
\label{sec:main_results_closed}
\subsubsection{Closed-Loop Decision Making}

\noindent \textbf{Reliability (Calibration)}:
Table~\ref{tab:calibration_main} highlights the calibration dynamics via Trajectory-ECE and BS. \textbf{ReAct} agents exhibit severe miscalibration on the process reliability metric ($\Phi_\text{min}$), confirming that snowballed hallucinations often go undetected by the agent itself. In contrast, our \textbf{Forward (\uamnospace-Only)} variant achieves superior T-ECE. By retaining verbalized uncertainty in memory, \uam effectively aligns the agent's confidence with its actual capabilities. However, while \uam is calibrated, \textbf{Inverse (\uarnospace-Only)} is decisive. It achieves the lowest BS by actively resolving knowledge gaps via reflection, polarizing the belief distribution towards certainty. \auq integrates these strengths, maintaining high calibration while significantly improving resolution compared to the baseline.

\begin{figure}[!h]
    \centering
    \includegraphics[width=\linewidth]{  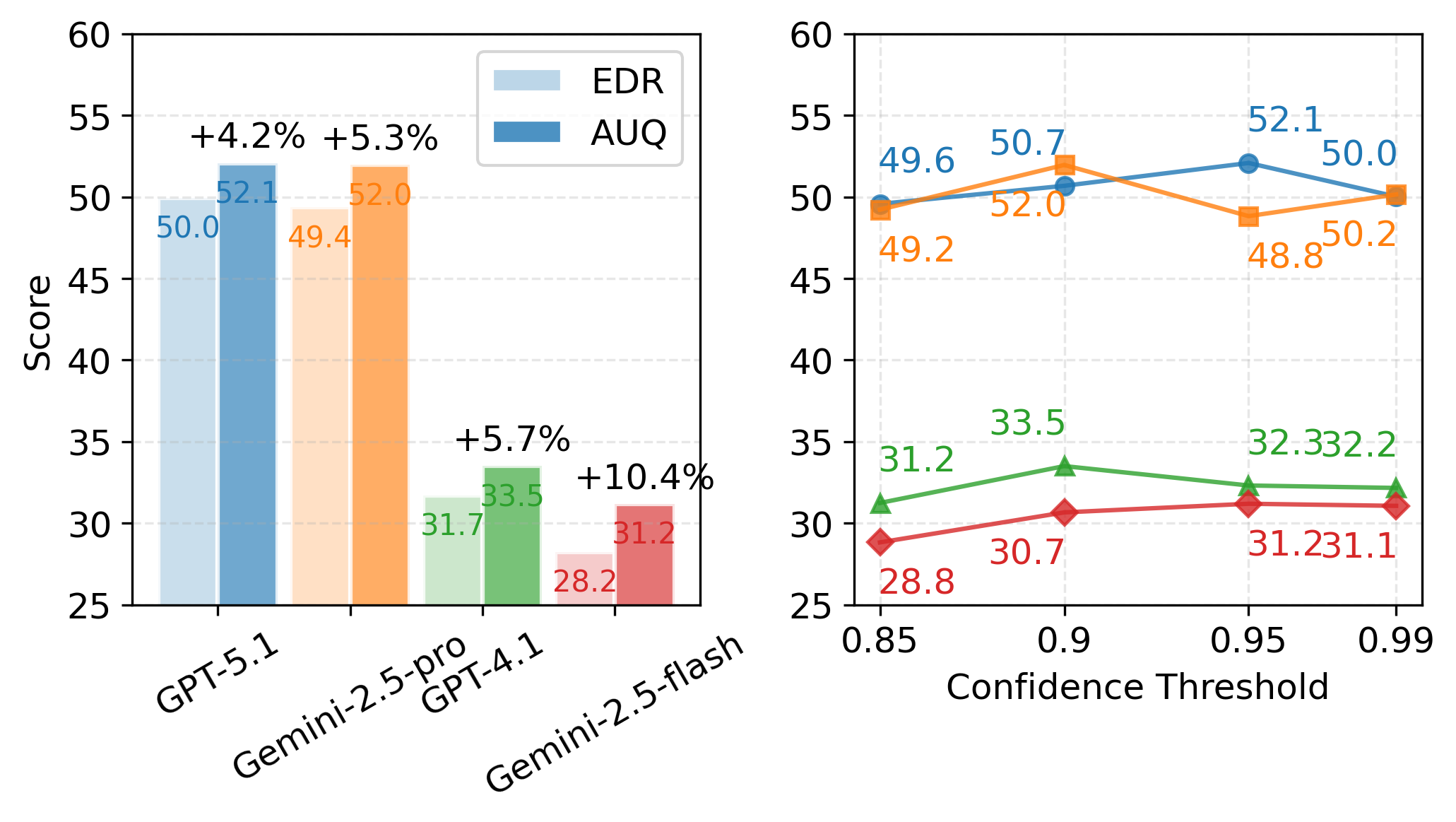} 
    
    \vspace{-3mm}
    \caption{{Generalization on Deep Research.} (Left) \auq outperforms the Enterprise Baseline (EDR) across four diverse LLM backends. \auq consistently outperforms the Enterprise Baseline (EDR). (Right) Sensitivity analysis of confidence threshold $\tau$. The performance is robust across models in the high-confidence regime.}
    \label{fig:deep_research_analysis}
\end{figure}

\noindent \textbf{Utility (Success)}:
Table~\ref{tab:performance_main} demonstrates that superior calibration translates directly into task performance. \textbf{Dual-Process (\auqnospace)} consistently outperforms baselines, achieving a remarkable \textbf{74.3\% SR} (\textbf{+10.7\%}) and  \textbf{42.5\% SR} (\textbf{+13.6\%}) on ALFWorld and WebShop respectively. This gain is particularly significant compared to \textbf{CoT-SC}, indicating that our \textit{Consistency-Weighted} reflection provides quality gains beyond simple ensembling. Furthermore, Dual-Process and its variants achieve superior \textbf{AUROC} against baselines, proving that its internal confidence signals reliably distinguish between success and failure. This discriminative power confirms that the system effectively allocates its System 2 budget to the trajectories that need it most, rather than reflecting blindly or randomly.

\begin{table*}[!h]
    \centering
    \resizebox{0.97\linewidth}{!}{
    \begin{tabular}{l|ccccc}
        \toprule
        \textbf{Agent systems} & \textbf{Overall} & \textbf{Comprehensiveness} & \textbf{Insight} & \textbf{Instruction} & \textbf{Readability} \\
        \midrule
        langchain-open-deep-research & 43.44 & 42.97 & 39.17 & 48.09 & 45.22 \\
        doubao-research & 44.34 & 44.84 & 40.56 & 47.95 & 44.69 \\
        kimi-research & 44.64 & 44.96 & 41.97 & 47.14 & 45.59 \\
        Claude-research & 45.00 & 45.34 & 42.79 & 47.58 & 44.66 \\
        Openai-deepresearch & 46.45 & 46.46 & 43.73 & 49.39 & 47.22 \\
        Gemini-2.5-pro-deepresearch & 49.71 & 49.51 & 49.45 & 50.12 & 50.00 \\
        \midrule
        WebWeaver (Qwen3-235b) & 50.62 & 51.29 & 51.00 & 49.98 & 48.89 \\
        WebWeaver (Claude-sonnet-4) & 50.58 & 51.45 & 50.02 & 50.81 & 49.79 \\
        Enterprise Deep Research (Gemini-2.5-pro) & 50.62 & 49.70 & 51.24 & 50.52 & 50.61 \\
        \midrule
        \rowcolor{gray!10} \auq (Ours, Gemini-2.5-Pro) & 51.97 & 49.19 & 53.64 & \textbf{51.67} & \textbf{51.26} \\
        \rowcolor{gray!10} \auq (Ours, GPT-5.1) & \textbf{52.09} & \textbf{51.60} & \textbf{54.21} & 50.69 & 50.13 \\
        \bottomrule
    \end{tabular}
    }

    \vspace{-2mm}
    \caption{{Evaluation on Open-Ended DeepResearch Bench.} We compare our approach against enterprise-grade systems and recent agent frameworks using the RACE rubric \cite{du2025deepresearch-auq}. The Dual-Process architecture achieves SOTA performance in Comprehensiveness, Insight, Instruction Following, and Readability.}
\label{tab:deepresearch_main}
\end{table*}


\begin{figure*}[!h]
    \centering
    \includegraphics[width=0.41\linewidth]{  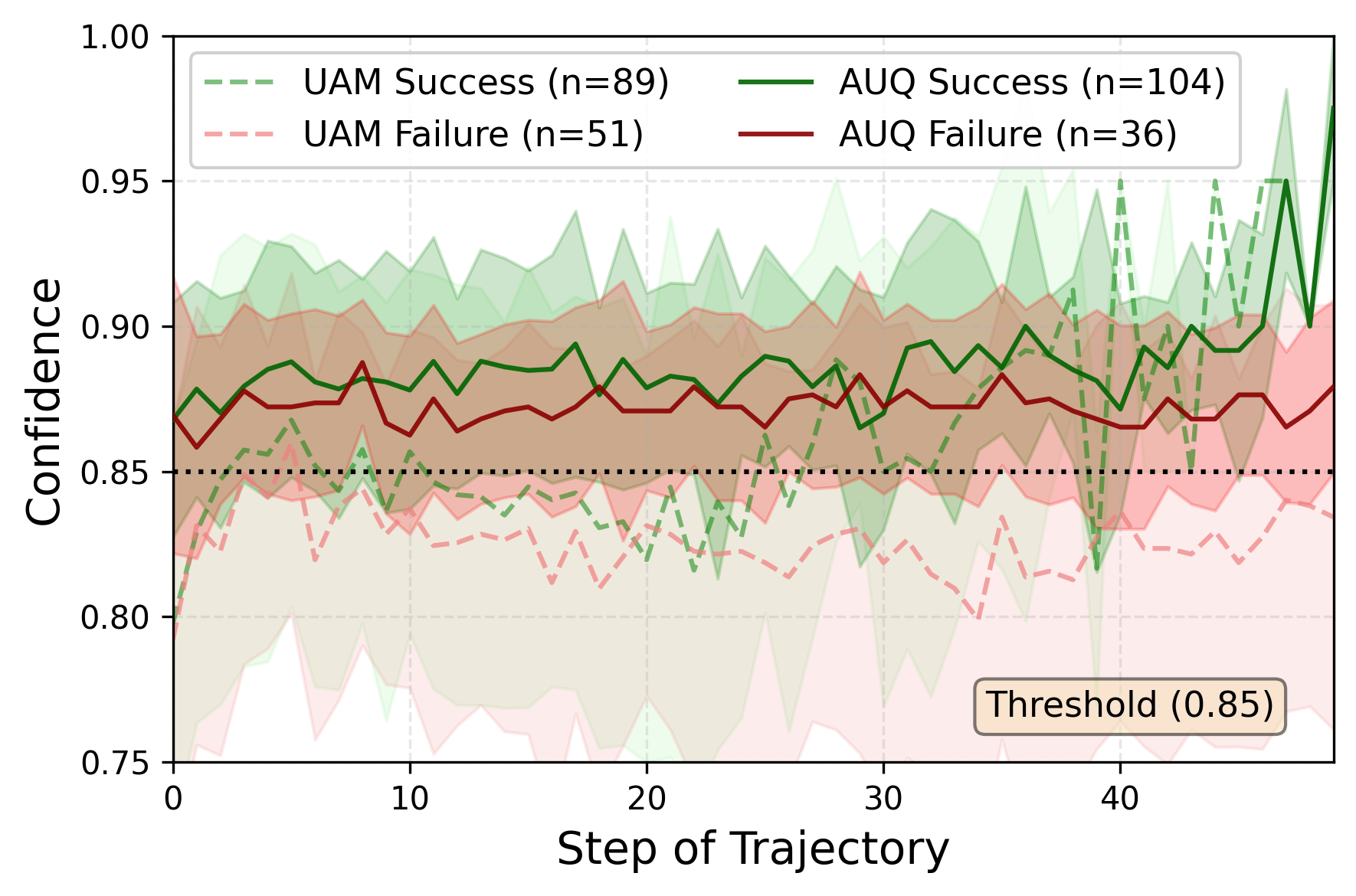}
    \includegraphics[width=0.28\linewidth]{  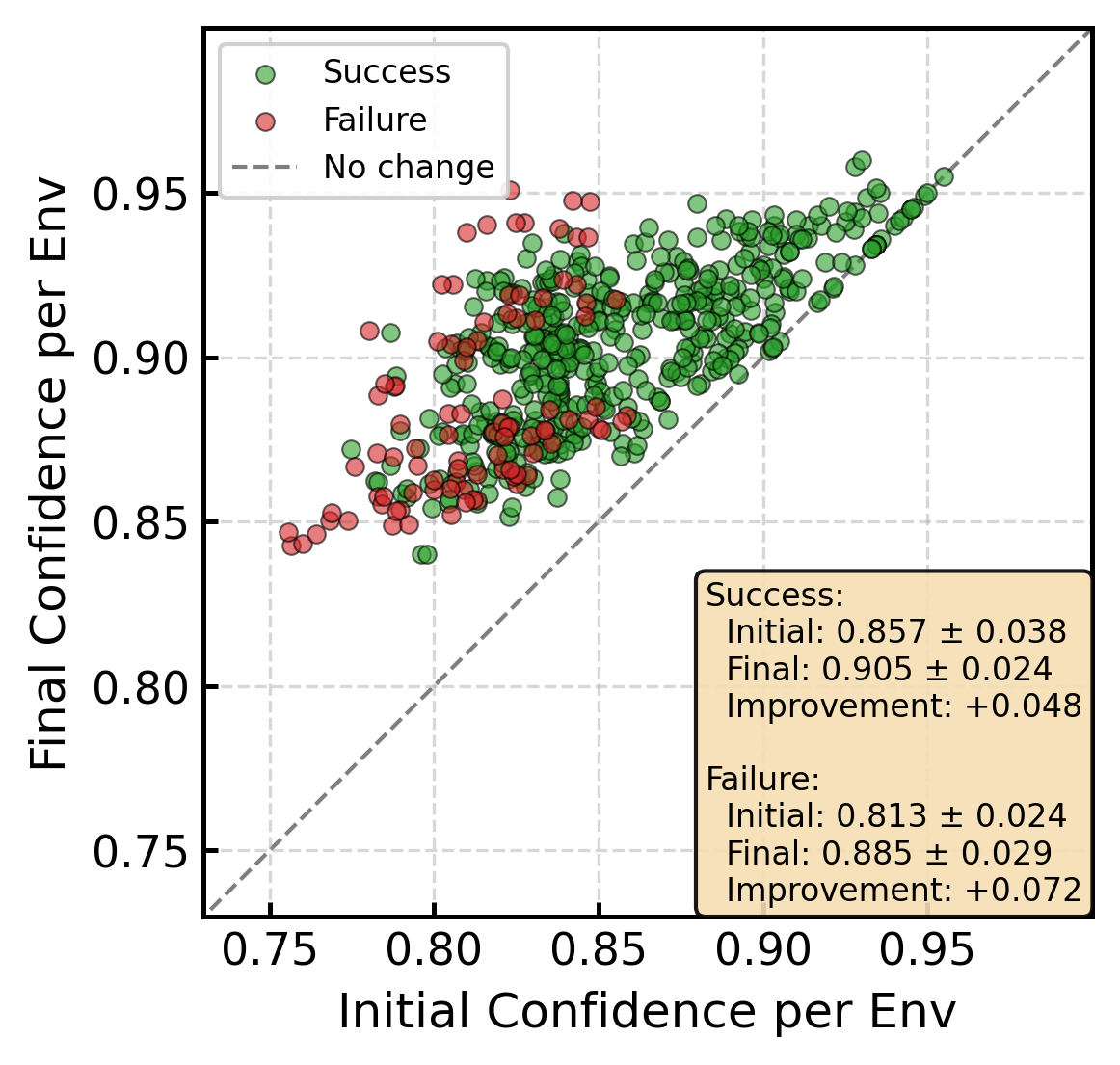}
    \includegraphics[width=0.28\linewidth]{  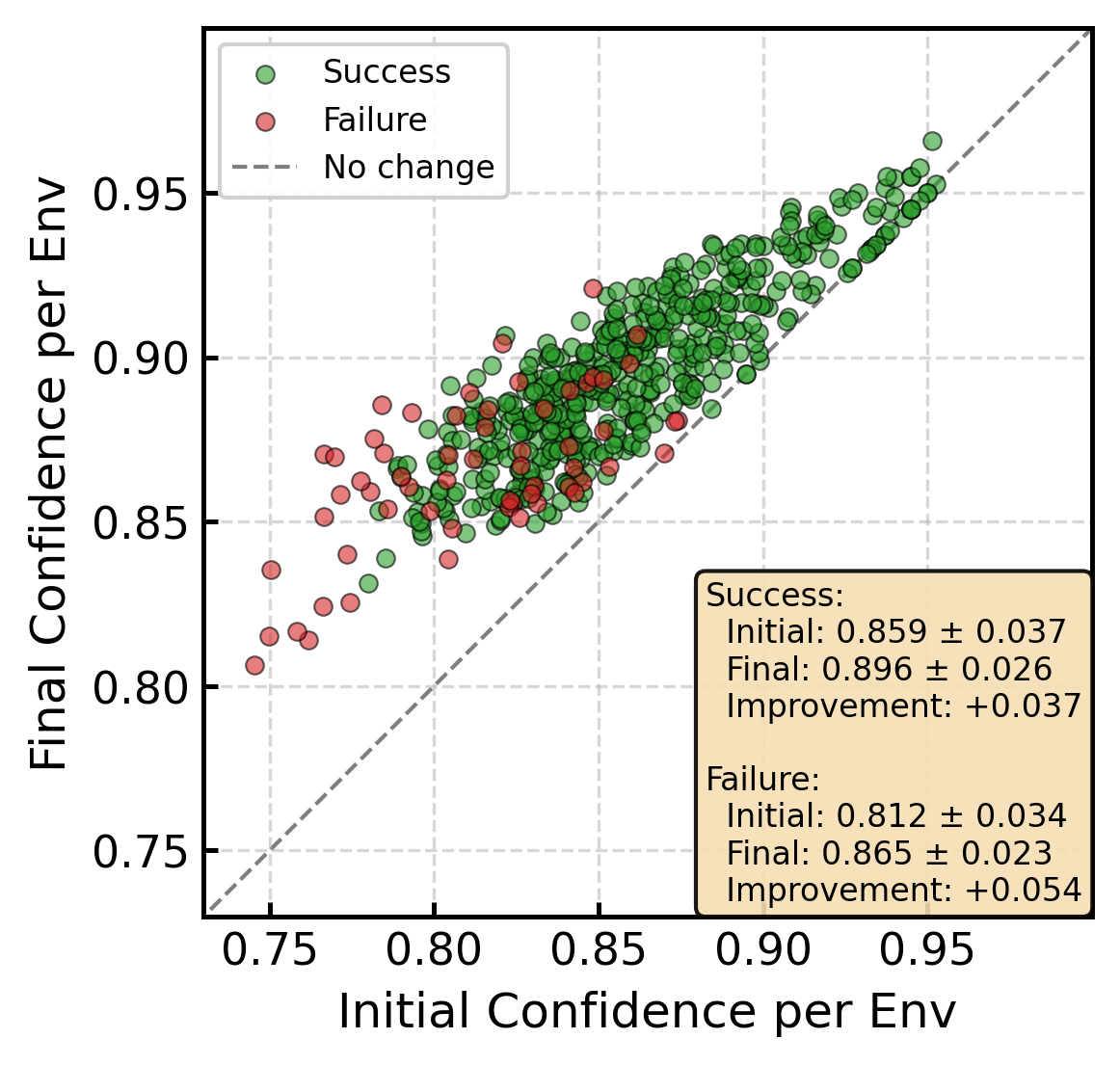}

    \vspace{-3mm}
\caption{{Internal Belief Dynamics on ALFWorld.} (Left) Trajectory confidence evolution comparing \textbf{UAM-Only} (Dashed) and \textbf{Dual-Process} \auq (Solid). (Right) Scatter plots of Pre- vs. Post-Reflection confidence (GPT-4o \& GPT-5.1). Yellow boxes show summary statistics. }
\label{fig:dynamics}
\end{figure*}

\vspace{-2mm}
\subsubsection{Open-Ended Deep Research}
\label{sec:main_results_open}
This task demands high-fidelity reasoning, where success is defined by the depth and coherence of the final report rather than binary completion. Table~\ref{tab:deepresearch_main} benchmarks our \textbf{Dual-Process} \auq framework against widely deployed \emph{Deep Research Agents} and open-source frameworks. \auq achieves a state-of-the-art Overall Score of \textbf{52.09}, outperforming the strongest closed baseline (\textbf{49.71}) and the best open-source competitor (\textbf{50.62}). Critically, this advantage stems from superior \textit{Insight} (\textbf{54.21}) and \textit{Comprehensiveness} (\textbf{51.60}). Baseline agents often exhibit ``information satisficing'', retrieving surface-level facts and stopping. In contrast, our Inverse UQ mechanism triggers a deep dive reflection when it detects epistemic gaps, transforming potential shallow summaries into rigorous, evidence-backed arguments. Figure~\ref{fig:deep_research_analysis} (Left) illustrates \auqnospace's impact across four diverse backends. We observe universal gains, with the relative improvement (e.g., \textbf{+10.4\%} on Gemini-2.5-flash, avg \textbf{+6.4\%} over all models). Figure~\ref{fig:deep_research_analysis} (Right) confirms hyperparameter stability: performance forms a stable plateau for $\tau$, indicating that verbalized confidence is a robust signal that does not require intensive tuning.

\vspace{-1mm}
\noindent \textbf{Qualitative Analysis.}
Beyond quantitative metrics, we analyze the agent's behavioral shifts in Appendix \ref{app:case_study}. Specifically, we provide detailed trajectories for \textbf{Deep Research} (Appendix \ref{app:DR1} and \ref{app:DR2}), illustrating how System 2 refines query decomposition; and \textbf{ALFWorld} (Appendix \ref{app:appendix_alfworld_example}), demonstrating how \auq detects and breaks snowballed hallucination in embodied tasks.


\vspace{-2mm}
\subsection{Analysis and Discussion}
\label{sec:analysis}

\paragraph{Internal Dynamics of Uncertainty.}
\label{sec:analysis_mechanism}

To demystify how \auq alters the reasoning trajectory, we analyze the evolution of belief in Figure~\ref{fig:dynamics}. Comparing the confidence traces of \textbf{\uamnospace-Only} and \textbf{Dual-Process} reveals distinct cognitive roles. \uam trajectories remain consistently lower, suppressing blind commitment to create a crucial \textit{Discriminative Margin} between safe and risky states. In contrast, \auq trajectories show a significant rebound, confirming that System 2 functions actively consume compute to eliminate uncertainty.  However, the scatter plots (Figure~\ref{fig:dynamics} Right) reveal a nuanced \emph{``Delusion Gap''} in this resolution process. While System 2 consistently boosts confidence, failure cases often exhibit significantly larger gains ($\Delta$) than successes. This is because successful trajectories often start with high confidence, where System 2 merely acts as a {validator}; while in intractable failure cases, aggressive reflection can sometimes lead to \textit{Delusional Confirmation}, creating overconfidence in a hallucinated plan. Despite this risk, the net impact remains overwhelmingly positive. As detailed in our comparative analysis (Appendix~\ref{app:comparative_analysis}), our framework corrects 14.3\% of ReAct's failures while regressing on only 3.6\% of its successes. 

\vspace{-3mm}
\begin{figure}[!h]
    \centering
    \includegraphics[width=\linewidth]{  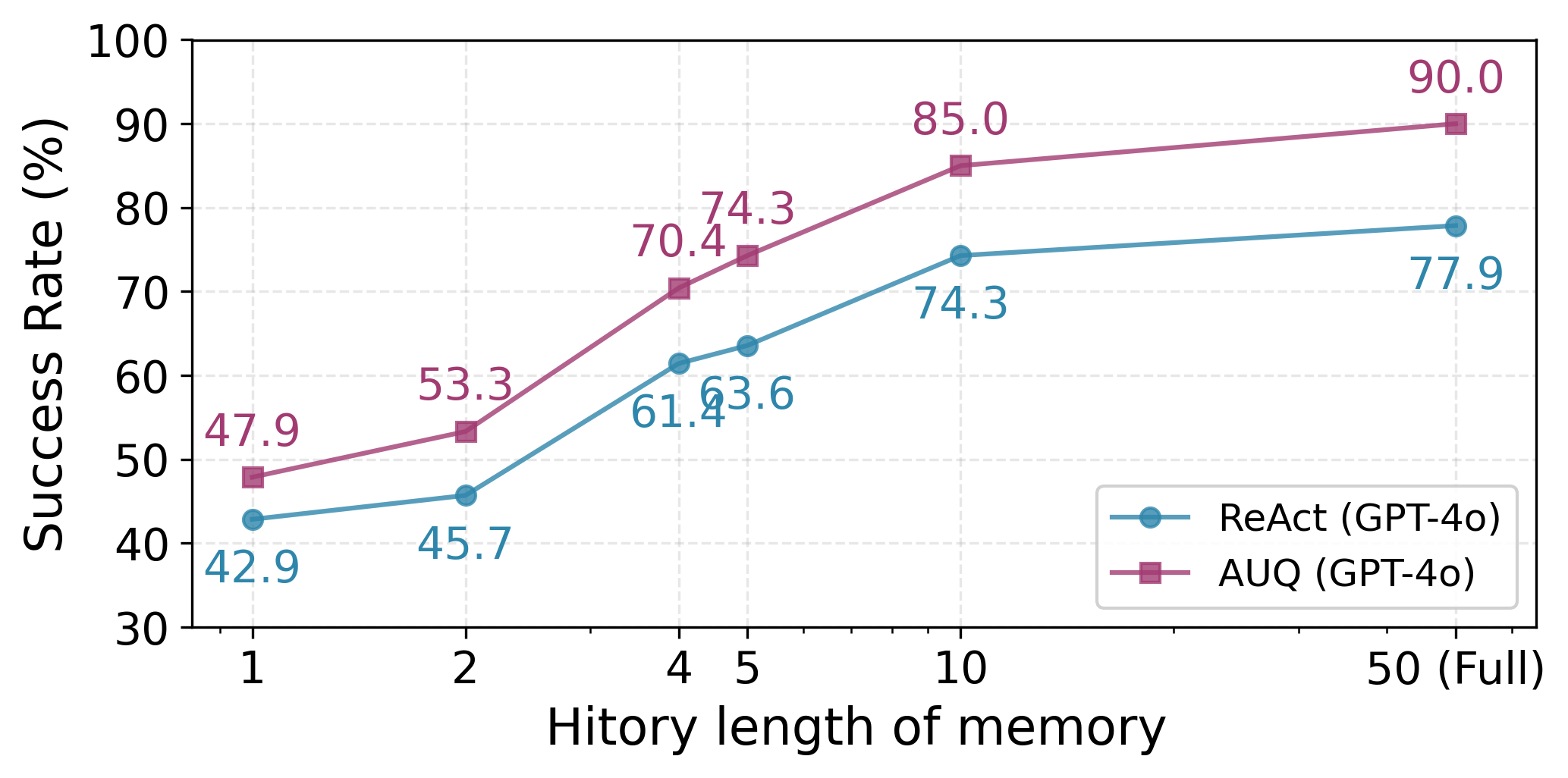}

    \vspace{-3mm}
    \caption{{Effect of memory length on \auq and ReAct.}}
    \label{fig:memory_length}
\end{figure}

\vspace{-4mm}
\paragraph{Role of Memory Length and Model Generalization.}
\label{sec:analysis_memory_model}


Figure~\ref{fig:memory_length} shows the performance degradation of the agent as we shorten the historical window $h$ from the full history to a single step. A significant difference emerges: ReAct's performance drops sharply with limited memory history because it cannot access the previous observations needed to maintain consistency. In contrast, \auq demonstrates superior resistance to forgetting. Even with $h=1$, it still maintains significantly higher performance (+5.0\%), indicating that the stored uncertainty metadata ($\hat{c}, \hat{e}$) encapsulates the risk state of the trajectory into a compact signal that persists even when the original observation log is truncated. Also, the performance improvement provided by \auq increases with the length of the memory.

\vspace{-3mm}
\begin{figure}[h!]
    \centering
    \includegraphics[width=\linewidth]{  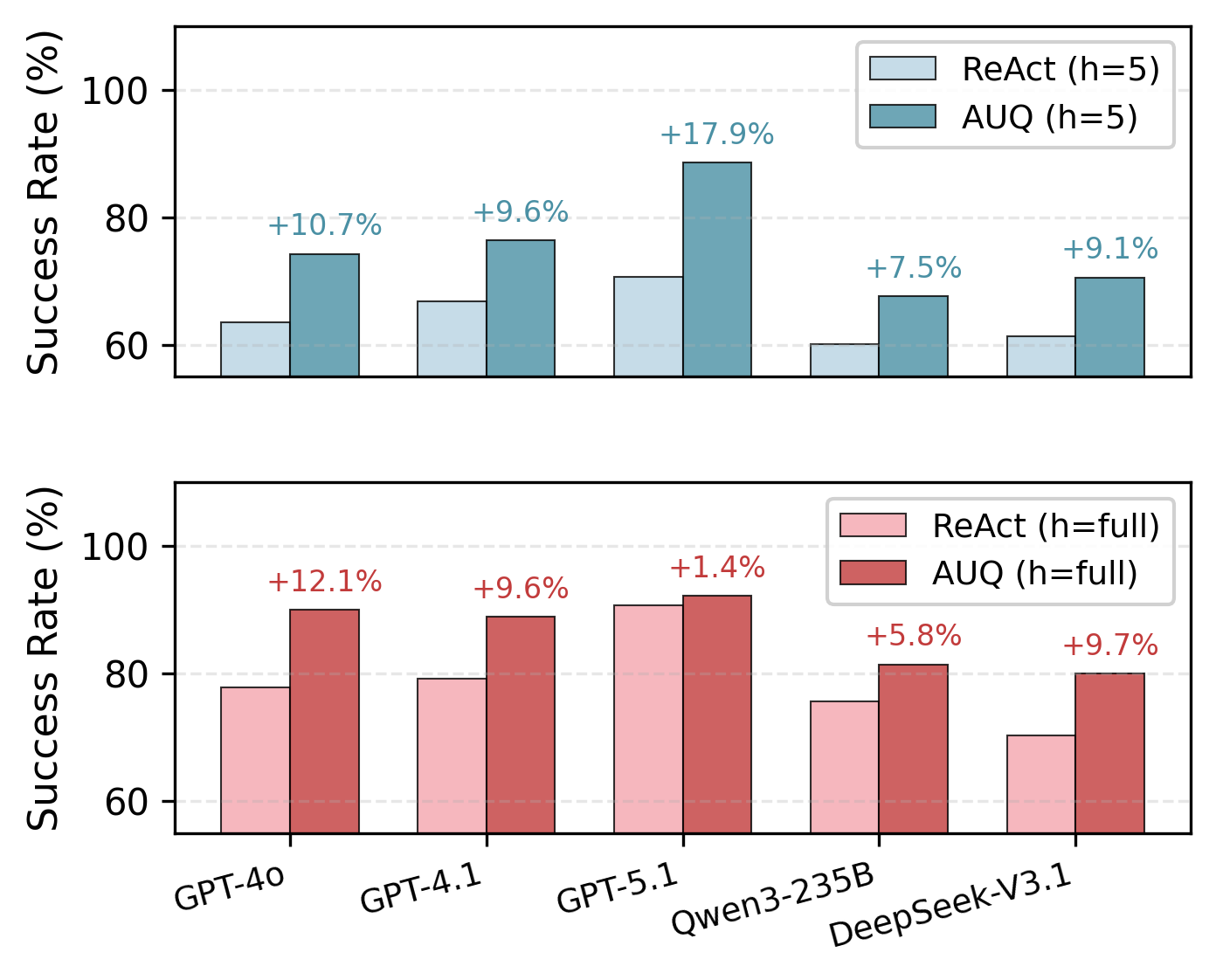}

    \vspace{-5mm}
    \caption{{Model generalization and memory expansion. }}
    \label{fig:model_scaling}
\end{figure}
\vspace{-3mm}

This robustness is further amplified by our memory expansion architecture, validated in Figure~\ref{fig:model_scaling}. In the constrained $h=5$ setting, agents often fail due to missing dependencies. \auq recovers this loss via \textbf{Adaptive Memory Expansion}, triggering full-context retrieval only when System 2 detects epistemic gaps. This mechanism yields massive gains (e.g., \textbf{+17.9\%} for GPT-5.1), proving that dynamic retrieval is superior to static windowing. Crucially, these benefits are universal. As shown in the cross-model comparison, \auq consistently outperforms ReAct across diverse models (e.g., \textbf{+11.0\%} for limited memory, and \textbf{+7.7\%} for full memory). 


\paragraph{Cost-Efficiency Analysis.}
\label{sec:analysis_cost}

\begin{figure}[h!]
    \centering
    \includegraphics[width=0.9\linewidth]{  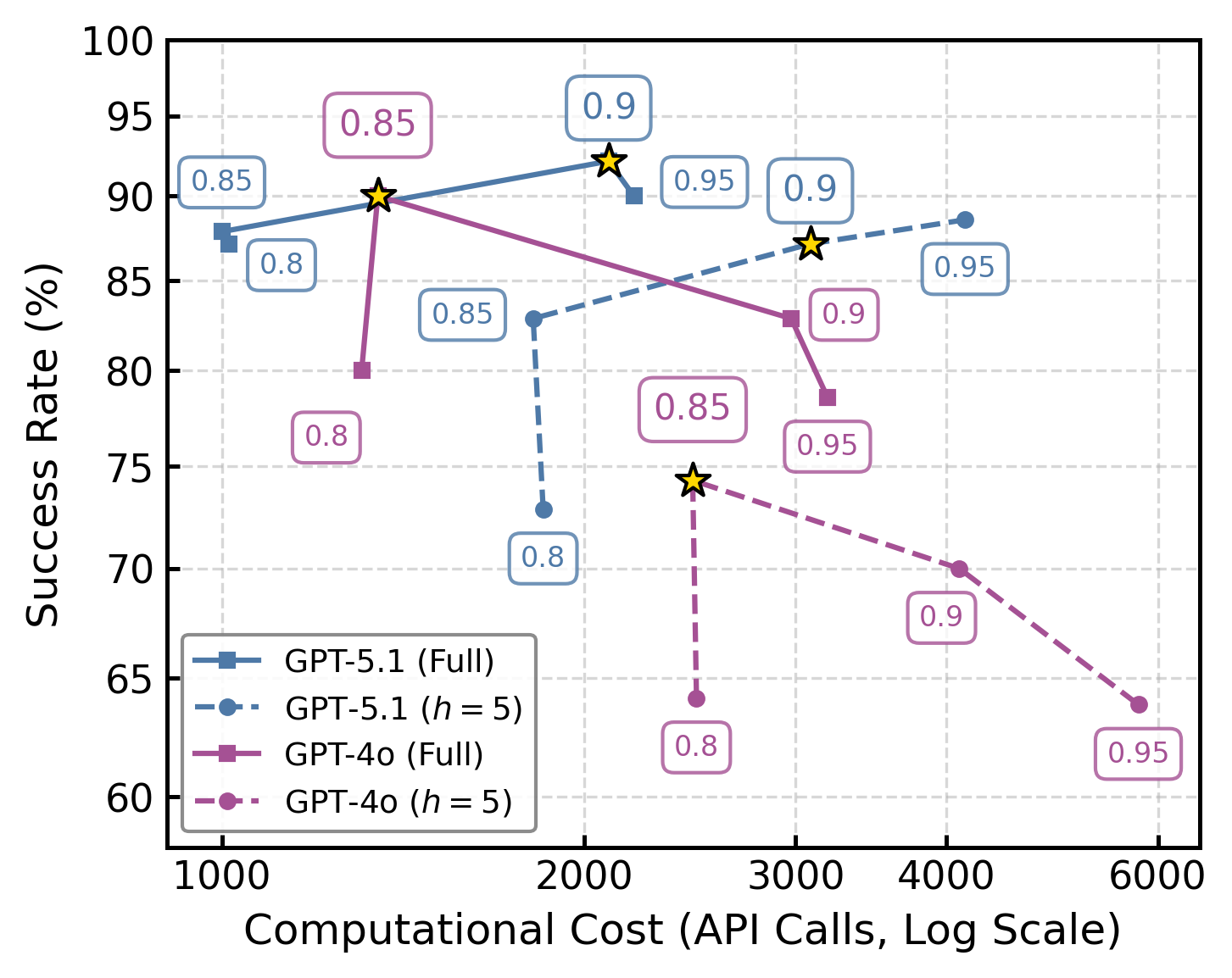}

    \vspace{-3mm}
    \caption{{Pareto Efficiency.} Success Rate vs. Computational Cost (log scale). \auq achieves a superior Pareto frontier, with optimal efficiency at $\tau \approx 0.9$ (starred).}
    \label{fig:pareto_efficiency}
\end{figure}

Figure~\ref{fig:pareto_efficiency} presents the Pareto frontier of Success Rate versus Inference Cost (avg. API calls).  
The curve exhibits a distinct inflection point at $\tau \approx 0.9$, representing the optimal efficiency sweet spot. Further increasing the threshold to $\tau=0.95$ leads to diminishing returns in accuracy while costs increase exponentially, as the agent begins to over-verify even trivial steps. Counterintuitively, our detailed analysis (see Appendix~\ref{app:cost_details}) shows that System 2 does not always increase the total computation; in many cases, by detecting low confidence early, the agent can prevent the lengthy and futile hallucination loops common in ReAct, thus transforming wasted failed computations into useful verification computations. Therefore, \auq effectively shifts the paradigm from cheap but wasteful generation to strategic and efficient reasoning (see Appendix \ref{sec:appendix_cost_discussion}).

\vspace{-3mm}

\vspace{-5mm}
\section{Conclusion}
\label{sec:conclusion}


We propose a dual-process agentic UQ framework that bridges the gap between calibration and autonomous reasoning. By decomposing uncertainty management into fast, memory-aware propagation (System 1) and slow, reflective calibration (System 2), we effectively mitigate the hallucination spiral problem that plagues long-horizon tasks. Our extensive experiments demonstrate that this approach not only achieves superior performance but also excels in calibration and self-awareness. These results suggest that principled agentic UQ is an effective approach to building more reliable and adaptive LLM agents.

\clearpage
\section*{Limitations}

Although our dual-process framework demonstrates significant reliability improvements across various benchmarks, we also acknowledge some limitations that define the scope of our current work and point to future research directions. (1) Our framework is based on the premise that the underlying large language model (LLM) possesses the latent ability to express uncertainty. While we observe a strong correlation between verbally expressed confidence and correctness in strong LLM models (e.g., GPT-5.1, Gemini-2.5-Pro), this capability diminishes in smaller models (e.g., models with fewer than 7 billion parameters). (2) The activation of System 2 (uncertainty-aware reflection) inevitably introduces additional inference latency due to the parallel execution of ``best-of-N''sampling and iterative critique loops. While our analysis shows that this latency is typically offset by a reduction in the total number of wasted steps (thus preventing prolonged failure loops), instantaneous latency spikes may be unacceptable for strictly real-time, low-latency applications (e.g., real-time conversational agents).

\section*{Ethical Considerations}
\label{sec:ethics}

This work introduces a framework for enhancing the reliability of autonomous agents. We identify two primary ethical implications: (1) \emph{Automation Bias and Over-Reliance.} While our AUQ framework improves calibration, there is a risk that users may over-rely on the agent's verbalized confidence ($\hat{c}_t$), perceiving high confidence as a guarantee of factual correctness. In high-stakes domains (e.g., medical or legal research), even calibrated agents can hallucinate. We emphasize that our system serves as a decision-support tool, and human oversight remains essential for final verification. (2) \emph{Computational Impact.} Our System 2 reflection mechanism (Best-of-N sampling) increases inference-time compute, potentially raising energy consumption. However, as noted in our Efficiency Analysis, this is often offset by preventing lengthy, futile failure loops in unguided agents. We advocate for "Adaptive Risk Budgeting" to deploy such compute-intensive reflection only when necessary, minimizing the environmental footprint.

\bibliography{reference}

@inproceedings{yao2022react-auq,
  title={React: Synergizing reasoning and acting in language models},
  author={Yao, Shunyu and Zhao, Jeffrey and Yu, Dian and Du, Nan and Shafran, Izhak and Narasimhan, Karthik R and Cao, Yuan},
  booktitle={The eleventh international conference on learning representations},
  year={2022}
}

@article{schick2023toolformer-auq,
  title={Toolformer: Language models can teach themselves to use tools},
  author={Schick, Timo and Dwivedi-Yu, Jane and Dess{\`\i}, Roberto and Raileanu, Roberta and Lomeli, Maria and Hambro, Eric and Zettlemoyer, Luke and Cancedda, Nicola and Scialom, Thomas},
  journal={Advances in Neural Information Processing Systems},
  volume={36},
  pages={68539--68551},
  year={2023}
}

@article{cemri2025multi-auq,
  title={Why do multi-agent llm systems fail?},
  author={Cemri, Mert and Pan, Melissa Z and Yang, Shuyi and Agrawal, Lakshya A and Chopra, Bhavya and Tiwari, Rishabh and Keutzer, Kurt and Parameswaran, Aditya and Klein, Dan and Ramchandran, Kannan and others},
  journal={arXiv preprint arXiv:2503.13657},
  year={2025}
}

@article{dziri2023faith-auq,
  title={Faith and fate: Limits of transformers on compositionality},
  author={Dziri, Nouha and Lu, Ximing and Sclar, Melanie and Li, Xiang Lorraine and Jiang, Liwei and Lin, Bill Yuchen and Welleck, Sean and West, Peter and Bhagavatula, Chandra and Le Bras, Ronan and others},
  journal={Advances in Neural Information Processing Systems},
  volume={36},
  pages={70293--70332},
  year={2023}
}

@inproceedings{zhang2024language-auq,
  title={How language model hallucinations can snowball},
  author={Zhang, Muru and Press, Ofir and Merrill, William and Liu, Alisa and Smith, Noah A},
  booktitle={Proceedings of the 41st International Conference on Machine Learning},
  pages={59670--59684},
  year={2024}
}

@article{kalai2025language-auq,
  title={Why language models hallucinate},
  author={Kalai, Adam Tauman and Nachum, Ofir and Vempala, Santosh S and Zhang, Edwin},
  journal={arXiv preprint arXiv:2509.04664},
  year={2025}
}

@inproceedings{zhangagent-auq,
  title={Which Agent Causes Task Failures and When? On Automated Failure Attribution of LLM Multi-Agent Systems},
  author={Zhang, Shaokun and Yin, Ming and Zhang, Jieyu and Liu, Jiale and Han, Zhiguang and Zhang, Jingyang and Li, Beibin and Wang, Chi and Wang, Huazheng and Chen, Yiran and others},
  booktitle={Forty-second International Conference on Machine Learning},
  year={2025}
}

@article{kadavath2022language-auq,
  title={Language models (mostly) know what they know},
  author={Kadavath, Saurav and Conerly, Tom and Askell, Amanda and Henighan, Tom and Drain, Dawn and Perez, Ethan and Schiefer, Nicholas and Hatfield-Dodds, Zac and DasSarma, Nova and Tran-Johnson, Eli and others},
  journal={arXiv preprint arXiv:2207.05221},
  year={2022}
}

@article{duan2025uprop-auq,
  title={UProp: Investigating the Uncertainty Propagation of LLMs in Multi-Step Agentic Decision-Making},
  author={Duan, Jinhao and Diffenderfer, James and Madireddy, Sandeep and Chen, Tianlong and Kailkhura, Bhavya and Xu, Kaidi},
  journal={arXiv preprint arXiv:2506.17419},
  year={2025}
}

@inproceedings{zhao2025uncertainty-auq,
  title={Uncertainty propagation on llm agent},
  author={Zhao, Qiwei and Li, Dong and Liu, Yanchi and Cheng, Wei and Sun, Yiyou and Oishi, Mika and Osaki, Takao and Matsuda, Katsushi and Yao, Huaxiu and Zhao, Chen and others},
  booktitle={Proceedings of the 63rd Annual Meeting of the Association for Computational Linguistics (Volume 1: Long Papers)},
  pages={6064--6073},
  year={2025}
}

@article{zhang2021modern-auq,
  title={Modern Monte Carlo methods for efficient uncertainty quantification and propagation: A survey},
  author={Zhang, Jiaxin},
  journal={Wiley Interdisciplinary Reviews: Computational Statistics},
  volume={13},
  number={5},
  pages={e1539},
  year={2021},
  publisher={Wiley Online Library}
}

@article{shinn2023reflexion-auq,
  title={Reflexion: Language agents with verbal reinforcement learning},
  author={Shinn, Noah and Cassano, Federico and Gopinath, Ashwin and Narasimhan, Karthik and Yao, Shunyu},
  journal={Advances in Neural Information Processing Systems},
  volume={36},
  pages={8634--8652},
  year={2023}
}

@article{madaan2023self-auq,
  title={Self-refine: Iterative refinement with self-feedback},
  author={Madaan, Aman and Tandon, Niket and Gupta, Prakhar and Hallinan, Skyler and Gao, Luyu and Wiegreffe, Sarah and Alon, Uri and Dziri, Nouha and Prabhumoye, Shrimai and Yang, Yiming and others},
  journal={Advances in Neural Information Processing Systems},
  volume={36},
  pages={46534--46594},
  year={2023}
}

@inproceedings{huanglarge-auq,
  title={Large Language Models Cannot Self-Correct Reasoning Yet},
  author={Huang, Jie and Chen, Xinyun and Mishra, Swaroop and Zheng, Huaixiu Steven and Yu, Adams Wei and Song, Xinying and Zhou, Denny},
  booktitle={The Twelfth International Conference on Learning Representations},
  year={2023}
}

@article{renze2024self-auq,
  title={Self-reflection in llm agents: Effects on problem-solving performance},
  author={Renze, Matthew and Guven, Erhan},
  journal={arXiv preprint arXiv:2405.06682},
  year={2024}
}

@book{kahneman2011thinking-auq,
  title={Thinking, Fast and Slow},
  author={Kahneman, Daniel},
  year={2011},
  publisher={Farrar, Straus and Giroux}
}

@article{li2025system-auq,
  title={From system 1 to system 2: A survey of reasoning large language models},
  author={Li, Zhong-Zhi and Zhang, Duzhen and Zhang, Ming-Liang and Zhang, Jiaxin and Liu, Zengyan and Yao, Yuxuan and Xu, Haotian and Zheng, Junhao and Wang, Pei-Jie and Chen, Xiuyi and others},
  journal={arXiv preprint arXiv:2502.17419},
  year={2025}
}

@article{kendall2017uncertainties-auq,
  title={What uncertainties do we need in bayesian deep learning for computer vision?},
  author={Kendall, Alex and Gal, Yarin},
  journal={Advances in neural information processing systems},
  volume={30},
  year={2017}
}

@article{cobbe2021training-auq,
  title={Training verifiers to solve math word problems},
  author={Cobbe, Karl and Kosaraju, Vineet and Bavarian, Mohammad and Chen, Mark and Jun, Heewoo and Kaiser, Lukasz and Plappert, Matthias and Tworek, Jerry and Hilton, Jacob and Nakano, Reiichiro and others},
  journal={arXiv preprint arXiv:2110.14168},
  year={2021}
}

@inproceedings{tian2023just-auq,
  title={Just Ask for Calibration: Strategies for Eliciting Calibrated Confidence Scores from Language Models Fine-Tuned with Human Feedback},
  author={Tian, Katherine and Mitchell, Eric and Zhou, Allan and Sharma, Archit and Rafailov, Rafael and Yao, Huaxiu and Finn, Chelsea and Manning, Christopher D},
  booktitle={Proceedings of the 2023 Conference on Empirical Methods in Natural Language Processing},
  pages={5433--5442},
  year={2023}
}

@inproceedings{zhonglaw-auq,
  title={Law of the Weakest Link: Cross Capabilities of Large Language Models},
  author={Zhong, Ming and Zhang, Aston and Wang, Xuewei and Hou, Rui and Xiong, Wenhan and Zhu, Chenguang and Chen, Zhengxing and Tan, Liang and Bi, Chloe and Lewis, Mike and others},
  booktitle={The Thirteenth International Conference on Learning Representations},
  year={2024}
}

@inproceedings{shridharalfworld-auq,
  title={ALFWorld: Aligning Text and Embodied Environments for Interactive Learning},
  author={Shridhar, Mohit and Yuan, Xingdi and Cote, Marc-Alexandre and Bisk, Yonatan and Trischler, Adam and Hausknecht, Matthew},
  booktitle={International Conference on Learning Representations},
  year={2021}
}

@article{yao2022webshop-auq,
  title={Webshop: Towards scalable real-world web interaction with grounded language agents},
  author={Yao, Shunyu and Chen, Howard and Yang, John and Narasimhan, Karthik},
  journal={Advances in Neural Information Processing Systems},
  volume={35},
  pages={20744--20757},
  year={2022}
}

@article{du2025deepresearch-auq,
  title={DeepResearch Bench: A Comprehensive Benchmark for Deep Research Agents},
  author={Du, Mingxuan and Xu, Benfeng and Zhu, Chiwei and Wang, Xiaorui and Mao, Zhendong},
  journal={arXiv preprint arXiv:2506.11763},
  year={2025}
}

@inproceedings{wangself-auq,
  title={Self-Consistency Improves Chain of Thought Reasoning in Language Models},
  author={Wang, Xuezhi and Wei, Jason and Schuurmans, Dale and Le, Quoc V and Chi, Ed H and Narang, Sharan and Chowdhery, Aakanksha and Zhou, Denny},
  booktitle={The Eleventh International Conference on Learning Representations},
  year={2022}
}

@article{comanici2025gemini-auq,
  title={Gemini 2.5: Pushing the frontier with advanced reasoning, multimodality, long context, and next generation agentic capabilities},
  author={Comanici, Gheorghe and Bieber, Eric and Schaekermann, Mike and Pasupat, Ice and Sachdeva, Noveen and Dhillon, Inderjit and Blistein, Marcel and Ram, Ori and Zhang, Dan and Rosen, Evan and others},
  journal={arXiv preprint arXiv:2507.06261},
  year={2025}
}

@article{yang2025qwen3-auq,
  title={Qwen3 technical report},
  author={Yang, An and Li, Anfeng and Yang, Baosong and Zhang, Beichen and Hui, Binyuan and Zheng, Bo and Yu, Bowen and Gao, Chang and Huang, Chengen and Lv, Chenxu and others},
  journal={arXiv preprint arXiv:2505.09388},
  year={2025}
}

@article{bi2024deepseek-auq,
  title={Deepseek llm: Scaling open-source language models with longtermism},
  author={Bi, Xiao and Chen, Deli and Chen, Guanting and Chen, Shanhuang and Dai, Damai and Deng, Chengqi and Ding, Honghui and Dong, Kai and Du, Qiushi and Fu, Zhe and others},
  journal={arXiv preprint arXiv:2401.02954},
  year={2024}
}

@article{prabhakar2025enterprise-auq,
  title={Enterprise Deep Research: Steerable Multi-Agent Deep Research for Enterprise Analytics},
  author={Prabhakar, Akshara and Ram, Roshan and Chen, Zixiang and Savarese, Silvio and Wang, Frank and Xiong, Caiming and Wang, Huan and Yao, Weiran},
  journal={arXiv preprint arXiv:2510.17797},
  year={2025}
}

@inproceedings{guo2017calibration-auq,
  title={On calibration of modern neural networks},
  author={Guo, Chuan and Pleiss, Geoff and Sun, Yu and Weinberger, Kilian Q},
  booktitle={International conference on machine learning},
  pages={1321--1330},
  year={2017},
  organization={PMLR}
}

@inproceedings{kuhnsemantic-auq,
  title={Semantic Uncertainty: Linguistic Invariances for Uncertainty Estimation in Natural Language Generation},
  author={Kuhn, Lorenz and Gal, Yarin and Farquhar, Sebastian},
  booktitle={The Eleventh International Conference on Learning Representations},
  year={2023}
}

@article{linteaching-auq,
  title={Teaching Models to Express Their Uncertainty in Words},
  author={Lin, Stephanie and Hilton, Jacob and Evans, Owain},
  journal={Transactions on Machine Learning Research},
  year={2022},
}

@article{groot2024overconfidence-auq,
  title={Overconfidence is key: Verbalized uncertainty evaluation in large language and vision-language models},
  author={Groot, Tobias and Valdenegro-Toro, Matias},
  journal={arXiv preprint arXiv:2405.02917},
  year={2024}
}

@inproceedings{kirchhofposition-auq,
  title={Position: Uncertainty Quantification Needs Reassessment for Large Language Model Agents},
  author={Kirchhof, Michael and Kasneci, Gjergji and Kasneci, Enkelejda},
  booktitle={Forty-second International Conference on Machine Learning Position Paper Track},
  year={2025}
}

@article{zhu2025llm-auq,
  title={Where llm agents fail and how they can learn from failures},
  author={Zhu, Kunlun and Liu, Zijia and Li, Bingxuan and Tian, Muxin and Yang, Yingxuan and Zhang, Jiaxun and Han, Pengrui and Xie, Qipeng and Cui, Fuyang and Zhang, Weijia and others},
  journal={arXiv preprint arXiv:2509.25370},
  year={2025}
}

@inproceedings{han2024towards-auq,
  title={Towards Uncertainty-Aware Language Agent},
  author={Han, Jiuzhou and Buntine, Wray and Shareghi, Ehsan},
  booktitle={Findings of the Association for Computational Linguistics ACL 2024},
  pages={6662--6685},
  year={2024}
}

@article{tsai2024efficient-auq,
  title={Efficient Non-Parametric Uncertainty Quantification for Black-Box Large Language Models and Decision Planning},
  author={Tsai, Yao-Hung Hubert and Talbott, Walter and Zhang, Jian},
  journal={arXiv preprint arXiv:2402.00251},
  year={2024}
}

@inproceedings{liu2024uncertainty-auq,
  title={Uncertainty Calibration for Tool-Using Language Agents},
  author={Liu, Hao and Dou, Zi-Yi and Wang, Yixin and Peng, Nanyun and Yue, Yisong},
  booktitle={Findings of the Association for Computational Linguistics: EMNLP 2024},
  pages={16781--16805},
  year={2024}
}

@article{lymperopoulos2025tools-auq,
  title={Tools in the Loop: Quantifying Uncertainty of LLM Question Answering Systems That Use Tools},
  author={Lymperopoulos, Panagiotis and Sarathy, Vasanth},
  journal={arXiv preprint arXiv:2505.16113},
  year={2025}
}

@inproceedings{goucritic-auq,
  title={CRITIC: Large Language Models Can Self-Correct with Tool-Interactive Critiquing},
  author={Gou, Zhibin and Shao, Zhihong and Gong, Yeyun and Yang, Yujiu and Duan, Nan and Chen, Weizhu and others},
  booktitle={The Twelfth International Conference on Learning Representations},
  year={2023}
}

@article{saunders2022self-auq,
  title={Self-critiquing models for assisting human evaluators},
  author={Saunders, William and Yeh, Catherine and Wu, Jeff and Bills, Steven and Ouyang, Long and Ward, Jonathan and Leike, Jan},
  journal={arXiv preprint arXiv:2206.05802},
  year={2022}
}

@article{wang2025steca-auq,
  title={Steca: Step-level trajectory calibration for llm agent learning},
  author={Wang, Hanlin and Wang, Jian and Leong, Chak Tou and Li, Wenjie},
  journal={arXiv preprint arXiv:2502.14276},
  year={2025}
}

@article{snell2024scaling-auq,
  title={Scaling llm test-time compute optimally can be more effective than scaling model parameters},
  author={Snell, Charlie and Lee, Jaehoon and Xu, Kelvin and Kumar, Aviral},
  journal={arXiv preprint arXiv:2408.03314},
  year={2024}
}

@article{achiam2023gpt-auq,
  title={Gpt-4 technical report},
  author={Achiam, Josh and Adler, Steven and Agarwal, Sandhini and Ahmad, Lama and Akkaya, Ilge and Aleman, Florencia Leoni and Almeida, Diogo and Altenschmidt, Janko and Altman, Sam and Anadkat, Shyamal and others},
  journal={arXiv preprint arXiv:2303.08774},
  year={2023}
}

@article{kirsch2024implicit-auq,
  title={(Implicit) Ensembles of Ensembles: Epistemic Uncertainty Collapse in Large Models},
  author={Kirsch, Andreas},
  journal={arXiv preprint arXiv:2409.02628},
  year={2024}
}

@article{liu2024lost-auq,
  title={Lost in the middle: How language models use long contexts},
  author={Liu, Nelson F and Lin, Kevin and Hewitt, John and Paranjape, Ashwin and Bevilacqua, Michele and Petroni, Fabio and Liang, Percy},
  journal={Transactions of the Association for Computational Linguistics},
  volume={12},
  pages={157--173},
  year={2024}
}

\clearpage

\appendix
\onecolumn
\setcounter{secnumdepth}{4} 
\setcounter{tocdepth}{4}    

\section{Appendix}
\etocsettocstyle{\section*{Appendix Contents}}{} 
\etocsetnexttocdepth{paragraph}
\etocsetlevel{paragraph}{4}

\localtableofcontents

\clearpage
\subsection{Related Work}
\label{sec:related_work}

\paragraph*{Uncertainty Quantification in LLMs.}
\label{sec:related_llm_calibration}

Accurate UQ is the cornerstone of reliable AI deployment. While classical calibration methods like temperature scaling \citep{guo2017calibration-auq} are effective for discriminative tasks, applying them to the open-ended generation of LLMs remains non-trivial \citep{kadavath2022language-auq}. Recent advancements have largely bifurcated into logit-based and linguistic-based approaches. Logit-based methods, such as Semantic Entropy \citep{kuhnsemantic-auq}, attempt to aggregate token probabilities over equivalent meanings, yet they often face accessibility challenges with black-box APIs.  Consequently, \textit{verbalized uncertainty} where models explicitly express their confidence via natural language prompting, has emerged as a promising alternative \citep{linteaching-auq, tian2023just-auq}. Studies show that sufficiently aligned models can produce well-calibrated verbal confidence \citep{groot2024overconfidence-auq}, a finding that serves as the theoretical basis for our \textbf{Forward UQ (System 1)} module. However, these methods predominantly treat UQ as a static, post-hoc metric for isolated queries. They do not address how such signals can be operationalized to dynamically control the branching and backtracking decisions in the continuous, multi-step trajectories of autonomous agents \citep{kirchhofposition-auq}.

\paragraph*{Autonomous Agents and Error Propagation.}
\label{sec:related_error_propagation}

While the reasoning-acting paradigm, pioneered by frameworks like ReAct \citep{yao2022react-auq} and Reflexion \citep{shinn2023reflexion-auq}, has empowered LLMs to solve long-horizon tasks, these systems exhibit significant brittleness in dynamic environments. Unlike single-turn generation, where errors are isolated, autonomous agents suffer from the “Curse of Recursion” or “Error Propagation”. Recent failure analyses have systematically codified this phenomenon. For instance, \citet{zhu2025llm-auq}  and \citet{cemri2025multi-auq} identify that the primary cause of task failure is not merely lack of knowledge, but \emph{“Spiral of Hallucination”} \cite{zhang2024language-auq, dziri2023faith-auq} where a minor hallucination in an early reasoning step (e.g., misinterpreting a tool output) pollutes the context window, biasing all subsequent planning steps towards an irreversible failure state.

To address this, a surge of recent work has focused on \emph{Automated Failure Attribution}. Approaches like \textit{AgentDebug} \citep{zhu2025llm-auq} and failure causality analysis \citep{zhangagent-auq} treat agent trajectories as diagnosable artifacts, utilizing separate critique models to trace root causes (e.g., planning brittleness vs. grounding errors) after the episode concludes. Other works have explored the dynamics of coordination collapse in multi-agent settings \citep{cemri2025multi-auq}, identifying how individual delusions can propagate to system-wide failures. While these contributions provide a valuable taxonomy of \textit{why} and \textit{where} agents break, they remain predominantly diagnostic or rely on distinct ``debugging layers'' \citep{zhu2025llm-auq}. They function as autopsies, explaining the death of the trajectory after it has occurred. Our Dual-Process framework fundamentally shifts this paradigm from post-hoc attribution to \textbf{runtime prevention}. By integrating UQ directly into the cognitive loop, we detect the onset of error propagation (System 1) and arrest the snowball effect via immediate reflection (System 2) before it cascades into total task failure.

\paragraph*{Uncertainty and Calibration in Agentic Systems.}
\label{sec:related_agent_calibration}

The study of uncertainty in LLM agents is an emerging but critical field, distinct from static text generation \citep{kirchhofposition-auq}. Pioneering works have begun to formalize the unique challenges agents present, particularly the sequential nature of confidence \citep{han2024towards-auq,tsai2024efficient-auq}. Frameworks like \textbf{UProp} \citep{duan2025uprop-auq} and \textbf{SAUP} \citep{duan2025uprop-auq} were among the first to explicitly model how uncertainty \textbf{propagates} through the sequential steps of an agent's trajectory, mathematically characterizing how local errors compound into global failures. Concurrently, other research has focused on quantifying the \textbf{external uncertainty} introduced by tool interactions, analyzing how API failures or noisy tool outputs impact overall reliability \citep{liu2024uncertainty-auq, lymperopoulos2025tools-auq}. While these studies laid the essential groundwork by identifying the core mechanics of propagation and external interaction, they primarily focus on \textit{high-level modeling} or \textit{passive diagnosis}. They quantify the risk but do not necessarily provide architectural mechanisms to resolve it during runtime. Our work bridges this gap by operationalizing these uncertainty signals. Instead of merely observing the propagation of doubt (as in UProp), our Dual-Process framework uses it as a trigger for \textbf{active intervention}, switching from fast System 1 execution to slow System 2 reflection to arrest the propagation before it becomes irreversible.

\paragraph*{Reflection and Self-Correction Mechanisms.}
\label{sec:related_reflection}

The ability to self-correct is a hallmark of autonomous agents. Early frameworks like \textbf{Self-Refine} \citep{madaan2023self-auq}, \textbf{Self-Reflection} \cite{renze2024self-auq}, and \textbf{Reflexion} \citep{shinn2023reflexion-auq} demonstrated that LLMs can improve their outputs through iterative refinement loops. However, these methods typically rely on \textit{explicit failure signals} or ground truth oracles to trigger correction. In open-ended reasoning tasks where such environmental feedback is absent, they often resort to ``blind'' reflection, leading to inefficient loops or the \textit{self-correction fallacy} \citep{huanglarge-auq} where the model confidently validates its own errors.

To address this, \textit{critique-and-refine} approaches like \textbf{CRITIC} \citep{goucritic-auq} and self-critiquing pipelines \citep{saunders2022self-auq} introduce verification steps using external tools or self-generated critiques. Yet, these methods suffer from the ``severity of hallucination'': if the model lacks the knowledge to solve the problem, it often lacks the knowledge to critique it, resulting in sycophantic confirmation. More recent advances like \textbf{STeCa} \citep{wang2025steca-auq} attempt to enforce calibration through trajectory-level reward modeling and supervised fine-tuning. While effective, STeCa requires expensive training on expert trajectories and is not easily adaptable to new base models. Our work aligns with the emerging trend of \textbf{Scaling Test-Time Compute} \citep{snell2024scaling-auq}, but with a critical distinction in \textit{activation}. Unlike Reflexion, which acts incessantly, or STeCa, which requires parameter updates, our \textbf{Dual-Process} framework offers a training-free, Pareto-optimal alternative. We utilize internal epistemic uncertainty as a calibrated ``Stop'' or ``Switch'' signal, ensuring that expensive reflective compute (System 2) is deployed \textit{only} when the agent detects a genuine risk of failure, thereby balancing reliability with token efficiency.

\subsection{Formal Mathematics in Problem Formulation}
\label{app:formal_math}

We provide the rigorous probabilistic definitions and derivations corresponding to the Problem Formulation in Section~\ref{sec:problem_formulation}.

\subsubsection{Forward Problem: Recursive Validity Estimation}
In Equation (1), we defined the forward problem as estimating the trajectory validity $P(V_t | h_t)$. Here, we explicate the recursive function $f_{\text{p}}$. Let $V_t \in \{0, 1\}$ be a binary random variable indicating that the trajectory up to step $t$ is free of critical epistemic errors. By the chain rule of probability, the validity of the current state depends on the validity of the history and the correctness of the current action:
\begin{align}
    P(V_t=1 | h_t) &= P(V_t=1 | V_{t-1}=1, a_t, h_{t-1}) \cdot P(V_{t-1}=1 | h_{t-1}) \\
    &= \underbrace{P(\text{Correct}(a_t) | h_t)}_{\text{Local Confidence } c_t} \cdot \underbrace{P(V_{t-1}=1 | h_{t-1})}_{\text{Historical Validity}}
\end{align}
This recursive product implies that $P(V_t=1)$ is monotonically non-increasing with respect to $t$. This mathematically formalizes the \textbf{Spiral of Hallucination}: a single failure at step $k$ ($P(\text{Correct}_k) \approx 0$) drives the joint probability to zero for all $t > k$, permanently invalidating the trajectory.
Our Forward UQ approximates this joint probability by aggregating verbalized confidence scores: $P(V_t | h_t) \approx \prod_{i=0}^t \hat{c}_i$ (or via a conservative minimum function $\min(\hat{c}_{0:t})$).

\subsubsection{Inverse Problem: Latent Variable Calibration}
We view this as \emph{Test-Time Calibration} \citep{cobbe2021training-auq}. When the forward process yields low confidence, we treat the generation of a reliable plan as an inverse search problem. By utilizing inference-time compute (e.g., reflection \citep{shinn2023reflexion-auq}), we approximate the posterior distribution of correct reasoning without updating model parameters.  

In Equation (2), we framed the inverse problem as maximizing the posterior via a latent reasoning path $z$. Here, we derive this form from Bayesian decision theory. We treat the task success as an optimality variable $\mathcal{O}$ (where $\mathcal{O}=1$ implies $\text{Succ}$). We introduce a latent variable $z$ representing the \textit{reasoning trace} (e.g., a Chain-of-Thought or a Reflection explanation). We aim to find the action $a^*$ that maximizes the posterior:
\begin{equation}
    a^* = \operatorname*{arg\,max}_{a} P(a | h_t, \mathcal{O}=1)
\end{equation}
By marginalizing over the latent reasoning $z$, we expand the posterior:
\begin{align}
    P(a | h_t, \mathcal{O}=1) &= \int P(a, z | h_t, \mathcal{O}=1) \, dz \\
    &= \int P(a | z, h_t) \cdot P(z | h_t, \mathcal{O}=1) \, dz
\end{align}
Using Bayes' rule on the second term:
\begin{equation}
    P(z | h_t, \mathcal{O}=1) \propto \underbrace{P(\mathcal{O}=1 | z, h_t)}_{\text{Likelihood (Consistency)}} \cdot \underbrace{\pi_{\text{prior}}(z | h_t)}_{\text{Generation Prior}}
\end{equation}
Substituting this back, we recover the objective in Equation (2).
Since the integral over all possible thoughts $z$ is intractable, our \textbf{Best-of-N Reflection} strategy performs a Monte Carlo approximation:
\begin{enumerate}
    \item \textbf{Sample:} Generate $N$ reasoning paths $\{z_i\}_{i=1}^N$ from the prior $\pi(z|h_t)$.
    \item \textbf{Reweight:} Estimate $P(\mathcal{O}=1 | z_i, h_t)$ using self-consistency or verbalized confidence as a proxy.
    \item \textbf{Argmax:} Select the action $a$ associated with the highest weighted $z$.
\end{enumerate}
This derivation proves that our method is a particle-based approximation of the optimal inverse calibration.


\subsection{Detailed Experimental Setup}
\label{app:setup_details}

\subsubsection{Datasets and Evaluation Protocols}
\label{app:datasets}

We selected three benchmarks that span the spectrum of agentic capabilities, ranging from rigid, logic-heavy planning to open-ended, creative synthesis. This diversity ensures that our Agentic UQ framework is not overfit to a specific modality but is generally applicable.

\paragraph*{ALFWorld: Embodied Decision Making (Logic \& Planning).}

ALFWorld aligns TextWorld with the ALFRED benchmark, creating a text-based simulated household. The agent must solve high-level goals (e.g., ``clean the apple and put it in the fridge'') by executing a sequence of low-level actions (e.g., \texttt{open fridge}, \texttt{put apple}).
\begin{itemize}
    \item It represents \textbf{Deterministic Planning}. The environment is logically consistent but requires long-horizon dependency tracking. A single missing step (e.g., forgetting to open the fridge before putting the apple) causes failure. This tests the agent's ability to maintain \textit{Cognitive Continuity}.
    \item We utilize the \textbf{Seen Evaluation Set}, comprising \textbf{140 unique environments}. This setting tests the agent's reliability in handling known domain structures. The evaluation metric is Success Rate (SR).
\end{itemize}

\paragraph*{WebShop: Noisy Web Navigation (Robustness \& Search).}
WebShop simulates a large-scale e-commerce website with 1.18 million products. The agent is given a user instruction (often containing implicit constraints like ``under \$50'') and must search, browse, and select the correct product options.
\begin{itemize}
    \item It represents \textbf{Stochastic Environment Interaction}. Unlike ALFWorld, WebShop is highly noisy: search engines return irrelevant results, and product descriptions are verbose and unstructured. This tests the agent's ability to use \textit{Inverse UQ} to filter noise and verify information before committing.
    \item To maintain sample size consistency with ALFWorld, we randomly sampled \textbf{140 episodes} from the standard \textbf{Development Set}. The evaluation metric is Success Rate (SR).
\end{itemize}

\paragraph*{DeepResearch Bench: Open-Ended Synthesis (Synthesis \& Depth).}

Created by 100+ domain experts, this benchmark consists of \textbf{100 PhD-level research tasks} (50 in Chinese, 50 in English) spanning 22 distinct fields. The agent must perform autonomous web research to generate a comprehensive report.
\begin{itemize}
    \item It represents \textbf{Open-Ended Reasoning}. There is no single ``correct'' sequence of actions. Success depends on the depth of insight and the coherence of the final report. This tests whether \textit{Confidence-Aware Reflection} can drive deeper investigation rather than superficial summarization.
    \item  We utilize the \textbf{Reference-based Adaptive Criteria-driven Evaluation (RACE)} framework \cite{du2025deepresearch-auq} with Dynamic Weighting. Unlike static grading, RACE dynamically adjusts evaluation criteria based on the reference report's complexity, assessing the agent on dimensions such as comprehensiveness, insight, instruction following, and readability. We focus on the RACE score to evaluate generation quality, excluding the FACT (citation) metric to isolate reasoning capabilities.
\end{itemize}

\subsubsection{Baselines and Variants}
\label{app:baselines}

Here we provide the exact operational details for each method to ensure reproducibility.

\paragraph*{Baselines Implementation}
\begin{itemize}
    \item \textbf{ReAct} \cite{yao2022react-auq}: The standard System 1 baseline. The agent reasons and acts in an interleaved manner without explicit self-evaluation or reflection. We use the standard prompt template provided in the original paper. The temperature is set to $0.0$ for deterministic greedy decoding.
    \item \textbf{Reflexion} \cite{shinn2023reflexion-auq}: A strong \textit{inter-episode} learning baseline. It persists through verbal reinforcement from past failures to improve future trials. We evaluate Reflexion after 2 accumulated failure trials to measure its few-shot adaptation capability. We implement the standard Reflexion loop: \texttt{Act} $\rightarrow$ \texttt{Fail} $\rightarrow$ \texttt{Reflect} $\rightarrow$ \texttt{Store in Long-term Memory}. For a fair comparison with our inference-time method, we allow Reflexion to accumulate experience over 2 failure trials on the same instance before measuring the performance on the 3rd trial. 
    \item \textbf{Self-Reflection} \cite{renze2024self-auq}: An \textit{intra-episode} baseline that triggers reflection on every step (or uses a heuristic). Unlike our method, it lacks confidence calibration and simply asks the model to ``double-check'' its action, serving as a baseline for unguided compute scaling. At every step, the agent generates a thought and action, then enters a mandatory ``Check''phase: \textit{``Review your proposed action. Is it correct? If not, generate a new one.``}  This method often suffers from ``double-checking fatigue,'' where the model blindly confirms its own action, or ``over-correction,'' where it changes a correct action due to anxiety.
    \item \textbf{CoT-SC} \cite{wangself-auq}: A statistical ensemble baseline. For every decision step, we sample $N=6$ independent trajectories (Action + Thought) with temperature $T=0.7$. We apply majority voting on the final Action string. If there is a tie, we select the one with the highest average log-probability. This tests whether our gains stem purely from sampling diversity rather than targeted reflection. 
\end{itemize}

\paragraph*{Dual-Process Variants Implementation}
All our variants use the same base prompt structure but differ in their control flow and memory content.

\begin{itemize}
    \item \textbf{Forward (\uamnospace-Only):} The agent generates and stores verbalized confidence in memory to constrain future exploration, but \textit{never} triggers the System 2 reflection loop. This tests the efficacy of predictive uncertainty propagation. 
    \begin{itemize}
        \item \textit{Process:} Step $t$ generates $({a}_t, \hat{c}_t, \hat{e}_t)$.
        \item \textit{Memory Update:} $\mathcal{M}_{t+1} \leftarrow \mathcal{M}_t \cup \{(o_t, {a}_t, \hat{c}_t, \hat{e}_t)\}$.
        \item \textit{Constraint:} The System 2 loop is strictly disabled (Threshold $\tau = -\infty$). The agent must rely on the presence of $\hat{e}_t$ in the context window to adjust its behavior for step $t+1$.
    \end{itemize}

    \item \textbf{Inverse (\uarnospace-Only):} The agent can trigger System 2 reflection to correct low-confidence steps, but does \textit{not} persist the confidence metadata $(\hat{c}, \hat{e})$ into long-term memory. This tests the benefit of local correction without cognitive continuity.
    \begin{itemize}
        \item \textit{Process:} If $\hat{c}_t < \tau$, trigger Reflection + Best-of-N to get corrected action $a^*_t$.
        \item \textit{Memory Update (Crucial Difference):} $\mathcal{M}_{t+1} \leftarrow \mathcal{M}_t \cup \{({a}^*_t, o_t)\}$.
        \item \textit{Constraint:} The confidence scores and explanations are \textbf{discarded} after the step is finalized. Future steps do not see the ``cognitive meta-data''of the past, simulating a ``forgetful'' but capable agent.
    \end{itemize}

    \item \textbf{Dual-Process (Full, \auqnospace):} The complete framework combining UAM propagation and UAR correction with adaptive switching.
    \begin{itemize}
        \item \textit{Process:} If $\hat{c}_t < \tau$, trigger Reflection.
        \item \textit{Memory Update:} $\mathcal{M}_{t+1} \leftarrow \mathcal{M}_t \cup \{(o_t, {a}^*_t, \hat{c}_{new}, \hat{e}_{new})\}$.
        \item \textit{Constraint:} Combines both local correction and long-term uncertainty propagation.
    \end{itemize}
\end{itemize}

\subsubsection{Implementation Details}
\label{app:implementation}

\subsection*{Calculation of Consistency-Weighted Confidence}
\label{sec:appendix_consistency_calc}
Given $N$ sampled reasoning paths $\{r_1, ..., r_N\}$, each producing a final action $a_i$ and a verbalized confidence $\hat{c}_i$, we first aggregate the paths into semantic clusters $C_1, ..., C_K$ such that all actions in a cluster are semantically equivalent. The score for a candidate action $a$ is calculated as:

\begin{equation}
    S(a) = \underbrace{\frac{|C_a|}{N}}_{\text{Consistency}} \times \underbrace{\frac{1}{|C_a|} \sum_{i \in C_a} \hat{c}_i}_{\text{Mean Confidence}}
\end{equation}

This formulation separates diversity (consistency) from epistemic strength (confidence).  Our consistency-weighted score $S_{cons}(a)$ (Equation \ref{eq:cw}) implicitly approximates the negative entropy of the System 2 distribution. A low consistency score indicates high variance among the $N$ sampled paths, serving as a proxy for \textit{Conflictual Uncertainty} (Epistemic), whereas a uniformly low $\hat{c}_t$ across consistent answers would indicate \textit{Task Difficulty} (Aleatoric). By filtering candidates based on $S_{cons}$, we specifically target epistemic hallucinations.  The semantic equivalence function $a_i \approx a_j$ is implemented differently depending on the task domain:
\begin{itemize}
\item Structured Environments (ALFWorld, WebShop): 
Since these environments require valid API calls, we employ \emph{Normalized String Matching}. We extract the content within the \texttt{<action>} tags, convert to lowercase, and strip trailing whitespace.
\begin{equation}
    \mathbb{I}(a_i, a_j) = \begin{cases} 1 & \text{if } \text{norm}(a_i) = \text{norm}(a_j) \\ 0 & \text{otherwise} \end{cases}
\end{equation}
\item Open-Ended Reasoning (Deep Research): 
For high-level planning tasks where identical intents may be phrased differently (e.g., ``Search for X then Y''vs. ``First find X, followed by Y''), strict string matching is insufficient. We employ a \emph{Model-Based Equivalence Check}. We define $\mathbb{I}(a_i, a_j) = 1$ if the model (System 1) predicts ``Yes'' to the prompt: \textit{``Do these two plans represent the same core information-seeking strategy?''}. In practice, to reduce latency, we often strictly enforce output formatting (e.g., specific JSON keys) to revert to string matching where possible.
\end{itemize}

\paragraph*{Model Specifications.}
We access proprietary models via their official APIs and host open-weights models using vLLM on a cluster of H200 GPUs. Table~\ref{tab:model_specs} lists the specific versions used in this study.

\begin{table}[h]
    \centering
    \small
    \begin{tabular}{l|l|l}
        \toprule
        \textbf{Model Family} & \textbf{Specific Checkpoint/Version} & \textbf{Role in Experiments} \\
        \midrule
        \multirow{3}{*}{OpenAI} & \texttt{gpt-5.1-preview} & Strong Reasoning Baseline \\
        & \texttt{gpt-4.1} & Strong Baseline \\
        & \texttt{gpt-4o} & Efficiency Baseline \\
        \midrule
        \multirow{2}{*}{Google} & \texttt{gemini-2.5-pro} & Long-Context Specialist \\
        & \texttt{gemini-2.5-flash} & High-Speed/Low-Cost \\
        \midrule
        \multirow{2}{*}{Open-source LLMs} & \texttt{Qwen3-235B-Instruct} & Strong Open Model \\
        & \texttt{DeepSeek-V3.1-Chat} & Reasoning-Dense Open Model \\
        \bottomrule
    \end{tabular}
    \caption{List of LLMs evaluated in our experiments.}
    \label{tab:model_specs}
\end{table}

\paragraph*{Hyperparameter Settings.}
We enforce strict hyperparameter control to ensure fair comparison. Table~\ref{tab:hyperparams} details the specific settings for the Dual-Process framework.

\begin{table}[h]
    \centering
    \small
    \begin{tabular}{l|c|l}
        \toprule
        \textbf{Parameter} & \textbf{Value} & \textbf{Description} \\
        \midrule
        \multicolumn{3}{l}{\textit{General Inference}} \\
        Max Trajectory Steps & 50 & Early stop for ALFWorld/WebShop \\
        Temperature (Greedy) & 0.0 & Used for ReAct and System 1 (Forward) \\
        Temperature (Sampling) & 0.7 & Used for System 2 (Inverse/Reflection) \\
        \midrule
        \multicolumn{3}{l}{\textit{Dual-Process Control}} \\
        Threshold $\tau$ (Standard) & $\{0.8, 0.85, 0.9, 0.95\}$ & Ablated range; 0.85 is default \\
        Threshold $\tau$ (Research) & $\{0.8, 0.85, 0.9, 0.95\}$ & Ablated range; 0.95 for DeepResearch Bench \\
        \midrule
        \multicolumn{3}{l}{\textit{Inverse UQ (System 2)}} \\
        Sampling Width $N$ & 3 & Number of parallel candidate paths \\
        Reflection Depth $D$ & 3 & Max iterative correction rounds per path \\
        Memory Expansion & True & Triggered if $h=\text{limit}$ and $\hat{c} < \tau$ after reflection \\
        \bottomrule
    \end{tabular}
    \caption{Hyperparameters for the Agentic UQ Framework.}
    \label{tab:hyperparams}
\end{table}

\paragraph*{Reflection and Memory Expansion Logic.}
The Inverse UQ process follows a hierarchical search:
\begin{enumerate}
    \item \textbf{Parallel Sampling:} Upon triggering System 2, we generate $N=3$ parallel responses using a temperature of $0.7$ to encourage diverse reasoning paths.
    \item \textbf{Iterative Refinement:} For each path, if the verbalized confidence is below $\tau$, the agent is prompted to critique and refine its action. This repeats for a maximum of $D=3$ turns.
    \item \textbf{Expansion Trigger (Conditional):} In the \textit{Limited Memory} setting ($h=5$), if the best candidate from the reflection loop still fails to meet the threshold $\tau$, the \textbf{Adaptive Memory Expansion} protocol is activated. The agent retrieves the full trajectory history from an external buffer and re-runs the reflection loop once. This ensures that expensive long-context processing is only reserved for the most stubborn uncertainties.
\end{enumerate}

\paragraph*{Integrate Agentic UQ with Enterprise Deep Research (EDR).}
\label{sec:appendix_edr_integration}

The standard EDR architecture operates as a hierarchical multi-agent system \cite{prabhakar2025enterprise-auq}: a \textbf{Planner} (“master research agent”) first decomposes the user topic into parallel sub-questions, which are then executed by \textbf{Researcher} (“specialized agent”) agents using tools (e.g., Tavily/Google Search), followed by an \textbf{Analyst} (“research report”) that synthesizes the results. We integrate our Dual-Process framework specifically at the \textbf{Planner-Researcher Interface}, utilizing a control strategy:

\begin{enumerate}
    \item \textbf{Planner Augmentation:} We modify the Planner's system prompt to include our \textit{Confidence Elicitation Protocol}. Instead of directly outputting a JSON list of sub-queries, the Planner is instructed to first output a \texttt{<confidence>} score and an \texttt{<explanation>} of potential knowledge gaps.
    
    \item \textbf{Execution Interception (The Switch):} 
    Before the Planner's output is parsed into the task queue for Researcher agents, the \auq module acts as a gatekeeper:
    \begin{itemize}
        \item \textbf{Pass-through (System 1):} If confidence $\ge \tau$, the decomposition is parsed immediately, and Researcher agents are dispatched in parallel. This incurs zero latency overhead.
        \item \textbf{Intervention (System 2):} If confidence $< \tau$, the \auq module {blocks} the dispatch of Researchers. It initiates the \textit{Reflection Loop}, feeding the Planner's own explanation back to it to regenerate a refined plan (e.g., switching from General Search to Academic Search tools, or adding granular sub-queries).
    \end{itemize}
    
    \item \textbf{Seamless Integration:} Once the System 2 loop resolves the uncertainty (or reaches max depth), the \textit{corrected} decomposition plan is injected back into the EDR pipeline. The Researcher agents then execute this improved plan, unaware that an intervention occurred.
\end{enumerate}

This architecture demonstrates that \auq is \textbf{framework-agnostic}: it improves the quality of the EDR system's output (Insight/Comprehensiveness) solely by optimizing the planning instructions, without requiring changes to the complex asynchronous search or synthesis logic.

\subsubsection{Evaluation Metrics and Protocols}
\label{app:metrics}

\paragraph*{Trajectory-Level Calibration Metrics.}
Standard token-level metrics are insufficient for sequential tasks. We define a trajectory confidence sequence $\mathbf{c} = \{\hat{c}_1, \dots, \hat{c}_T\}$. We map this to a scalar trajectory belief $C(\tau) = \Phi(\mathbf{c})$ using three aggregation strategies: \textit{End-State} ($\Phi_{\text{last}}$), \textit{Overall Quality} ($\Phi_{\text{avg}}$), and \textit{Process Reliability} ($\Phi_{\text{min}}$).
\begin{itemize}
\item \textbf{Trajectory-ECE (T-ECE).} We partition test trajectories into $M$ bins $\{\mathcal{B}_m\}_{m=1}^M$ based on their aggregated belief $C(\tau)$. T-ECE measures the weighted absolute difference between confidence and accuracy:
\begin{equation}
    \text{T-ECE}_{\Phi} = \sum_{m=1}^M \frac{|\mathcal{B}_m|}{N} \left| \underbrace{\text{acc}(\mathcal{B}_m)}_{\text{Avg } Y} - \underbrace{\text{conf}(\mathcal{B}_m)}_{\text{Avg } C(\tau)} \right|
\end{equation}
where $Y \in \{0, 1\}$ is the binary task success.
\item \textbf{Trajectory Brier Score (T-BS).} To jointly evaluate calibration and sharpness (decisiveness), we utilize the Mean Squared Error of the probabilistic prediction:
\begin{equation}
    \text{T-BS}_{\Phi} = \frac{1}{N} \sum_{i=1}^N (C(\tau^{(i)}) - Y^{(i)})^2
\end{equation}
\item \textbf{Discriminative AUROC.} We calculate the AUROC by treating the aggregated confidence $C(\tau)$ as a binary classifier score for task success. A score of 0.5 indicates random guessing, while 1.0 indicates perfect separation between success and failure trajectories.
\end{itemize}

\paragraph*{Deep Research Judge Protocol.}
Evaluating open-ended research reports requires nuanced judgment. We utilize \textbf{Gemini-2.5-Pro} as the evaluator. The model follows the RACE protocol \cite{du2025deepresearch-auq}, scoring on a 1-10 Likert scale across four dimensions (comprehensiveness, insight/depth, instruction following, readability), which are then normalized to a 0-100 scale.
\subsection{Dual-Process Agentic UQ Framework (\auqnospace)}
\label{sec:appendix_algorithm}

We provide the formal execution flow of the Dual-Process Agentic UQ framework in Algorithm \ref{alg:dual_process}.  The algorithm proceeds in three phases at each time step $t$:
\begin{itemize}
    \item \textbf{System 1 - Uncertainty Propagation:} The agent observes the current state $o_t$ and conditions on the accumulated uncertainty-aware memory $\mathcal{M}_t$. It jointly generates a tentative action $\hat{a}_t$, a scalar confidence $\hat{c}_t$, and a semantic explanation $\hat{e}_t$. The explanation serves as a ``rationale cue'' for potential debugging.
    \item \textbf{System 2 - Uncertainty Reflection:} The framework evaluates the confidence against the threshold $\tau$. 
    \begin{itemize}
        \item If $\hat{c}_t \ge \tau$, the system trusts the fast intuition and proceeds immediately.
        \item If $\hat{c}_t < \tau$, System 2 is activated. The agent uses the generated explanation $\hat{e}_t$ as a specific query to guide a \textit{Best-of-N} sampling process. The final action $a_t$ is selected via confidence-weighted consistency voting, effectively filtering out hallucinated paths.
    \end{itemize}
    \item \textbf{Memory Consolidation:} Crucially, the final executed tuple $(o_t, a_t, \hat{c}_t, \hat{e}_t)$ is appended to $\mathcal{M}$. This ensures that future steps are conditioned on the agent's past epistemic states, enabling the ``Cognitive Damper'' effect where past doubts influence future caution.
\end{itemize}

\begin{algorithm}[H]
\caption{Dual-Process Agentic Uncertainty Quantification}
\label{alg:dual_process}
\begin{algorithmic}[1]
\Require Task Instruction $I$, Threshold $\tau$, Sampling Count $N$, Max Steps $T_{max}$
\State \textbf{Initialize:} Memory $\mathcal{M}_0 \leftarrow \emptyset$, Observation $o_0 \leftarrow \text{Env}.\text{reset}(I)$

\For{$t = 0$ \textbf{to} $T_{max}$}
    \State \textbf{// Phase 1: System 1 Fast Execution}
    \State Construct prompt $P_t \leftarrow f_{prompt}(I, \mathcal{M}_t, o_t)$
    \State Generate initial proposal: 
    \State \quad $\hat{a}_t, \hat{c}_t, \hat{e}_t \sim \pi_{\theta}(\cdot | P_t)$ 
    
    \State \textbf{// Phase 2: Uncertainty Switch}
    \If{$\hat{c}_t \ge \tau$}
        \State \textbf{// High Confidence: Pass-through}
        \State Set final action $a_t \leftarrow \hat{a}_t$
    \Else
        \State \textbf{// Low Confidence: System 2 Reflection (Triggered)}
        \State Define candidate set $\mathcal{A}_{cand} \leftarrow \emptyset$
        \State Construct reflection prompt $P_{ref}$ using rational cue $\hat{e}_t$
        
        \For{$n = 1$ \textbf{to} $N$}
            \State Sample reasoning path: $r^{(n)}, a^{(n)}, \hat{c}^{(n)} \sim \pi_{\theta}(\cdot | P_{ref})$
            \State Add to candidates: $\mathcal{A}_{cand} \leftarrow \mathcal{A}_{cand} \cup \{(a^{(n)}, \hat{c}^{(n)})\}$
        \EndFor
        
        \State \textbf{// Inverse UQ: Consistency-Weighted Selection}
        \State Calculate score for each unique action $a$:
        \State \quad $S(a) = \sum_{(a^{(i)}, \hat{c}^{(i)}) \in \mathcal{A}_{cand}} \mathbb{I}(a^{(i)} \approx a) \cdot \hat{c}^{(i)}$
        \State Select optimal action: $a_t \leftarrow \arg\max_{a} S(a)$
        \State Update confidence/explanation to match selected path's values
    \EndIf
    
    \State \textbf{// Phase 3: Execution \& Memory Update}
    \State Execute action: $o_{t+1}, \text{done} \leftarrow \text{Env}.\text{step}(a_t)$
    \State Update Uncertainty-Aware Memory:
    \State \quad $\mathcal{M}_{t+1} \leftarrow \mathcal{M}_t \cup \{ (o_t, a_t, \hat{c}_t, \hat{e}_t) \}$
    
    \If{$\text{done}$}
        \State \textbf{Break}
    \EndIf
\EndFor
\end{algorithmic}
\end{algorithm}

\subsection{Additional Experimental Results and Analysis}

\subsubsection{Extended Analysis of Internal Dynamics and Risks}
\label{app:internal_dynamics}



The scatter plots in Figure~\ref{fig:dynamics} (Right) revealed a counter-intuitive phenomenon where failure cases often exhibit larger confidence gains ($\Delta$) than successes. To explain this, we categorize the System 2 intervention outcomes into three distinct modes based on initial confidence $c_{init}$ and final status:

\paragraph*{1. Validation (High $c_{init} \to$ High $c_{final}$, Success).}
This is the dominant mode for successful trajectories. The agent starts with high confidence (typically $c_{init} > 0.9$). The reflection loop serves as a sanity check, confirming the plan is sound. Since the confidence is already near the ceiling (1.0), the potential gain $\Delta$ is minimal. This explains the tight clustering of green points in the top-right of the scatter plots.

\paragraph*{2. True Correction (Low $c_{init} \to$ High $c_{final}$, Success).}
This represents the ideal System 2 intervention. The agent starts in the ``Ambiguous Zone''($c_{init} \approx 0.6 - 0.8$), correctly identifying a gap in its reasoning. Reflection synthesizes a corrected plan, boosting confidence significantly (e.g., $\Delta \approx +0.3$).

\paragraph*{3. Delusional Confirmation (Low $c_{init} \to$ High $c_{final}$, Failure).}
This is the root of the ``delusional gap.''In challenging situations, the agent initially has low confidence. However, instead of realizing the task is impossible, it generates a seemingly plausible but actually incorrect explanation through its reflection mechanism. The agent then adopts this fabricated solution with extremely high confidence. This significant increase in confidence (e.g., from 0.5 to 0.9) distorts the average statistics, creating the illusion that failing agents are ``more confident''in their improvements than successful agents.


\subsection*{Threshold Sensitivity and Calibration Profiles}
\label{app:threshold_model_analysis}
We further analyze how the sensitivity threshold $\tau$ and model scale influence the trade-off between the correction modes defined above.

\begin{figure}[!h]
    \centering
    \includegraphics[width=0.4\linewidth]{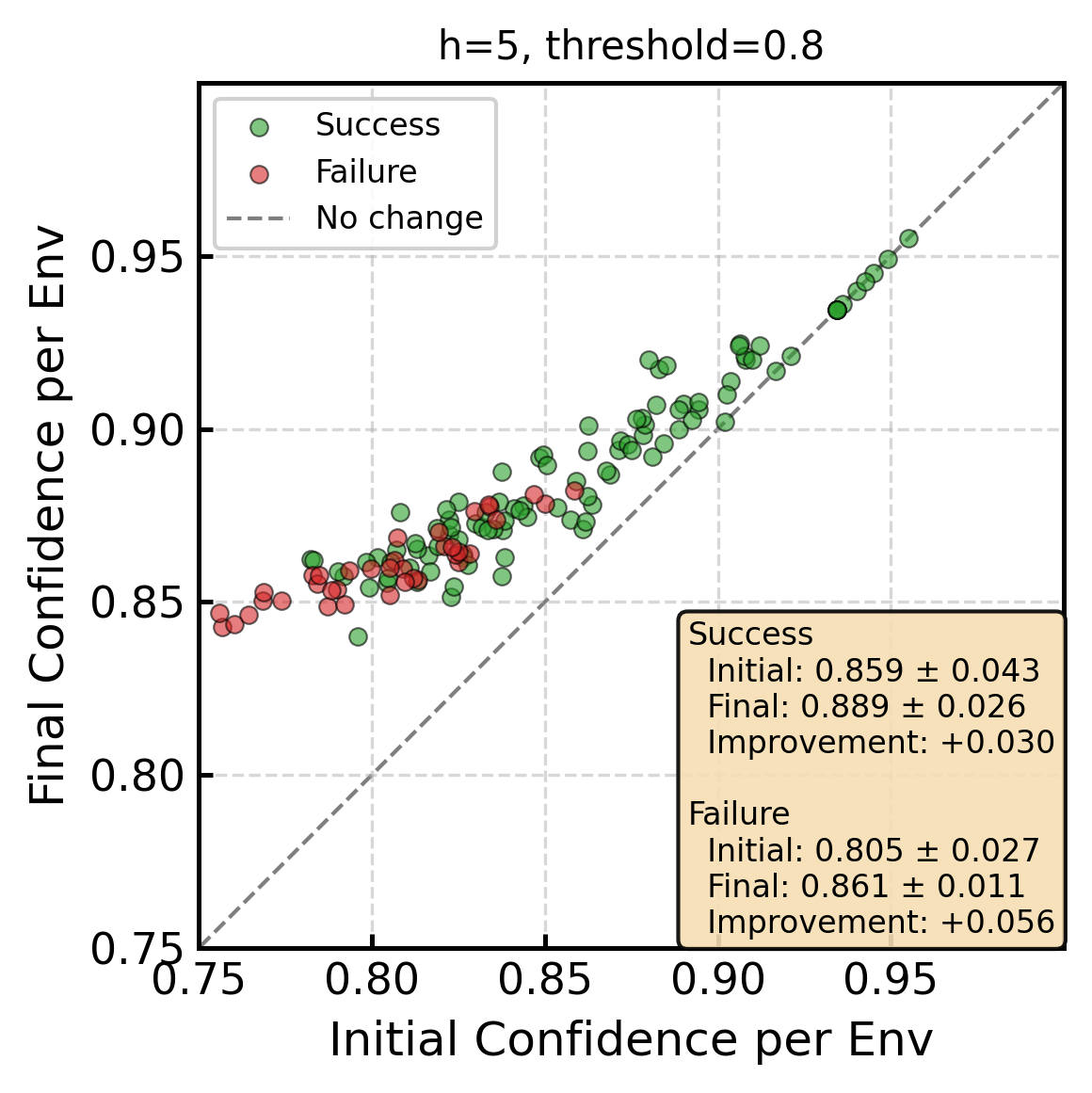} 
    \includegraphics[width=0.4\linewidth]{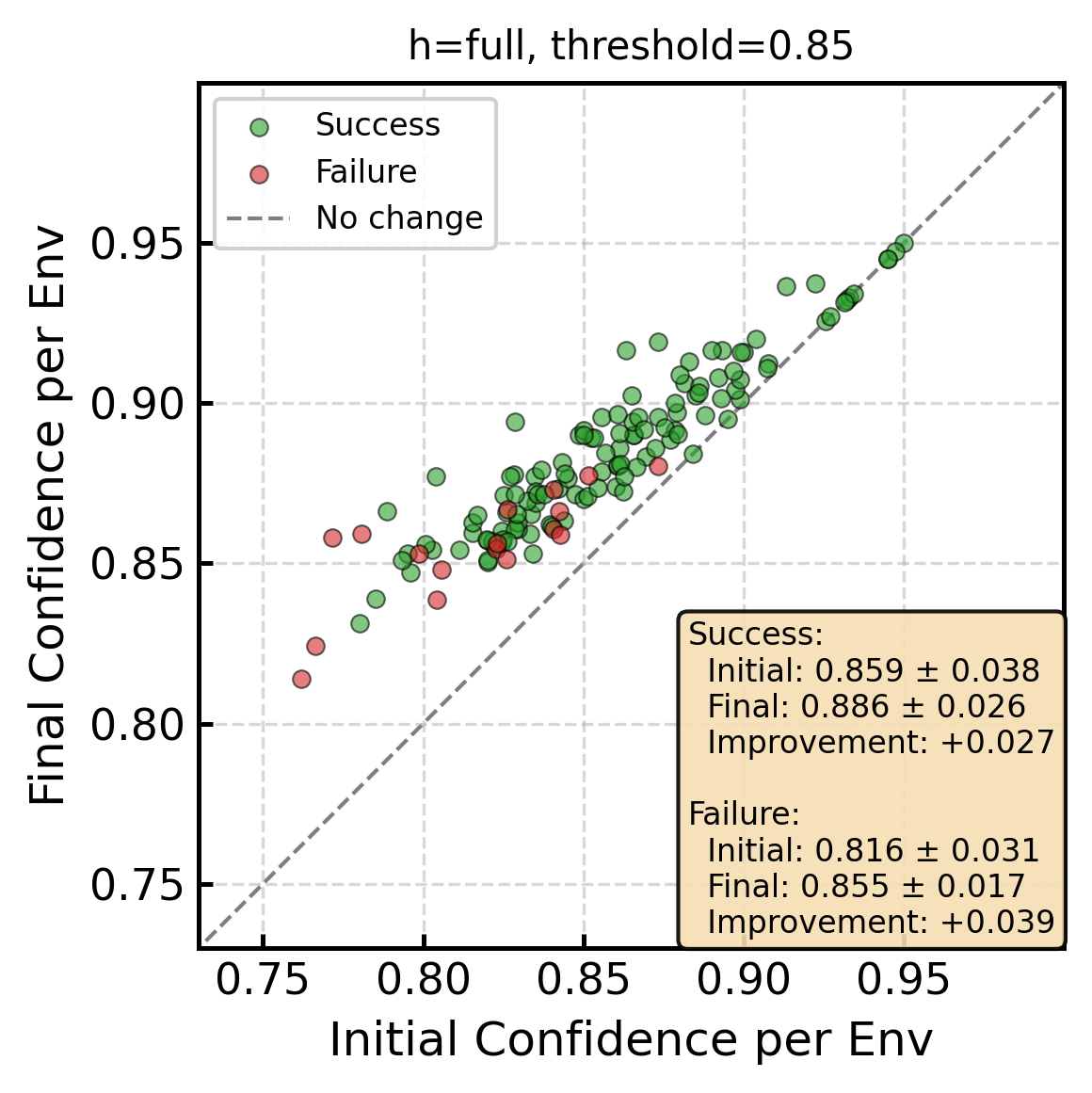} 
    \includegraphics[width=0.4\linewidth]{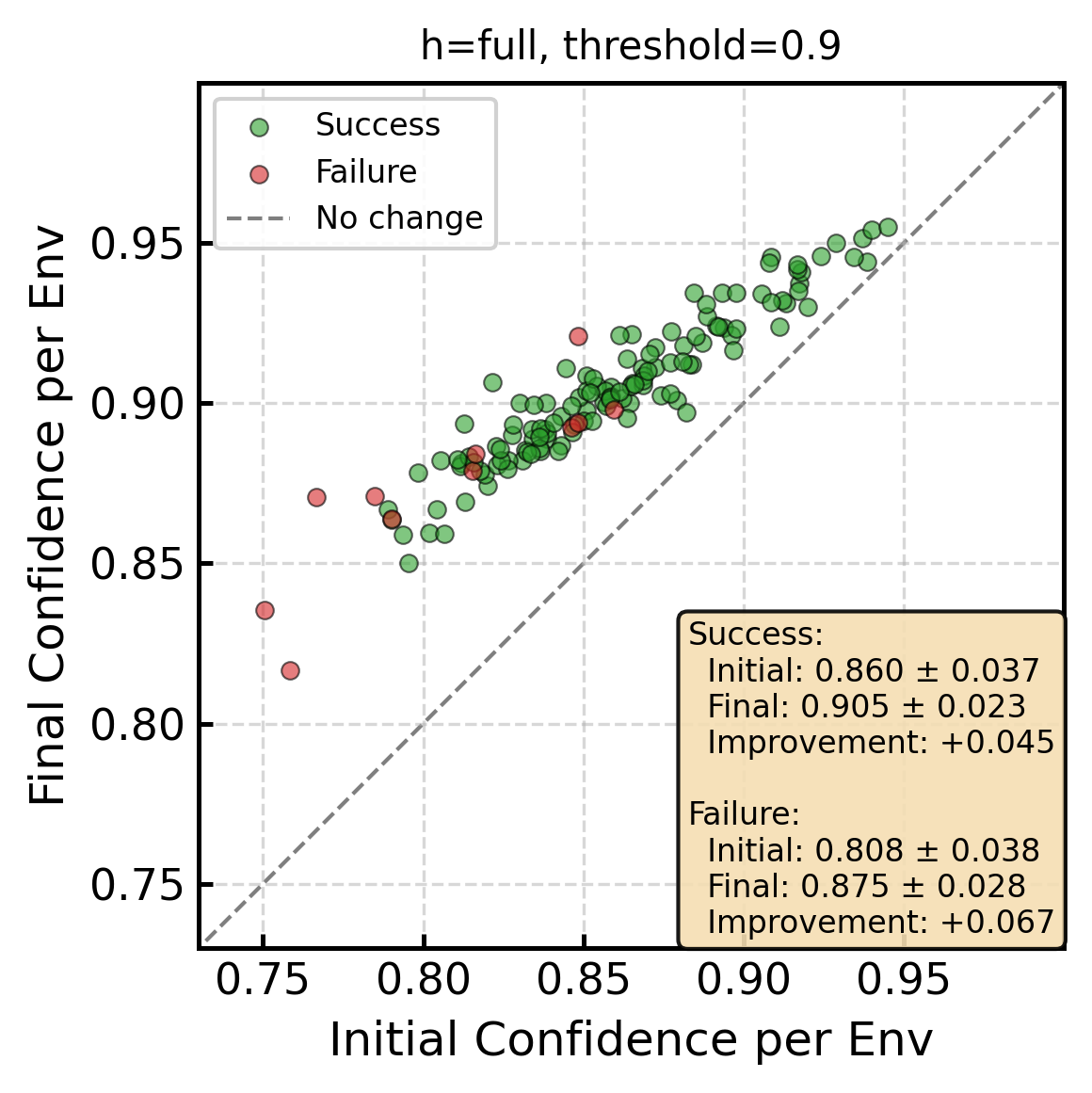} 
    \includegraphics[width=0.4\linewidth]{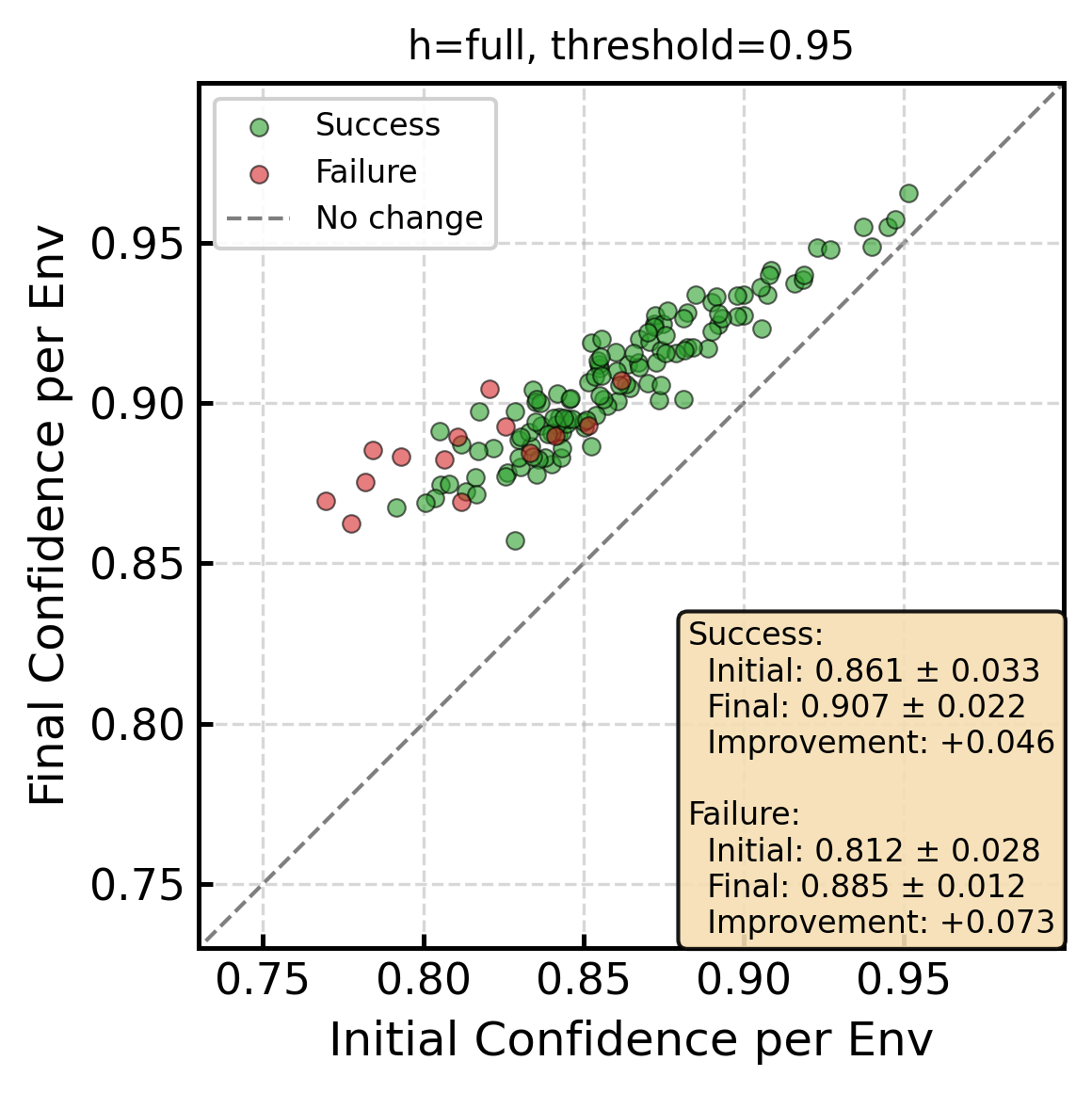} 
    \caption{\textbf{Detailed Confidence Dynamics for GPT-4o across Thresholds.} We visualize the reflection gain for $\tau \in \{0.8, 0.85, 0.9, 0.95\}$. Lower thresholds (e.g., 0.8) trigger sparsely on obvious errors, while higher thresholds (e.g., 0.95) induce broad recalibration. The positive shift (points above the diagonal) remains consistent across all settings.}
    \label{fig:gpt4o_breakdown}
\end{figure}

\paragraph*{The Trigger-Efficiency Trade-off.}
The choice of threshold modulates the system's behavior:
\begin{itemize}
    \item \textbf{Low Threshold ($\tau=0.8$, ``Loose Filter''):} The mechanism only activates for obvious errors. While it minimizes the risk of Regression (breaking good steps), it misses many ``True Corrections'' in the Ambiguous Zone, limiting the overall Success Rate gain.
    \item \textbf{High Threshold ($\tau=0.95$, ``Strict Forcer''):} The system acts aggressively, verifying even moderately confident steps. This maximizes the error correction rate (recall), but also increases the probability of incorrect confirmations. Crucially, however, Figure~\ref{fig:gpt4o_breakdown} shows that \textbf{System 2 rarely downgrades a correct high-confidence step} (few green points below the diagonal). This ``safe failure''characteristic indicates that increasing sensitivity increases computational cost but does not significantly compromise reliability.
\end{itemize}

\begin{figure}[!h]
    \centering
    \includegraphics[width=0.4\linewidth]{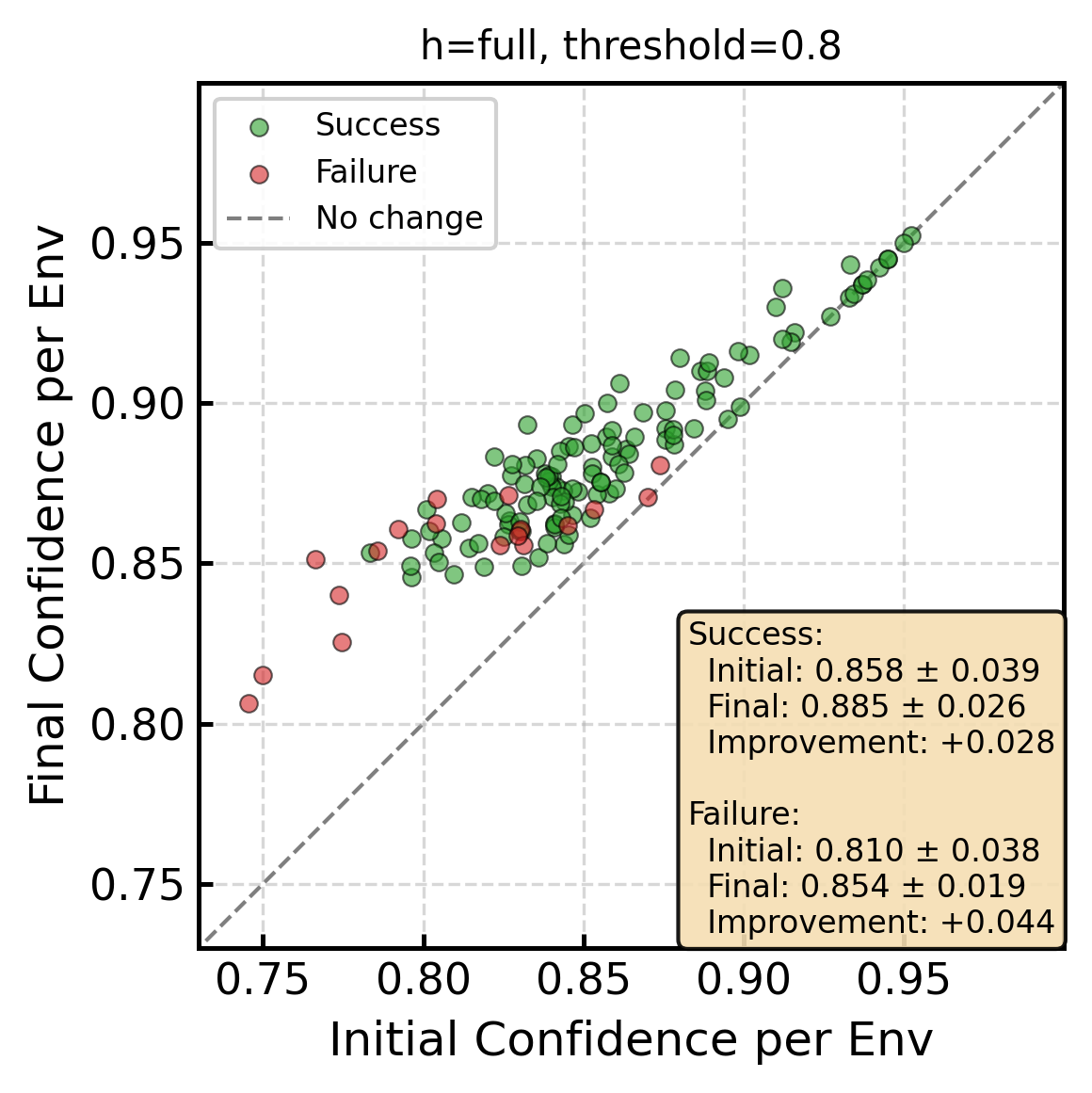} 
    \includegraphics[width=0.4\linewidth]{gpt51_full_threshold_0.85_scatter.png} 
    \includegraphics[width=0.4\linewidth]{gpt51_full_threshold_0.9_scatter.png} 
    \includegraphics[width=0.4\linewidth]{gpt51_full_threshold_0.95_scatter.png} 
    \caption{{Detailed Confidence Dynamics for GPT-5.1 across Thresholds.} Compared to GPT-4o, GPT-5.1 exhibits tighter clustering and higher intrinsic confidence. Even at strict thresholds ($\tau=0.95$), the mechanism effectively nudges uncertain predictions towards certainty (1.0) or rejection (0.0).}
    \label{fig:gpt51_breakdown}
\end{figure}

\paragraph*{Model-Specific Calibration.}
Comparing GPT-4o and GPT-5.1 reveals distinct personalities:
\begin{itemize}
    \item \textbf{GPT-4o (Figure~\ref{fig:gpt4o_breakdown}):} Shows a wider spread in initial confidence. The reflection mechanism often functions as a ``booster,'' lifting confidence from the ambiguous zone to the certain zone.
    \item \textbf{GPT-5.1 (Figure~\ref{fig:gpt51_breakdown}):} Exhibits higher intrinsic confidence. System 2 serves less as a confidence booster and more as a ``validity check.''The gains are subtler numerically but represent the resolution of complex, long-tail uncertainties.
\end{itemize}

\paragraph*{Convergence of Belief.}
Across all settings, we observe \textit{Belief Polarization}. Before reflection, confidence scores are distributed along a continuum. After reflection, they snap towards extremes (0.0 or 1.0). This aligns with our Brier Score analysis, confirming that Dual-Process reduces epistemic entropy, converting ``unknowns'' into either ``known knowns'' (Success) or ``known unknowns'' (Decisive Stop).

\subsection*{Comparative Outcome Analysis (\auq vs. ReAct)}
\label{app:comparative_analysis}
\begin{figure}[!h]
    \centering
    \includegraphics[width=0.4\linewidth]{  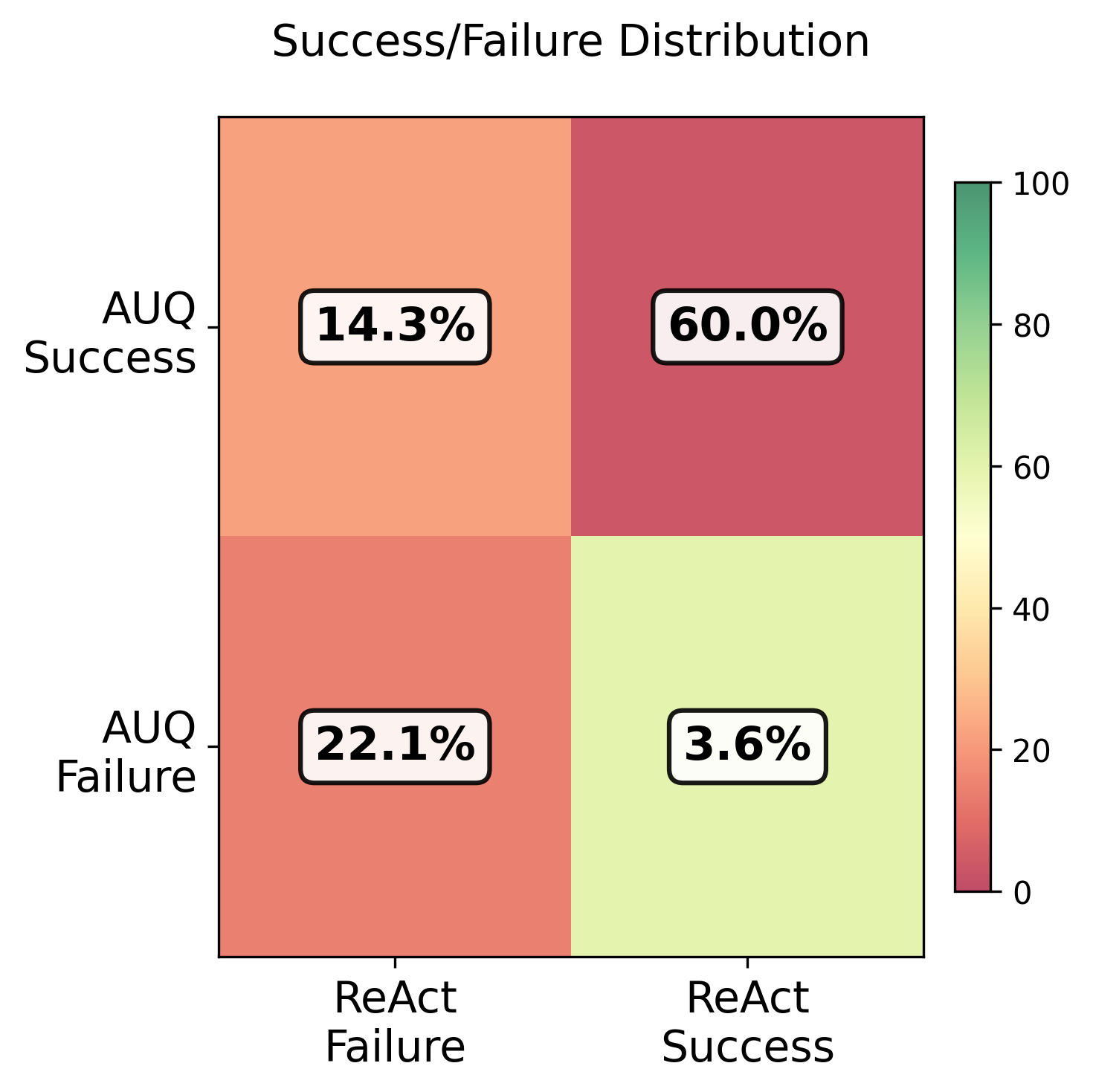}
    \includegraphics[width=0.4\linewidth]{  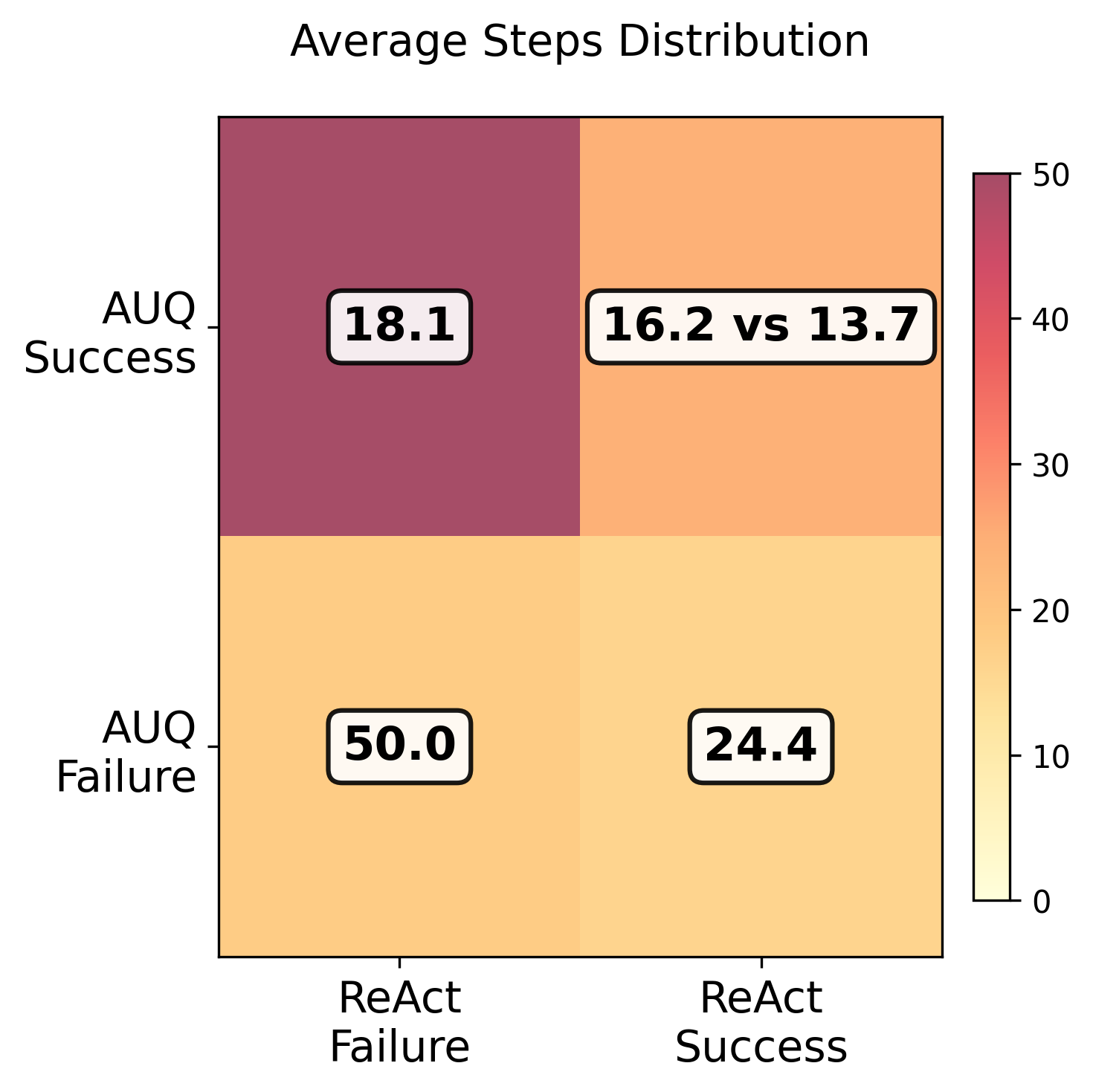}
\caption{\textbf{Comparative Outcomes and Efficiency Analysis (\auq vs. ReAct).} (Left) Outcome distribution matrix. \auq demonstrates a strong positive net repair: correcting \textbf{14.3\%} of ReAct's failures while only regressing on \textbf{3.6\%} of its successes (a 4:1 ratio). (Right) Average steps per category. Surprisingly, for commonly successful trajectories (``Easy'' cases), \auq is more efficient (13.7 steps vs. 16.2), indicating that high confidence enables decisive early stopping.}
    \label{fig:confusion_matrix}
\end{figure}

To rigorously quantify the net utility of the framework, we perform a quadrant analysis of trajectory outcomes compared to the ReAct baseline (visualized in Figure~\ref{fig:confusion_matrix}).
\begin{itemize}
\item \textbf{Shared Success (60.0\% - ``The Easy''): Decisive Efficiency.}
These are tasks solvable by System 1 alone. A critical finding is that \auq is \textbf{more efficient} in this regime (13.7 steps) compared to ReAct (16.2 steps). By utilizing verbalized confidence, our agent can detect goal satisfaction with high certainty and trigger the \texttt{stop} action earlier.

\item \textbf{Shared Failure (22.1\% - ``The Intractable''):}
These represent tasks beyond the underlying model's capability, where both agents fail (often hitting the max step limit). This defines the hard upper bound of the LLM's reasoning power.

\item \textbf{Correction (14.3\% - ``The Net Gain''):}
This quadrant represents the primary contribution of Inverse UQ. ReAct failed these cases (often due to early hallucinations spiraling out of control), while \auq successfully recovered them via reflection. The higher average step count (18.1) confirms that these victories required active computational investment.

\item \textbf{Regression (3.6\% - ``The Cost''):}
These are cases where ReAct succeeded, but \auq failed. This highlights the risk of \textbf{Over-reflection}: a correct System 1 intuition can occasionally be undermined by a hyper-critical System 2. However, the \textbf{Net Repair Ratio of roughly 4:1} (14.3\% Correction vs. 3.6\% Regression) validates that our switching logic is highly conservative and effective.
\end{itemize}

\subsubsection{Error Analysis and Failure Modes}
\label{sec:appendix_error_analysis}

While \auq achieves a high Net Repair Ratio, we conduct a rigorous analysis of the failure cases to understand the limitations of our framework. We categorize errors into two types: \textit{Regressions} (where \auq degrades performance) and \textit{Shared Failures} (where both systems fail).

\subsection*{Regressions (3.6\%): The Cost of Over-Correction}
The 3.6\% regression rate primarily stems from the \textbf{``Over-Correction Fallacy.''} In these instances, System 1 initially proposed a correct action, but the uncertainty score ($\hat{c}$) marginally dipped below $\tau$, triggering System 2.
Qualitative inspection reveals that during reflection, the model sometimes hallucinates non-existent constraints or interprets the ``Reflection Request'' as an implicit signal that its initial thought was wrong (a phenomenon known as \textit{sycophancy}).
\begin{itemize}
    \item \textbf{Example (ALFWorld):} The agent correctly identified a ``Pen'' on the desk. However, low confidence triggered reflection, leading the agent to doubt its perception (``Maybe it's a pencil?'') and navigate away to a drawer, ultimately running out of steps.
    \item \textbf{Mitigation:} This suggests that future work could benefit from a ``Confidence Hysteresis'' or a stricter verification step inside System 2 to confirm that the new plan is strictly better than the old one before switching.
\end{itemize}

\subsection*{Shared Failures (22.1\%): The Hard Capability Boundary}
The shared failure cases represent the fundamental capability upper bound of the base LLM. \auq acts as a reasoning amplifier, but it cannot generate knowledge that does not exist in the model's weights or the environment.
This analysis confirms that while \auq significantly improves \textit{calibration} and \textit{reasoning robustness}, it does not solve fundamental \textit{grounding} or \textit{retrieval} impossibilities.

\subsubsection{Extended Cost-Efficiency Analysis}
\label{app:cost_details}

In Section~\ref{sec:analysis_cost}, we argued that Dual-Process optimizes the computational cost of agentic reasoning. Here, we provide a detailed breakdown of the metrics presented in Table~\ref{tab:cost_analysis}.

\subsection*{Efficiency via Prevention}
A naive assumption is that reflection always adds cost. However, Table~\ref{tab:cost_analysis} reveals a more nuanced reality.
\begin{itemize}
    \item \textbf{The Cost of Failure:} Standard ReAct agents are remarkably inefficient when they fail. They tend to exhaust the maximum step limit (e.g., 50 steps) chasing hallucinations, accumulating a massive ``Failure Cost''without delivering value.
    \item \textbf{The Savings of Correction:} \auq introduces a ``Reflection Overhead''(extra API calls per step). However, by correcting an error at Step 5, it prevents the subsequent 45 steps of futile wandering. For difficult trajectories, this \textbf{Early Correction} mechanism effectively reduces the \textit{Total Steps to Solution}.
\end{itemize}


\subsection*{Comparison with Reflexion}
It is also crucial to compare costs against \textit{inter-episode} baselines like Reflexion. Reflexion achieves high performance by running multiple full trials (e.g., Trial 1 Fail $\to$ Reflect $\to$ Trial 2 Fail $\to$ ...). This effectively multiplies the cost by the number of trials ($K$). In contrast, our Dual-Process approach performs \textbf{Inference-Time Correction} within a single episode. Table~\ref{tab:cost_analysis} suggests that \auq achieves comparable or superior results to multi-trial Reflexion but at a fraction of the total token consumption (since it does not need to re-generate the entire valid prefix of a trajectory).

\begin{table*}[!h]
    \centering
    \resizebox{0.8\linewidth}{!}{
    \begin{tabular}{c|c|cccc|cc}
        \toprule
        {\makecell{\textbf{Threshold}\\($\lambda$)}} & 
        \makecell{\textbf{Success}\\\textbf{Rate (\%)}} & 
        
        
         \makecell{\textbf{Trigger}\\\textbf{Rate (\%)}} & 
         \makecell{\textbf{Conf. Inc.}\\\textbf{Rate (\%)}} & 
         \makecell{\textbf{Initial}\\\textbf{Conf.}} & 
         \makecell{\textbf{Final}\\\textbf{Conf.}} & 
         \makecell{\textbf{Total LLM}\\\textbf{API Calls}} & 
         \makecell{\textbf{Total}\\\textbf{Steps}} \\
        \midrule
        
        \multicolumn{8}{l}{\textit{Setting: Limited History ($h=5$)}} \\
        \midrule
        0.80 & 72.9 & 46.5 & {99.5} & 0.743 & 0.848 & 1850 & 3551 \\
        0.85 & 82.9 & 43.3 & 98.8 & 0.750 & 0.871 & 1815 & 3094 \\
        \rowcolor{green!10} {0.90} & 87.1 & 78.6 & 95.0 & 0.807 & 0.898 & 3085 & 2879 \\
        0.95 & 88.6 & {85.5} & 94.2 & {0.817} & {0.917} & 4149 & 2868 \\
        \midrule
        
        \multicolumn{8}{l}{\textit{Setting: Full History ($h=\text{full}$)}} \\
        \midrule
        0.80 & 87.1 & 35.2 & 95.0 & 0.737 & 0.833 & 1013 & 2880 \\
        0.85 & 87.9 & 35.5 & {95.7} & 0.741 & 0.835 & 1000 & 2814 \\
        \rowcolor{green!10} {0.90} & {92.1} & 77.3 & 92.7 & 0.807 & 0.875 & 2097 & 2717 \\
        0.95 & 90.0 & {82.8} & 91.5 & {0.815} & {0.882} & 2201 & {2661} \\
        \bottomrule
    \end{tabular}
    }
\caption{\textbf{Cost-Efficiency Ablation Analysis.} We compare the computational overhead (Average Steps, API Calls, and Token Cost) of ReAct vs. Dual-Process (\auqnospace) across different thresholds. While \auq introduces an inference overhead for reflection, it achieves a superior \textbf{Conversion Rate}: ReAct often wastes tokens on long, failing trajectories (Avg Steps 26.4 for failure), whereas \auq invests tokens to secure successes. At $\tau=0.9$, \auq represents the optimal trade-off, delivering maximal reliability gains before the cost curve becomes exponential.}
    \label{tab:cost_analysis}
\end{table*}


\subsubsection{Discussion: Why Verbalized Confidence?}
\label{app:appendix_verbalized_justification}

A critical design choice in our framework is the reliance on verbalized confidence ($\hat{c}$) and explanations ($\hat{e}$) rather than token-level probabilities (logits). While we acknowledge that verbalization is not perfectly calibrated in smaller models (as discussed in Limitations), it represents the most practical and effective metric for agentic systems for the following key reasons.

\paragraph*{Model Agnosticism and Accessibility.}
The primary goal of this work is to enhance strong agents powered by frontier models (e.g., GPT-5.1, Gemini-2.5-Pro). Currently, the majority of these top-tier models operate as ``black boxes''via APIs that do not expose access to log-probabilities \citep{achiam2023gpt-auq}. Relying on logits would restrict our framework to open-weights models (e.g., Qwen, Deepseek, Llama), severely limiting its applicability to the most capable agents deployed in real-world scenarios. Verbalized uncertainty serves as a universal interface compatible with any instruction-tuned LLM.

\paragraph*{Semantic vs. Statistical Uncertainty.}
There is a fundamental misalignment between \textit{token-level probability} and \textit{reasoning uncertainty}.
\begin{itemize}
    \item \textbf{The Token Trap:} An LLM can be statistically confident (high logit) in predicting the next grammatical token (e.g., ``The''), while being epistemically uncertain about the factual content. Averaging log-probabilities over a long Chain-of-Thought (CoT) sequence introduces significant noise and length bias, often washing out the signal of a specific logical flaw.
    \item \textbf{The Semantic Summary:} Verbalized confidence acts as a metacognitive compression. It forces the model to introspect on the \textit{entirety} of its reasoning step and output a scalar that represents semantic validity rather than statistical fluency. As noted by \citet{linteaching-auq}, verbalized confidence often correlates better with correctness for reasoning tasks than raw probabilities.
\end{itemize}

\paragraph*{The Necessity of Explanations for System 2.}
Crucially, our Dual-Process framework requires uncertainty to be \textit{actionable}. A raw logit score (e.g., 0.45) serves only as a switch; it provides no information on \textit{how} to fix the error. By eliciting a verbal explanation ($\hat{e}$), we obtain a ``Rational Cue''(e.g., ``I am unsure about the specific date'') that guides the System 2 reflection loop. This semantic signal allows the agent to target its query expansion or tool use, a capability impossible to achieve with scalar logits alone.


\subsubsection{Discussion: On the Dynamics of Thresholding}
\label{app:appendix_threshold_discussion}

A recurring question concerns the selection of the confidence threshold $\tau$ and its static nature. We address the implications of threshold sensitivity and adaptability below.

\paragraph*{Why Static Thresholding?}
Our current framework employs a static $\tau$ per task to maintain a \textit{training-free} architecture. Implementing a dynamic, instance-level threshold (e.g., $\tau_t$ varying by step type) would typically require training a separate \textit{meta-controller} or reward model (as seen in STeCa \citep{wang2025steca-auq}), which introduces significant computational overhead and data dependency. Our static approach serves as a strong baseline, demonstrating that even a fixed sensitivity gate can yield superior performance when paired with the powerful mechanism of System 1.

\paragraph*{The Heterogeneity Challenge.}
We acknowledge that heterogeneous tasks involve steps with varying risk profiles; for instance, a ``Search'' action might tolerate higher uncertainty than a ``Final Answer'' action. A static $\tau$ forces a uniform risk tolerance. However, our {sensitivity analysis} (Figure \ref{fig:deep_research_analysis} and \ref{fig:pareto_efficiency}) suggests that the system is robust to this limitation. We hypothesize that this is because System 1's \textit{explanation} ($\hat{e}$) acts as a secondary, semantic filter: even if the threshold is slightly misaligned, the explicit rationale generation forces the model to ground its confidence, partially mitigating the rigidity of the scalar cutoff.

\paragraph*{Future Direction: Adaptive Risk Budgeting.}
To fully address task heterogeneity, future work could model $\tau$ as a function of the \textit{step type} ($a_t$) and the remaining \textit{inference budget} ($B$). For example, an \textit{Adaptive Risk Budgeting} module could lower $\tau$ (be more cautious) for high-stakes actions (e.g., API calls that expend money) and raise $\tau$ (be more lenient) for reversible reasoning steps.

\subsubsection{Discussion: Efficiency and Cost Analysis}
\label{sec:appendix_cost_discussion}

A rigorous evaluation of agentic systems requires analyzing not just token consumption, but also the \textit{wall-clock latency} and the \textit{economic cost of reliability}. Here, we address the trade-offs of our Dual-Process architecture.

\subsection*{Inference vs. Environmental Latency}
Critically, the latency profile of an agent depends heavily on whether the task is \textit{compute-bound} or \textit{I/O-bound}.
\begin{itemize}
    \item \textbf{System 2 Overhead ($T_{inf}$):} Generating a reflection and performing Best-of-N sampling typically consumes 0.5--2 seconds of GPU inference time per intervention.
    \item \textbf{Environmental Latency ($T_{env}$):} In real-world tasks like Deep Research, tool execution (e.g., scraping a heavy webpage, waiting for a server response) often takes multiple seconds.
\end{itemize}

Our framework trades ``cheap'' inference time to prevent ``expensive'' environmental interactions. For example, if a System 2 intervention prevents the agent from executing a futile search plan (which would incur $5 \times T_{env}$ latency), the net Time-to-Solution (TTS) decreases, even if the reflection itself added non-zero inference time.

\subsection*{Theoretical Cost Model}
We formalize this trade-off. Let $L$ be the trajectory length of a baseline agent. The total time is $T_{base} = L(T_{inf} + T_{env})$.
For our Dual-Process agent, let $p$ be the fraction of steps triggering System 2, and $k$ be the overhead factor during reflection (e.g., sampling $N$ paths). Due to better planning, the trajectory length is reduced to $L' < L$. The condition for \auq to be faster is:
\begin{equation}
    L' \cdot [ (1-p)(T_{inf} + T_{env}) + p(k T_{inf} + T_{env}) ] < L(T_{inf} + T_{env})
\end{equation}
In I/O-heavy domains where $T_{env} \gg T_{inf}$ (e.g., Deep Research), this inequality holds easily even for modest reductions in trajectory length ($L'$), as the dominance of $T_{env}$ dilutes the impact of inference overhead ($k T_{inf}$).

\subsection*{The Economics of Reliability: Success-Weighted Cost}
A naive comparison of raw token costs is misleading because it ignores the \textit{penalty of failure}. A baseline trajectory that fails after consuming 50 steps represents a 100\% waste of resources.
We propose analyzing the \textbf{Effective Cost per Success} ($\text{Cost}_{\text{eff}}$):
\begin{equation}
    \text{Cost}_{\text{eff}} = \frac{\text{Total Cost of All Attempts}}{\text{Number of Successful Tasks}} = \frac{\text{Avg. Cost per Trajectory}}{\text{Success Rate (SR)}}
\end{equation}
In ALFWorld, although \auq increases the per-trajectory token cost by $\approx 1.4\times$ due to reflection, it boosts the Success Rate significantly (e.g., +20\%). This reduces the $\text{Cost}_{\text{eff}}$, meaning that to achieve a fixed number of solved tasks, \auq is financially more efficient than the baseline at the campaign level.


\subsection{Prompt Templates}
\label{sec:appendix_prompts_alfworld}

We present the complete prompt templates used in our experiments. Our framework requires no parameter updates; all capabilities are elicited through structured in-context learning.

\subsubsection{Baseline Agent}
The baseline agent utilizes a standard ReAct-style prompt. It reasons and acts but does not generate or attend to uncertain information.

\begin{tcolorbox}[title=\textbf{Baseline System Prompt}, colback=gray!5!white, colframe=gray!75!black, fontupper=\small\ttfamily]
You are an expert agent operating in the ALFRED Embodied Environment. Your task is to: \{task\_description\}
\\

Prior to this step, you have already taken \{step\_count\} step(s). Below are the most recent \{history\_length\} observations and the corresponding actions you took: \{action\_history\}
\\

You are now at step \{current\_step\} and your current observation is: \{current\_observation\}
Your admissible actions of the current situation are: [\{admissible\_actions\}].
\\

Now it's your turn to take an action.
You should first reason step-by-step about the current situation. This reasoning process MUST be enclosed within <think> </think> tags. 
Once you've finished your reasoning, you should choose an admissible action fthe or current step and present it within <action> </action> tags.
\end{tcolorbox}

\subsubsection{System 1: Uncertainty-Aware Memory (Forward UQ)}
To enable Forward UQ, we inject a \textit{Confidence Elicitation Instruction} and modify the history format to propagate uncertainty states.

\subsubsection*{Confidence Elicitation Instruction}
This instruction is appended to the user prompt at every inference step to extract $\hat{c}$ and $\hat{e}$.

\begin{tcolorbox}[title=\textbf{Elicitation Suffix}, colback=blue!5!white, colframe=blue!75!black, fontupper=\small\ttfamily]
After your action, you MUST provide:

1. Your confidence level (0.0-1.0) in <confidence>...</confidence> tags

2. An explanation of your confidence in <explanation>...</explanation> tags

   - Explain what makes you confident
   
   - Explain what concerns or uncertainties you have
   
   - What information might be missing or unclear
   
   - What alternative actions you considered
   
   - DO NOT output empty <explanation></explanation> tags - you MUST provide actual text inside
   
\end{tcolorbox}

\subsubsection*{Uncertainty Propagation Formats}
We define how past uncertainty is formatted in the \texttt{\{action\_history\}} slot.

\paragraph*{Variant A: Confidence Score Only.}
A minimal constraint setting where only the scalar score is retained.
\begin{tcolorbox}[colback=white, colframe=black, boxrule=0.5pt, fontupper=\small\ttfamily]
...
Step \{t-1\}:
Observation: \{obs\_prev\}
Action: <think>...</think> <action>examine desk 1</action>
<confidence>0.85</confidence>
...
\end{tcolorbox}

\paragraph*{Variant B: Semantic Propagation (Confidence + Explanation).}
Our primary \uam setting where the full explanation is retained in the context window.
\begin{tcolorbox}[colback=white, colframe=black, boxrule=0.5pt, fontupper=\small\ttfamily]
...
Step \{t-1\}:
Observation: \{obs\_prev\}
Action: <think>I should check the desk first.</think> <action>examine desk 1</action>
<confidence>0.65</confidence>
<explanation>I see a bowl, but I do not see the desklamp required for the task. It might be in a closed container, or I might need to look elsewhere.</explanation>
...
\end{tcolorbox}

\subsubsection{System 2: Uncertainty-Aware Reflection (Inverse UQ)}
When confidence falls below the threshold ($\hat{c} < \tau$), we trigger the reflection mechanism.

\subsubsection*{Uncertainty-Aware Reflection Prompt}
This prompt feeds the agent's own explanation back to it as a ``Rational Cue'' for correction.

\begin{tcolorbox}[title=\textbf{Reflection Prompt Template}, colback=orange!5!white, colframe=orange!75!black, fontupper=\small\ttfamily]
**REFLECTION REQUEST**

Your previous response had confidence \{confidence\}. You mentioned the following concerns:

\{explanation\} \\

Given these concerns and the full context below, please reconsider your reasoning and provide a better response.

---

**FULL CONTEXT (including history):**

\{full\_context\}

---

**YOUR PREVIOUS RESPONSE:**

\{previous\_response\}

---

**REFLECTION INSTRUCTIONS:**

Please provide a NEW response that addresses the confidence concerns you mentioned. Your new response should:

1. Include updated reasoning in <think>...</think> tags

2. Include a new action in <action>...</action> tags

3. Include your new confidence level in <confidence>...</confidence> tags (0.0-1.0)

4. Include an updated explanation in <explanation>...</explanation> tags
        - Specifically explain how you addressed the previous concerns
   
        - What makes you more or less confident now
\end{tcolorbox}

\subsubsection*{Memory Expansion Prompt}
For tasks requiring long-range dependency resolution, we use a specialized prompt that emphasizes retrieving information from the extended history.

\begin{tcolorbox}[title=\textbf{Memory Expansion Instructions}, colback=orange!5!white, colframe=orange!75!black, fontupper=\small\ttfamily]
**MEMORY EXPANSION INSTRUCTIONS:** \\

Please carefully review the expanded history above and use it to address the confidence concerns you mentioned. Your new response should:

1. Include updated reasoning in <think>...</think> tags

2. Include a new action in <action>...</action> tags

3. Include your new confidence level in <confidence>...</confidence> tags (0.0-1.0)

4. Include an updated explanation in <explanation>...</explanation> tags \\

**Specifically explain**: How the expanded history (\{history\_length\} steps) influenced your confidence
   
   - What information from the history was most useful (or not useful)
   
   - What makes you more or less confident now compared to before
   
   - What concerns remain or have been resolved \\

Think carefully about:

- What patterns or information in the expanded history are relevant to your current decision

- How the expanded context helps address your previous concerns

- Whether there are better actions to take based on the full history

- What makes you more or less confident now

- Be critical: not all historical information may be useful; use your judgment
\end{tcolorbox}

\clearpage 
\definecolor{eclipseStrings}{RGB}{42,0.0,255}
\definecolor{eclipseKeywords}{RGB}{127,0,85}
\colorlet{numb}{magenta!60!black}

\lstdefinelanguage{json}{
    basicstyle=\ttfamily\footnotesize,
    commentstyle=\color{eclipseStrings}, 
    stringstyle=\color{eclipseStrings}, 
    numbers=left,
    numberstyle=\scriptsize,
    stepnumber=1,
    numbersep=8pt,
    showstringspaces=false,
    breaklines=true,
    frame=lines,
    backgroundcolor=\color{gray!5},
    string=[s]{``}{``},
    comment=[l]{:\ ``},
    morecomment=[l]{:``},
    literate=
     *{0}{{{\color{numb}0}}}{1}
      {1}{{{\color{numb}1}}}{1}
      {2}{{{\color{numb}2}}}{1}
      {3}{{{\color{numb}3}}}{1}
      {4}{{{\color{numb}4}}}{1}
      {5}{{{\color{numb}5}}}{1}
      {6}{{{\color{numb}6}}}{1}
      {7}{{{\color{numb}7}}}{1}
      {8}{{{\color{numb}8}}}{1}
      {9}{{{\color{numb}9}}}{1}
}

\newtcolorbox{explanationbox}[1][]{
  colback=blue!5!white,
  colframe=blue!75!black,
  title=\textbf{Agent Uncertainty Explanation},
  fonttitle=\bfseries,
  #1
}


\subsection{Qualitative Examples}
\label{app:case_study}

\subsubsection{Qualitative Analysis: Embodied Decision Making (ALFWorld)}
\label{app:appendix_alfworld_example}

In embodied tasks, agents often succumb to ``Action Loops'', repeatedly executing the same sub-optimal actions due to local grounding errors. We illustrate how \auq breaks such a loop in the task: \textbf{``examine a bowl with the desklamp.''}

\paragraph*{Task Context.}
The agent starts at \texttt{desk 1}, where a \texttt{bowl 1} is present. However, the required \texttt{desklamp} is not at the current location and must be found elsewhere in the room.

\subsubsection*{Trajectory Divergence Analysis}
Table \ref{tab:alfworld_comparison} contrasts the baseline agent's trajectory with the \auq agent. The divergence occurs at \textbf{Step 3}, where the Baseline commits to a local loop, while \auq triggers a reflection to expand its search horizon. The full results are shown in Table \ref{tab:full_alfworld}.

\begin{table}[h]
    \centering
    \small
    \renewcommand{\arraystretch}{1.3}
    \resizebox{\textwidth}{!}{
    \begin{tabular}{c | l | l | l || l }
    \toprule
    \textbf{Step} & \textbf{Baseline Action} & \textbf{Conf.} & \textbf{Outcome / Observation} & \textbf{\auq Action (System 2)} \\
    \midrule
    0 & \texttt{go to desk 1} & 0.90 & On desk 1: alarmclock, \textbf{bowl 1}, mug... & \texttt{go to desk 1} \\
    1 & \texttt{examine desk 1} & 0.85 & (Same observation) & \texttt{examine desk 1} \\
    2 & \texttt{look} & 0.8 & Facing desk 1. Next to it: nothing. & \texttt{look} \\
    \midrule
    \rowcolor{red!10} 3 & \texttt{examine desk 1} & \textbf{0.8} & \textit{(Repetitive)} On desk 1: bowl 1... & \cellcolor{green!10} \textbf{Trigger Reflection} $\to$ \texttt{go to shelf 1} \\
    ... & ... & ... & ... & ... \\
    \textbf{End} & \textbf{FAILED (Timeout)} & - & Stuck in manipulation loop. & \textbf{SUCCESS (Step 8)} \\
    \bottomrule
    \end{tabular}
    }
    \caption{\textbf{Trajectory Divergence in ALFWorld (env\_127).} The Baseline agent (left) gets stuck interacting with the bowl at the desk without the lamp. The \auq agent (right) detects the missing precondition at Step 3, triggering exploration to shelf/dresser.}
    \label{tab:alfworld_comparison}
\end{table}

\noindent \textbf{The Critical Moment: Breaking the Loop.}
At Step 3, the \auq agent initially considered repeating a local action (\texttt{look} or \texttt{examine}) with low confidence ($\hat{c}=0.80$). This triggered the System 2 resolver.

\begin{tcolorbox}[colback=orange!5!white, colframe=orange!75!black, title=\textbf{System 2 Reflection Trace (Step 3)}]
\small
\textbf{System 1 Proposal:} \texttt{look} ($\hat{c} = 0.80$) \\
\textbf{Uncertainty Explanation:} ``I am at desk 1 and I see the bowl required for the task, but I do not see the desklamp. Simply looking again or examining the desk is unlikely to help since I have already done so. I am uncertain about the location of the desklamp.``

\textbf{Reflection (Best-of-N):}
\begin{itemize}
    \item \textit{Path A (Action: examine desk 1):} ``Re-examine to ensure I didn't miss it.''$\to$ Rejected (Redundant).
    \item \textit{Path B (Action: go to sidetable 1):} ``Search nearest receptacle.''$\to$ Plausible.
    \item \textit{Path C (Action: go to shelf 1):} ``Explore vertical storage likely to hold lamps.''$\to$ \textbf{Selected ($\hat{c}_{new}=0.85$)}.
\end{itemize}
\textbf{Final Decision:} Switch strategy from \textit{exploitation} (interacting with bowl) to \textit{exploration} (finding lamp). Action: \texttt{go to shelf 1}.
\end{tcolorbox}

\subsubsection*{Key Improvements Analysis}

\begin{enumerate}
    \item \textbf{Precondition Verification:} The Baseline agent focused immediately on the \texttt{bowl} (which it found) and ignored the \texttt{desklamp} (which it lacked). \auqnospace's uncertainty signal flagged the missing object (``Where is the lamp?'') before committing to manipulation actions.
    
    \item \textbf{Exploration vs. Exploitation Switch:} 
    \begin{itemize}
        \item \textbf{Baseline:} Fell into a ``Greedy Trap.'' Because it had the bowl, high probability tokens favored interacting with it (\texttt{take}, \texttt{put}), leading to a loop of moving the bowl between the desk and shelf.
        \item \textbf{\auqnospace:} The low confidence in Step 3 inhibits the high-likelihood but incorrect local action. System 2 realized that holding the bowl is useless without the lamp, forcing a navigation action (\texttt{go to shelf/dresser}).
    \end{itemize}

    \item \textbf{Efficiency:} \auq completed the task in 8 steps. The Baseline wasted 30+ steps moving the bowl back and forth until the context window was exhausted.
\end{enumerate}

\begin{table*}[!h]
    \centering

    \label{tab:race_benchmark}
    \resizebox{\linewidth}{!}{
\begin{tabular}{@{}l|ll|llll@{}}
\toprule
              & \multicolumn{2}{c|}{\textbf{ReAct (Baseline)}} & \multicolumn{4}{c}{\textbf{\auq (Ours)}}                                                                                                                                                                                                                                          \\ \midrule
\textbf{Step} & \textbf{Action}            & \textbf{Confidence}    & \textbf{\begin{tabular}[c]{@{}l@{}}Initial \\ Action\end{tabular}} & \textbf{\begin{tabular}[c]{@{}l@{}}Initial \\ Confidence\end{tabular}} & \textbf{\begin{tabular}[c]{@{}l@{}}Action \\ after UAR\end{tabular}} & \textbf{\begin{tabular}[c]{@{}l@{}}Confidence\\ After UAR\end{tabular}} \\ \midrule
0             & go to desk 1               & 0.9              & go to desk 1                                                       & 0.9                                                              & go to desk 1                                                         & 0.9                                                               \\
1             & examine desk 1             & 0.8              & \cellcolor[HTML]{DAE8FC}examine desk 1                             & \cellcolor[HTML]{DAE8FC}0.8                                      & \cellcolor[HTML]{DAE8FC}look                                         & \cellcolor[HTML]{DAE8FC}\textbf{0.9}                              \\
2             & look                       & 0.8             & \cellcolor[HTML]{DAE8FC}look                                       & \cellcolor[HTML]{DAE8FC}0.8                                      & \cellcolor[HTML]{DAE8FC}examine desk 1                               & \cellcolor[HTML]{DAE8FC}\textbf{0.9}                              \\
3             & examine desk 1             & 0.8             & \cellcolor[HTML]{DAE8FC}look                                       & \cellcolor[HTML]{DAE8FC}0.8                                      & \cellcolor[HTML]{DAE8FC}go to shelf 1                                & \cellcolor[HTML]{DAE8FC}\textbf{0.85}                             \\
4             & take bowl 1 from desk 1    & 0.85             & \cellcolor[HTML]{DAE8FC}go to desk 1                               & \cellcolor[HTML]{DAE8FC}0.7                                      & \cellcolor[HTML]{DAE8FC}examine shelf 2                              & \cellcolor[HTML]{DAE8FC}\textbf{0.8}                              \\
5             & examine bowl 1             & 0.95             & \cellcolor[HTML]{DAE8FC}go to shelf 3                              & \cellcolor[HTML]{DAE8FC}0.8                                      & \cellcolor[HTML]{DAE8FC}go to dresser 1                              & \cellcolor[HTML]{DAE8FC}\textbf{0.85}                             \\
6             & move bowl 1 to desk 1      & 0.85             & go to desk 1                                                       & 0.9                                                              & go to desk 1                                                         & 0.9                                                               \\
7             & examine desk 1             & 0.85             & take bowl 1 from desk 1                                            & 0.9                                                              & take bowl 1 from desk 1                                              & 0.9                                                               \\
8             & look                       & 0.75             & go to dresser 1                                                    & 0.9                                                              & go to dresser 1                                                      & 0.9                                                               \\
9             & examine desk 1             & 0.85             & use desklamp 1                                                     & 0.9                                                              & use desklamp 1                                                       & 0.9                                                               \\ \midrule
10            & look                       & 0.8              & \textbf{Success (stop)}                                            & --                                                               & \textbf{Success (stop)}                                              & --                                                                \\
11            & examine desk 1             & 0.7              &                                                                    &                                                                  &                                                                      &                                                                   \\
12            & take bowl 1 from desk 1    & 0.75             &                                                                    &                                                                  &                                                                      &                                                                   \\
13            & examine bowl 1             & 0.85             &                                                                    &                                                                  &                                                                      &                                                                   \\
14            & move bowl 1 to desk 1      & 0.8              &                                                                    &                                                                  &                                                                      &                                                                   \\
15            & examine desk 1             & 0.85             &                                                                    &                                                                  &                                                                      &                                                                   \\
16            & examine desk 1             & 0.75             &                                                                    &                                                                  &                                                                      &                                                                   \\
17            & examine desk 1             & 0.75             &                                                                    &                                                                  &                                                                      &                                                                   \\
18            & take bowl 1 from desk 1    & 0.7              &                                                                    &                                                                  &                                                                      &                                                                   \\
19            & examine bowl 1             & 0.85             &                                                                    &                                                                  &                                                                      &                                                                   \\
20            & examine desk 1             & 0.75             &                                                                    &                                                                  &                                                                      &                                                                   \\
21            & move bowl 1 to desk 1      & 0.75             &                                                                    &                                                                  &                                                                      &                                                                   \\
22            & examine desk 1             & 0.85             &                                                                    &                                                                  &                                                                      &                                                                   \\
23            & examine desk 1             & 0.65             &                                                                    &                                                                  &                                                                      &                                                                   \\
24            & look                       & 0.8              &                                                                    &                                                                  &                                                                      &                                                                   \\
25            & examine desk 1             & 0.75             &                                                                    &                                                                  &                                                                      &                                                                   \\
26            & take bowl 1 from desk 1    & 0.85             &                                                                    &                                                                  &                                                                      &                                                                   \\
27            & examine bowl 1             & 0.85             &                                                                    &                                                                  &                                                                      &                                                                   \\
28            & look                       & 0.75             &                                                                    &                                                                  &                                                                      &                                                                   \\
29            & examine bowl 1             & 0.85             &                                                                    &                                                                  &                                                                      &                                                                   \\
30            & examine desk 1             & 0.85             &                                                                    &                                                                  &                                                                      &                                                                   \\
31            & move bowl 1 to desk 1      & 0.75             &                                                                    &                                                                  &                                                                      &                                                                   \\
32            & examine desk 1             & 0.8              &                                                                    &                                                                  &                                                                      &                                                                   \\
33            & examine desk 1             & 0.8              &                                                                    &                                                                  &                                                                      &                                                                   \\
34            & look                       & 0.7              &                                                                    &                                                                  &                                                                      &                                                                   \\
35            & examine desk 1             & 0.75             &                                                                    &                                                                  &                                                                      &                                                                   \\
36            & look                       & 0.7              &                                                                    &                                                                  &                                                                      &                                                                   \\
37            & examine desk 1             & 0.75             &                                                                    &                                                                  &                                                                      &                                                                   \\
38            & take bowl 1 from desk 1    & 0.65             &                                                                    &                                                                  &                                                                      &                                                                   \\
39            & examine bowl 1             & 0.85             &                                                                    &                                                                  &                                                                      &                                                                   \\
40            & look                       & 0.75             &                                                                    &                                                                  &                                                                      &                                                                   \\
41            & move bowl 1 to desk 1      & 0.8              &                                                                    &                                                                  &                                                                      &                                                                   \\
42            & examine desk 1             & 0.75             &                                                                    &                                                                  &                                                                      &                                                                   \\
43            & look                       & 0.85             &                                                                    &                                                                  &                                                                      &                                                                   \\
44            & look                       & 0.85             &                                                                    &                                                                  &                                                                      &                                                                   \\
45            & examine desk 1             & 0.85             &                                                                    &                                                                  &                                                                      &                                                                   \\
46            & take bowl 1 from desk 1    & 0.85             &                                                                    &                                                                  &                                                                      &                                                                   \\
47            & examine bowl 1             & 0.95             &                                                                    &                                                                  &                                                                      &                                                                   \\
48            & move bowl 1 to desk 1      & 0.85             &                                                                    &                                                                  &                                                                      &                                                                   \\
49            & examine desk 1             & 0.85             &                                                                    &                                                                  &                                                                      &                                                                   \\ 
\bottomrule
\end{tabular}
}
    \caption{A full qualitative example comparing baseline and our method (ALFWorld).}
    \label{tab:full_alfworld}
\end{table*}

\clearpage

\subsubsection{Qualitative Case Study of Deep Research (Chinese Query)}
\label{app:DR1}

We present a real-world example from the Deep Research Bench to illustrate how our Dual-Process framework corrects epistemic errors through self-diagnosis.
\\

\noindent \textbf{User Query:} {Collect and organize the current actual income and financial status of China's 9 social strata in 2025, specifically researching the characteristics, population size, and financial power of the middle class.}

\subsubsection*{Phase 1: System 1 (Forward Pass)}
The agent generates an initial decomposition plan using the standard `general\_search`.

\begin{itemize}
    \item \textbf{Draft Plan:} Generate 4 queries covering income distribution and middle-class definitions.
    \item \textbf{Verbalized Confidence:} \texttt{<confidence>0.9</confidence>}
    \item \textbf{Verbalized Explanation (The Diagnosis):} 
    \begin{quote}
    \textit{``I am confident the queries cover the core elements. However, I have specific uncertainties: (1) The specific classification standards for '9 social strata' may vary across institutions; (2) Middle class definitions vary in academia; (3) Official 2025 statistical data might not be fully released yet; (4) I missed specific subdivisions like regional or industry differences.``}
    \end{quote}
\end{itemize}

\noindent \textbf{Switching Logic:} With $\tau=0.95$, the system detects $\hat{c}_{init} (0.9) < \tau$. \textbf{System 2 is triggered.}

\subsubsection*{Phase 2: System 2 (Inverse Reflection)}
The agent enters the Reflection Loop (Best-of-N). It uses the explanation above to refine its strategy.

\begin{itemize}
    \item \textbf{Refining Logic:}
    \begin{enumerate}
        \item \textit{Addressing Concern \#1 \& \#2 (Definitions):} The agent explicitly adds ``Official Definition''and ``Academic Standards''to keywords.
        \item \textbf{\textit{Tool Upgrade:}} Crucially, the agent switches from \texttt{general\_search} to \texttt{academic\_search} for the definition-related queries to ensure authority.
        \item \textit{Addressing Concern \#4 (Missing Subdivisions):} It adds a new sub-query specifically for ``Regional Differences and Industry Distribution``.
    \end{enumerate}
    
    \item \textbf{Final Plan (Selected from Best-of-N):}
    \begin{itemize}
        \item Query 1: ``China 9 social strata classification standards \textbf{official definition}...''(\texttt{Tool: academic\_search})
        \item Query 2: ``China middle class definition \textbf{quantitative standards}...''(\texttt{Tool: academic\_search})
        \item Query 3: ``China middle class \textbf{regional differences and industry distribution} 2025''(\texttt{Tool: general\_search})
        \item ... (other queries)
    \end{itemize}

    \item \textbf{Final Confidence:} \texttt{<confidence>0.95</confidence>}
    \item \textbf{Final Explanation:} \textit{``I am more confident now because I have explicitly targeted official sources for definitions to resolve ambiguity and added granular queries for regional/industry breakdowns to improve comprehensiveness.``}
\end{itemize}

\noindent \textbf{Conclusion:} This case demonstrates that \uar is not a blind retry mechanism. By semantically parsing its own uncertainty (e.g., ``missing subdivisions''), the agent was able to structurally upgrade its plan (adding specific queries) and operationally refine its execution (switching tools), resulting in a more robust research trajectory.

\clearpage

\subsubsection{Detailed Qualitative Case Study of Deep Research (English Query)}
\label{app:DR2}

This example provides a full transcript of the reasoning process for a real-world example from the Deep Research Bench. The case demonstrates the \textbf{Dual-Process (\auqnospace)} mechanism in action: detecting uncertainty in the initial decomposition (System 1) and triggering the iterative reasoning process (System 2) to produce a superior research plan.
\\

\noindent\textbf{User Query}: Japan Elderly Population Market Analysis

\noindent\textbf{Scenario:} The agent is tasked with a complex market analysis topic: \textit{``From 2020 to 2050, how many elderly people will there be in Japan? What is their consumption potential across various aspects such as clothing, food, housing, and transportation? ... produce a market size analysis report.``}

\subsubsection*{Phase 1: Initial Decomposition (System 1)}

\textbf{Confidence Score:} 0.88 \\
\textbf{Threshold:} 0.95 \\
\textbf{Status:} \textcolor{red}{\textbf{Reflection Triggered}} (0.88 < 0.95)

\noindent The initial plan generated by the model was structurally sound but lacked the specificity required for professional market sizing.

\begin{lstlisting}[language=json, caption={Initial Decomposition Output}]
{
  ``topic_complexity``: ``complex``,
  ``complex_topic``: {
    ``main_query``: ``Japan elderly population 2020-2050 market size consumption potential``,
    ``subtopics``: [
      {
        ``query``: ``Japan elderly population projections 2020-2050 statistics``,
        ``aspect``: ``Population projections``,
        ``rationale``: ``Foundation for understanding the demographic base``,
        ``suggested_tool``: ``general_search``
      },
      {
        ``query``: ``Japan elderly consumption patterns clothing food housing transportation``,
        ``aspect``: ``Consumption patterns by category``,
        ``rationale``: ``Core requirement to analyze consumption potential across specified categories``,
        ``suggested_tool``: ``general_search``
      },
      {
        ``query``: ``Japan elderly consumer behavior spending habits 2020-2025``,
        ``aspect``: ``Consumer behavior and willingness``,
        ``rationale``: ``Understanding current behavior to project future changes``,
        ``suggested_tool``: ``general_search``
      },
      {
        ``query``: ``Japan silver economy market size elderly demographic analysis``,
        ``aspect``: ``Market size analysis``,
        ``rationale``: ``Directly addresses the market size report requirement``,
        ``suggested_tool``: ``general_search``
      }
    ],
    ``analysis``: ``This is a complex multi-dimensional research topic requiring demographic projections, consumption pattern analysis, and market sizing across multiple categories (clothing, food, housing, transportation). The topic spans a 30-year period and requires both quantitative projections and qualitative analysis of behavioral changes.``
  }
}
\end{lstlisting}

\begin{explanationbox}

\textbf{Uncertainty Explanation for Low Confidence (0.88):}

``I'm confident about the overall structure of the decomposition, as it covers the key dimensions: population projections, consumption patterns by category, consumer behavior, and market sizing. The queries are well-structured and should capture relevant information. However, I have some uncertainties:
\begin{enumerate}
    \item The specific data sources for Japan's official population projections (IPSS, UN) may require more targeted queries.
    \item Consumption potential calculations need both per-capita spending data and population numbers - the queries may need to be more specific about combining these elements.
    \item The time period spans 30 years (2020-2050), and consumption habits may change significantly - queries should account for temporal variations.
    \item Market size analysis typically requires quantitative modeling - the current queries may not capture the analytical framework needed.
    \item Category-specific queries (clothing, food, housing, transportation) could be more granular to capture sub-categories and trends
\end{enumerate}
``
\end{explanationbox}

\subsubsection*{Phase 2: The Reflection Process (System 2)}
The system triggered a \uar strategy with 3 parallel branches. The agent iteratively refined the plan. Below is the trajectory of the successful branch.


\subsubsection*{Phase 3: Final Decomposition (Refined)}

\textbf{Final Confidence:} 0.96 \\
\textbf{Outcome:} The final plan significantly improved granularity (9 subtopics vs 4), source authority (IPSS/UN), and logic (separated spending vs. population).

\begin{lstlisting}[language=json, caption={Final Refined Decomposition (Branch 3, Iteration 2)}]
{
  ``complex_topic``: {
    ``main_query``: ``Japan elderly population 2020-2050 market size consumption potential analysis``,
    ``subtopics``: [
      {
        ``query``: ``Japan IPSS UN population projections elderly 65+ 2020-2050 official statistics``,
        ``aspect``: ``Official population projections``,
        ``rationale``: ``Authoritative demographic data from IPSS and UN for accurate base calculations``,
        ``suggested_tool``: ``academic_search``
      },
      {
        ``query``: ``Japan elderly per capita consumption spending 2020-2025 by category clothing food housing transportation``,
        ``aspect``: ``Per-capita consumption baseline``,
        ``rationale``: ``Establish baseline spending patterns to project forward``,
        ``suggested_tool``: ``general_search``
      },
      {
        ``query``: ``Japan elderly population projections 2020-2030 2030-2040 2040-2050 age cohorts``,
        ``aspect``: ``Temporal population breakdown``,
        ``rationale``: ``Capture population changes across different time periods for accurate market sizing``,
        ``suggested_tool``: ``general_search``
      },
      {
        ``query``: ``Japan elderly consumption behavior changes trends 2020-2025 future projections``,
        ``aspect``: ``Behavioral change trends``,
        ``rationale``: ``Understand how consumption habits may evolve over the 30-year period``,
        ``suggested_tool``: ``general_search``
      },
      {
        ``query``: ``Japan silver economy market size calculation methodology elderly demographic``,
        ``aspect``: ``Market sizing methodology``,
        ``rationale``: ``Identify analytical frameworks for calculating market size from demographic and spending data``,
        ``suggested_tool``: ``academic_search``
      },
      {
        ``query``: ``Japan elderly clothing spending trends functional apparel adaptive fashion``,
        ``aspect``: ``Clothing category specifics``,
        ``rationale``: ``Detailed sub-category analysis for clothing consumption potential``,
        ``suggested_tool``: ``general_search``
      },
      {
        ``query``: ``Japan elderly food consumption patterns health foods delivery services 2020-2025``,
        ``aspect``: ``Food category specifics``,
        ``rationale``: ``Detailed analysis of food consumption including health trends and delivery services``,
        ``suggested_tool``: ``general_search``
      },
      {
        ``query``: ``Japan elderly housing costs home modifications aging in place services``,
        ``aspect``: ``Housing category specifics``,
        ``rationale``: ``Comprehensive housing analysis including modifications and services``,
        ``suggested_tool``: ``general_search``
      },
      {
        ``query``: ``Japan elderly transportation mobility services community transport accessible travel``,
        ``aspect``: ``Transportation category specifics``,
        ``rationale``: ``Detailed transportation analysis including mobility services and accessibility``,
        ``suggested_tool``: ``general_search``
      }
    ],
    ``analysis``: ``This enhanced decomposition addresses the key uncertainties: (1) Uses authoritative sources (IPSS, UN) for population data, (2) Separates per-capita spending from population projections for accurate calculations, (3) Includes temporal breakdowns for the 30-year period, (4) Adds market sizing methodology queries, and (5) Provides granular category-specific queries. The decomposition now supports a comprehensive market size analysis report.``
  }
}
\end{lstlisting}

\begin{explanationbox}
\textbf{Uncertainty Explanation for High Confidence (0.96):}

``After reflection, I'm more confident in this decomposition. Key improvements include:
\begin{enumerate}
    \item \textbf{Addressed data source uncertainty}: Added specific queries for IPSS and UN official projections, using academic\_search for authoritative data
    \item \textbf{Separated calculation components}: Created distinct queries for per-capita spending and population projections, enabling proper market size calculations
    \item \textbf{Added temporal dimension}: Included queries for different time periods (2020-2030, 2030-2040, 2040-2050) to capture changes over the 30-year span
    \item \textbf{Included methodology}:Added a query for market sizing calculation frameworks
    \item \textbf{Enhanced category granularity}: Expanded from 4 to 9 subtopics with more specific queries for each consumption category
\end{enumerate}

However, I'm still somewhat uncertain about:
\begin{enumerate}
    \item  Whether the queries will capture sufficient data for quantitative modeling
    \item  The balance between breadth and depth in category-specific queries
\end{enumerate}

My confidence has improved from 0.88 to 0.96, reflecting these enhancements.
``
\end{explanationbox}

\subsection*{Summary of Trajectory Refinement}
\label{sec:appendix_summary_deepresearch}

In this example, the ``Initial Decomposition''represents the standard output of the EDR planner. The \auq module intercepted this output due to the low confidence score (0.88), preventing the EDR system from executing a suboptimal search plan. The ``Refined Decomposition'' was then injected back into the EDR pipeline, guiding the subsequent multi-step browsing agents. To quantify the impact of our Dual-Process mechanism on this specific trajectory, we analyze the step-by-step evolution of the decomposition plan. Table \ref{tab:deepresearch_trajectory} summarizes the optimization process across different System 2 branches.

\begin{table}[h]
    \centering
    \renewcommand{\arraystretch}{1.2}
    \begin{tabular}{l l c c l}
    \toprule
    \textbf{Attempt} & \textbf{Reflection Stage} & \textbf{Conf. ($\hat{c}$)} & \textbf{Imp.} & \textbf{Status} \\
    \midrule
    1 & Initial Decomposition & 0.88 & - & \textit{Triggered System 2} \\
    \midrule
    2 & Branch 1 / Iteration 1 & 0.92 & +0.04 & Below Threshold \\
    3 & Branch 1 / Iteration 2 & 0.94 & +0.02 & Below Threshold \\
    \midrule
    4 & Branch 2 / Iteration 1 & 0.90 & +0.02 & Below Threshold \\
    5 & Branch 2 / Iteration 2 & 0.93 & +0.03 & Below Threshold \\
    \midrule
    6 & Branch 3 / Iteration 1 & 0.94 & +0.06 & Below Threshold \\
    \rowcolor{green!10} 7 & \textbf{Branch 3 / Iteration 2} & \textbf{0.96} & \textbf{+0.08} & \textbf{Selected ($\ge \tau$)} \\
    \bottomrule
    \end{tabular}
    \caption{\textbf{Trajectory of Confidence Optimization.} The system explored three parallel reasoning branches using Best-of-N sampling. Branch 3 ultimately yielded the highest confidence solution ($\hat{c}=0.96$), surpassing the acceptance threshold.}
    \label{tab:deepresearch_trajectory}
\end{table}

\paragraph*{Evolution of the Decomposition Plan.}
The System 2 reflection process transformed the initial generic plan into a rigorously structured research strategy. Key qualitative improvements include:

\begin{itemize} [leftmargin=10pt]
    \item \textbf{Source Authority \& Tool Selection:} The initial plan relied on generic queries. The final plan explicitly targets authoritative bodies (IPSS, UN projections) and strategically switches to \texttt{academic\_search} for demographic data while retaining \texttt{general\_search} for consumer trends.
    \item \textbf{Analytical Structure:} The system recognized a ``calculation gap''in the original plan. The final output structurally decouples variables, requesting separate data for \textit{per-capita spending} and \textit{population projections}, to ensure the downstream Analyst agent can perform accurate market sizing.
    \item \textbf{Temporal \& Granularity Expansion:} 
    \begin{itemize}
        \item \textit{Temporal:} Added specific time-horizon queries (2020-2030, 2030-2040, 2040-2050) to capture non-linear growth trends over the 30-year period.
        \item \textit{Categorical:} Expanded the taxonomy from 4 to 9 subtopics, adding granular queries for specific consumption categories.
    \end{itemize}
    \item \textbf{Methodological Grounding:} Crucially, the final plan includes a meta-query for ``market sizing calculation frameworks,'' ensuring the research is grounded in established economic methodologies.
\end{itemize}

\paragraph*{Key Insights.}
This trajectory highlights the \textbf{autonomous self-correction} capabilities of the \auq framework:
\begin{enumerate}
    \item \textbf{Iterative Refinement:} The improvement was not instantaneous but cumulative. Each reflection step (Iter 1 $\to$ Iter 2) built upon the identified gaps, progressively closing the knowledge/reasoning gap. 
    \item \textbf{Parallel Exploration:} By maintaining three active branches, the system avoided getting stuck in local optima (e.g., Branch 2 plateaued at 0.93), allowing it to discover the superior reasoning path in Branch 3.
    \item \textbf{Transparent Decision Making:} Unlike black-box optimizations, every improvement is logged with a specific \textit{rationale} (e.g., ``Address data source uncertainty''), providing full interpretability of the agent's metacognitive process.
\end{enumerate}

\subsection{LLM Usage}
We have used LLM to polish writing for this paper.

\end{document}